\documentclass[11pt]{article}

\usepackage{amsmath}
\usepackage{amssymb}
\usepackage{tabularx}
\usepackage{booktabs}
\usepackage{mathtools}
\usepackage{upgreek}
\usepackage{hyperref} 
\usepackage{color,colortbl} 
\usepackage{multirow}
\usepackage{booktabs}
\usepackage{rotating}
\usepackage{soul}
\usepackage{graphicx,import} 
\usepackage{tikz}  
\usepackage[flushleft]{threeparttable}
\usetikzlibrary{decorations.text}
\usepackage[font={small,it}]{caption} 
\usepackage[semicolon,authoryear]{natbib}
\usepackage{algorithm}
\usepackage[noend]{algpseudocode}
\usepackage{enumitem}

\makeatletter 
 
\@addtoreset{algorithm}{section} 
\makeatother

\newcommand{\pos}{\mathrm{pos}}
\newcommand{\ant}{\mathrm{ant}}
\newcommand{\suc}{\mathrm{suc}}
\newcommand{\lb}{\mathrm{lb}}
\newcommand{\ub}{\mathrm{ub}}
\newcommand{\size}{\mathrm{size}}

\newcommand{\cro}{\mathrm{cro}}
\newcommand{\mut}{\mathrm{mut}}
\newcommand{\best}{\mathrm{best}}
\DeclareMathOperator{\delay}{delay} 
\DeclareMathOperator*{\argmin}{argmin}
\DeclareMathOperator*{\argmax}{argmax}

\begin{document}

\title{Metaheuristics for the Online Printing Shop Scheduling Problem}

\author{
  Willian T. Lunardi\thanks{University of Luxembourg, 29 John F Kennedy, L-1855, Luxembourg, Luxembourg. e-mail: willian.tessarolunardi@uni.lu. Corresponding author.}
  \and
  Ernesto G. Birgin\thanks{Department of Computer Science, Institute of Mathematics and Statistics, University of S\~ao Paulo, Rua do Mat\~ao, 1010, Cidade Universit\'aria, 05508-090, S\~ao Paulo, SP, Brazil. e-mail: egbirgin@ime.usp.br}
  \and 
  D\'ebora P. Ronconi\thanks{Department of Production Engineering, Polytechnic School, University of S\~ao Paulo, Av. Prof. Luciano Gualberto, 1380, Cidade Universit\'aria, 05508-010, S\~ao Paulo, SP, Brazil. e-mail: dronconi@usp.br}
  \and
  Holger Voos\thanks{University of Luxembourg, 29 John F Kennedy, L-1855, Luxembourg, Luxembourg. e-mail: holger.voos@uni.lu}
}

\date{June 4, 2020\footnote{Revision made on November 30, 2020.}.}

\maketitle

\begin{abstract}
In this work, the online printing shop scheduling problem is considered. This challenging real-world scheduling problem, that emerged in the present-day printing industry, corresponds to a flexible job shop scheduling problem with sequencing flexibility; and it presents several complicating requirements such as resumable operations, periods of unavailability of the machines, sequence-dependent setup times, partial overlapping between operations with precedence constraints, and fixed operations, among others. A local search strategy and metaheuristics are proposed and evaluated. Based on a common representation scheme, trajectory and populational metaheuristics are considered. Extensive numerical experiments on large-sized instances show that the proposed methods are suitable for solving practical instances of the problem; and that they outperform a half-heuristic-half-exact off-the-shelf solver by a large extent. In addition, numerical experiments on classical instances of the flexible job shop scheduling problem show that the proposed methods are also competitive when applied to this particular case.\\


\end{abstract}

\section{Introduction}\label{sec:introduction}

This paper deals with the online printing shop (OPS) scheduling problem introduced in~\cite{ops1}. The problem is a flexible job shop (FJS) scheduling problem with sequencing flexibility and a wide variety of challenging features, such as non-trivial operations' precedence relations given by an arbitrary directed acyclic graph (DAG), partial overlapping among operations with precedence constraints, periods of unavailability of the machines, resumable operations, sequence-dependent setup times, release times, and fixed operations. The goal is the minimization of the makespan. 

The OPS scheduling problem represents a real-world problem of the present-day printing industry. Online printing shops receive a wide variety of online orders of diverse clients per day. Orders include the production of books, brochures, calendars, cards (business, Christmas, or greetings cards), certificates, envelopes, flyers, folded leaflets, as well as beer mats, paper cups, or napkins, among many others. Naturally, the production of these orders includes a printing operation. Aiming to reduce the production cost, a cutting stock problem is solved to join the printing operations of different placed orders. These merged printing operations are known as \textit{ganging operations}. The production of the orders whose printing operations were ganged constitutes a single job. Operations of a job also include cutting, embossing (e.g., varnishing, laminating, hot foil), and folding operations. Each operation must be processed on one out of multiple machines with varying processing times. Due to their nature, the structure of the jobs, i.e., the number of operations and their precedence relations, as well as the routes of the jobs through the machines, are completely different. Multiple operations of the same type may appear in a job structure. For example, in the production of a book, multiple independent printing operations corresponding to the book cover and the book pages are commonly required. Disassembling and assembling operations are also present in a job structure, e.g., at some point during the production of a book, cover and pages must be gathered together. A simple example of a disassembling operation is the cutting of the printed material of a ganged printing operation. Another example of a disassembling operation occurs in the production of catalogs. Production of catalogs for a franchise usually presents a complex production plan composed of several operations (e.g., printing, cutting, folding, embossing). Once catalogs are produced, the production is branched into several independent sequences of operations, i.e., one sequence for each franchise partner. This is due to the fact that for each partner a printing operation must be performed in the catalog cover to denote the partner's address and other information. Subsequently, each catalog must be delivered to its respective partner.

Several important factors that have a direct impact on the manufacturing system and its efficiency, must be taken into consideration in the OPS scheduling problem. Machines are flexible, which means they can perform a wide variety of tasks. To produce something in a flexible machine requires the machine to be configured. The configuration or setup time of a machine depends on its current configuration and the characteristics of the operation to be processed. A printing operation has characteristics related to the size of the paper, its weight, the required set of colors, and the type of varnishing, among others. Consider now two consecutive operations that are processed on the same machine; the more different the two operations are, the more time consuming the setup will be. Thus, setup operations are sequence-dependent. Working days are divided into three eight-hour shifts, namely, morning, afternoon/evening, and overnight shift, in which different groups of workers perform their duties. However, the presence of all three shifts depends on the working load. When a shift is not present, the machines are considered unavailable. In addition to shift patterns, other situations such as machines' maintenance, pre-scheduling, and overlapping of two consecutive time planning horizons imply machines' downtimes. Operations are resumable, in the sense that the processing of an operation can be interrupted by a period of unavailability of the machine to which the operation has been assigned; the operation being resumed as soon as the machine returns to be active. On the other hand, setup operations cannot be interrupted; the end of a setup operation must be immediately followed by the beginning of its associated regular operation. This is because a setup operation might include cleaning the machine before the execution of an operation. If we assume that a period of unavailability of a machine corresponds to pre-scheduled maintenance, the machine cannot be opened and half-cleaned, the maintenance operation executed, and then the cleaning operation finished after the interruption. The same situation occurs if the period of unavailability corresponds to a night shift during which the store is closed. In this case, the half-cleaned opened machine could get dirty because of dust or insects during the night. Operations that compose a job are subject to precedence constraints. The classical conception of precedence among a pair of operations called predecessor and successor means that the predecessor must be fully processed before the successor can start to be processed. However, in the OPS scheduling problem, some operations connected by a precedence constraint may overlap to a certain predefined extent. For instance, a cutting operation preceded by a printing operation may overlap its predecessor: if the printing operation consists in printing a certain number of copies of something, already printed copies can start to be cut while some others are still being printed. Fixed operations (i.e., with starting time and machine established in advance) can also be present in the OPS. This is due to the fact that customers may choose to visit the OPS to check the quality of the outcome product associated with that operation. This is mainly related to printing quality, so most fixed operations are printing operations. Fixed operations are also useful to assemble the schedule being executed with the schedule of a new planning horizon.

The OPS scheduling problem is NP-hard, since it includes as a particular case the job shop scheduling problem which is known to be strongly NP-hard~\citep{garey}. In this work, a heuristic method able to tackle the large-sized practical instances of the OPS scheduling problem is proposed. First, we extend the local search strategy introduced in~\cite{mastrolilli2000effective} to deal with the FJS scheduling problem. The local search is based on the representation of the operations' precedences as a graph in which the makespan is given by the longest path from the ``source'' to the ``target'' node. In the present work, this underlying graph is extended to cope with the sequencing flexibility and, more relevantly, with resumable operations and machines' downtimes. With the help of the redefined graph, the main idea in~\cite{mastrolilli2000effective}, which consists in defining reduced neighbor sets, is also extended. The reduction of the neighborhood, that greatly speeds up the local search procedure, relies on the fact that the reduction of the makespan of the current solution requires the reallocation of an operation in a \textit{critical path}, i.e., a path that realizes the makespan. With all these ingredients a local search for the OPS scheduling problem is proposed. To enhance the probability of finding better solutions, the local search procedure is embedded in metaheuristic approaches. A relevant ingredient of the metaheuristic approaches is the representation of a solution with two arrays of real numbers of the size of the number of non-fixed operations. One of the arrays represents the assignment of non-fixed operations to machines; while the other represents the sequencing of the non-fixed operations within the machines. This is an indirect representation, i.e., it does not encode a complete solution. Thus, another relevant ingredient is the development of a \textit{decoder}, i.e., a methodology to construct a feasible solution from the two arrays. One of the challenging tasks of the decoder is to sequence the fixed operations besides constructing a feasible semi-active schedule. The representation scheme, the decoder and the local search strategy are evaluated in connection with four metaheuristics. Two of the metaheuristics, genetic algorithms (GA) and differential evolution (DE), are populational methods; while the other two, namely iterated local search (ILS) and tabu search (TS), are trajectory methods. Since the proposed GA and DE include a local search, they can be considered memetic algorithms.

The paper is structured as follows. Section~\ref{sec:litreview} presents a literature review. Section~\ref{sec:prob_description} describes the OPS scheduling problem. Section~\ref{repre} introduces the way in which the two key elements of a solution (assignment of operations to machines and sequencing within the machines) are represented and how a feasible solution is constructed from them. Section~\ref{local} introduces the proposed local search. The metaheuristic approaches are given in Section~\ref{sec:metaheuristic}. Numerical experiments are presented and analyzed in Section~\ref{sec:exp}. Final remarks and conclusions are given in the last section.

\section{Literature review}\label{sec:litreview}

Many works in the literature deal with the FJS scheduling problem; see~\cite{chaudhry2016research} for a recent review and~\cite{cinar2015} for a taxonomy. On the other hand, only a few papers, mostly inspired by practical applications, tackle the FJS scheduling problem with sequencing flexibility. The literature review below aims to show that no published work addressed an FJS scheduling problem with sequencing flexibility including simultaneously all the complicating features that are present in the OPS scheduling problem. As it will be shown in the forthcoming sections, these features are crucial in the development of the proposed method.

The FJS with sequencing flexibility was recently described through mixed integer linear programming (MILP) and constraint programming (CP) formulations. In~\cite{ozguven2010mathematical}, a MILP model for the FJS was considered. This model was adapted to the sequencing flexibility scenario in~\cite{birgin2014milp}, where an alternative MILP model was also presented. In both models, precedence constraints among operations are given by a DAG. A model for an FJS scheduling problem with sequencing and process plan flexibility, in which precedences between operations are given by an AND/OR graph, was proposed in~\cite{lee2012}. The MILP model introduced in~\cite{birgin2014milp} was extended to encompass all the requeriments of the OPS scheduling problem in~\cite{ops1}, where a CP model for the OPS scheduling problem was also proposed. The model proposed in~\cite{birgin2014milp} was extended in a different direction in~\cite{andradescheduling} to consider dual resources (machines and workers with different abilities). 

In~\cite{gan} a practical application of the mold manufacturing industry that can be seen as an FJS scheduling problem with sequencing and process plan flexibility is considered. The problem is tackled with a branch and bound algorithm. The simultaneous optimization of the process plan and the scheduling problem is uncommon in the literature, as well as the usage of an exact method. In~\cite{kim}, where the same problem is addressed, a symbiotic evolutionary algorithm is proposed. (Note that the problem addressed in~\cite{gan} and~\cite{kim} does not possess any of the complicating features of the OPS scheduling problem.) Due to its computational complexity, most papers in the literature tackle the FJS with sequencing flexibility using heuristic approaches. A problem originated in the glass industry is described in~\cite{alvarez2005heuristic}. The problem they addressed includes some of the characteristics of the OPS scheduling problem such as resumable operations, periods of unavailability of the machines, and partial overlapping. In addition, some operations present no-wait constraints. The minimization of a non-regular criterion based on due dates is proposed. To solve the problem, a heuristic method combining priority rules and local search is presented. However, no numerical results are shown and no mathematical formulation of the problem is given. In~\cite{vilcot2008tabu}, a scheduling problem that arises in the printing industry is addressed with a bi-objective genetic algorithm based on the NSGA~II. Unlike in the OPS scheduling problem, in the version of the problem they investigated, operations’ precedence constraints are limited to the case in which each operation can have at most one successor. 

An MILP model for the FJS with sequencing flexibility that allows for precedence constraints given by a DAG was introduced in~\cite{birgin2014milp}. For this problem, heuristic approaches were presented in~\cite{birgin2015list} and~\cite{lunardi2019}. In~\cite{birgin2015list} a list scheduling algorithm and its extension to a beam search method were introduced. In~\cite{lunardi2019}, a hybrid method that combines an imperialist competitive algorithm and tabu search was proposed. In~\cite{rossi}, an FJS scheduling problem in the context of additive/subtractive manufacturing is tackled. Process planning and sequencing flexibility are simultaneously considered. Both features are modeled through a precedence graph with conjunctive and disjunctive arcs and nodes. Numerical experiments using an ant colony optimization procedure aiming to minimize the makespan are presented to validate the proposed approach. With respect to the features of the OPS scheduling problem, only the sequence-dependent setup time is considered. In~\cite{soto}, the minimization of the weighted tardiness and the makespan in an FJS with sequencing flexibility is addressed. Precedences between operations are given by a DAG as introduced in~\cite{birgin2014milp}. For this problem, the authors introduce an MILP model and a biomimicry hybrid bacterial foraging optimization algorithm hybridized with simulated annealing. The method makes use of a local search based on the reallocation of critical operations. Numerical experiments with classical instances and a case study are presented to illustrate the performance of the proposed approach. The considered problem does not include any of the additional characteristics of the OPS scheduling problem. The FJS with sequencing flexibility in which precedences are given by a DAG, and that allows for sequence-dependent setup times, was also considered in~\cite{cao2019b}. For this problem, a knowledge-based cuckoo search algorithm was introduced that exhibits a self-adaptive parameters control based on reinforcement learning. However, other features such as machines' downtimes and resumable operations are absent in the considered problem. The scheduling of repairing orders and allocation of workers in an automobile repair shop is addressed in~\cite{andradescheduling}. The underlying scheduling problem is a dual-resource FJS scheduling problem with sequencing flexibility that aims to minimize a combination of makespan and mean tardiness. For this problem, a constructive iterated greedy heuristic is proposed.

\section{Problem description} \label{sec:prob_description}

\begin{figure}[ht!]
\centering
\resizebox{0.75\textwidth}{!}{
\begin{tabular}{ccc}
\includegraphics{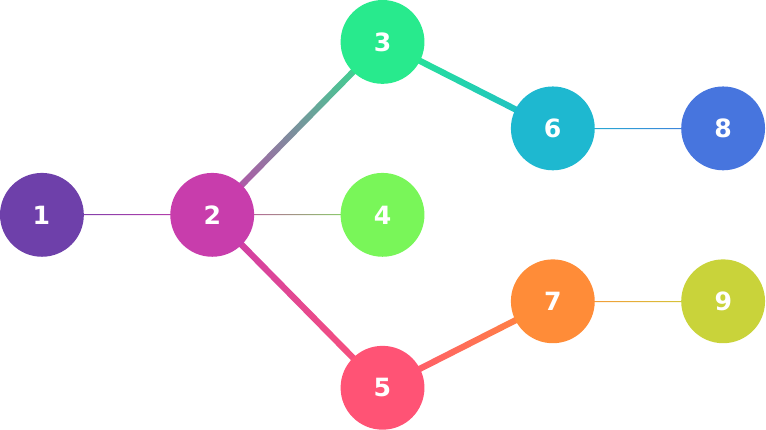} & \quad \quad &
\includegraphics{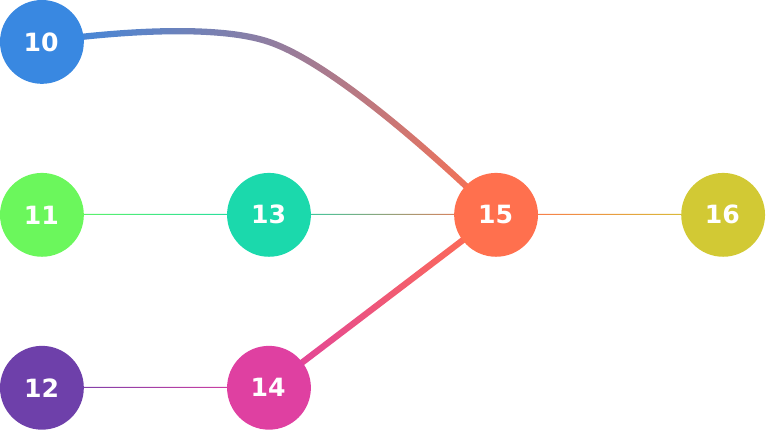} \\ 
& & \\
\small (a) Job 1 with 9 operations & \quad \quad & \small (b) Job 2 with 7 operations
\end{tabular}}
\caption{Directed acyclic graph representing precedence constraints between operations of two different jobs with~$9$ and $7$ operations, respectively. Nodes represent operations and arcs, directed from left to right, represent precedence constraints. Operations are numbered consecutively from~$1$ to~$16$. So, $V=\{1,2,\dots,16\}$ and $A = \{ (1,2), (2,3), (2,4), (2,5), (3,6), (5,7), (6,8), (7,9), (10,15), (11,13), (12,14), (13,15), (14,15), (15,16) \}$.}
\label{fig1}
\end{figure}

In the OPS scheduling problem, there are $n$ jobs and $m$ machines. Each job~$i$ is decomposed into~$o_i$ operations with arbitrary precedence constraints represented by a directed acyclic graph (DAG). For simplicity, it is assumed that operations are numbered consecutively from~$1$ to~$o := \sum_{i=1}^n o_i$; and all~$n$ disjoint DAGs are joined together into a single DAG $(V,A)$, where $V=\{1,2,\dots,o\}$ and~$A$ is the set all arcs of the~$n$ individual DAGs. (See Figure~\ref{fig1}.)  For each operation~$i \in V$, there is a set $F(i) \subseteq \{1,\dots,m\}$ of machines by which the operation can be processed; the processing time of executing operation~$i$ on machine $k \in F(i)$ is given by $p_{ik}$. Each operation~$i$ has a release time~$r_i$.

\begin{figure}[ht!]
\centering
\begin{tabular}{cc}
\multicolumn{2}{c}{\includegraphics{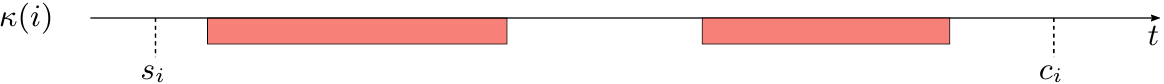}}\\[-3mm]
\multicolumn{2}{c}{\footnotesize{(a)}} \\[2mm]
\includegraphics{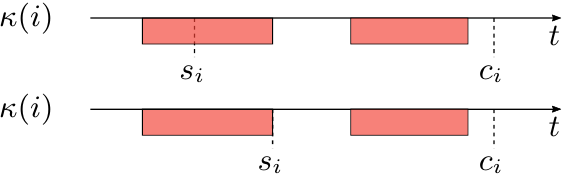} &
\includegraphics{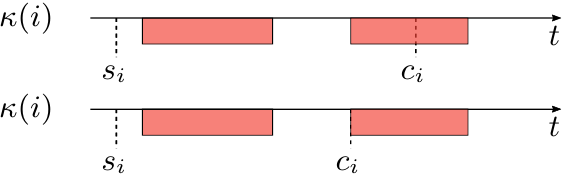} \\
\footnotesize{(b)} & \footnotesize{(c)}
\end{tabular}
\caption{Allowed and forbidden relations between the starting time~$s_i$, the completion time~$c_i$, and the periods of unavailability of machine~$\kappa(i)$. In (a), allowed positions are illustrated. For further reference, it is worth mentioning that the sum of the sizes of the two periods of unavailability in between $s_i$ and~$c_i$ is named~$u_i$; so the relation~$s_i + p_{i,\kappa(i)} + u_i = c_i$ holds. The top picture in (b) shows the forbidden situation $s_i \in [\underline{u}^{\kappa(i)}_{\ell},\bar u^{\kappa(i)}_{\ell})$ for some $\ell$, that corresponds to a fictitious prior starting time. The valid value for~$s_i$ that corresponds to the same situation is illustrated in the bottom picture in (b). The top picture in (c) shows a forbidden situation in which $c_i \in (\underline{u}^{\kappa(i)}_{\ell},\bar u^{\kappa(i)}_{\ell}]$ for some $\ell$, that corresponds to a fictitious delayed completion time. The valid value for~$c_i$ that corresponds to the same situation is illustrated in the bottom picture in (c).}
\label{unavs1amao}
\end{figure}

Machines $k=1,\dots,m$ have periods of unavailability given by $[\underline{u}^k_1,\bar u^k_1], \dots, [\underline{u}^k_{q_k},\bar u^k_{q_k}]$, where $q_k$ is the number of unavailability periods of machine~$k$. Although preemption is not allowed, the execution of an operation can be interrupted by periods of unavailability of the machine to which it was assigned; i.e., operations are resumable. The starting time~$s_i$ of an operation~$i$ assigned to a machine~$\kappa(i)$ must be such that $s_i \notin [\underline{u}^{\kappa(i)}_{\ell},\bar u^{\kappa(i)}_{\ell})$ for all $\ell = 1,\dots,q_{\kappa(i)}$. This means that the starting time may coincide with the end of a period of unavailability (the possible existence of a non-null setup time is being ignored here), but it cannot coincide with its beginning nor belong to its interior, since these two situations would represent a fictitious prior starting time\footnote{If a machine is unavaliable between instants~$5$ and~$10$ and we say the starting time of an operation in this machine is~$7$, then this is a ``fictitious prior starting time'' because the actual starting time is~$10$.}. In an analogous way, the completion time~$c_i$ must be such that $c_i \notin (\underline{u}^{\kappa(i)}_{\ell},\bar u^{\kappa(i)}_{\ell}]$ for all $\ell = 1,\dots,q_{\kappa(i)}$, since violating these constraints would correspond to allowing a fictitious delayed completion time. It is clear that if operation~$i$ is completed at time~$c_i$ and $c_i \in (\underline{u}^{\kappa(i)}_{\ell},\bar u^{\kappa(i)}_{\ell}]$ for some~$\ell$ then it is because the operation is actually completed at instant~$\underline{u}^{\kappa(i)}_{\ell}$; see Figure~\ref{unavs1amao}. 

\begin{figure}[ht!]
\centering 
\includegraphics{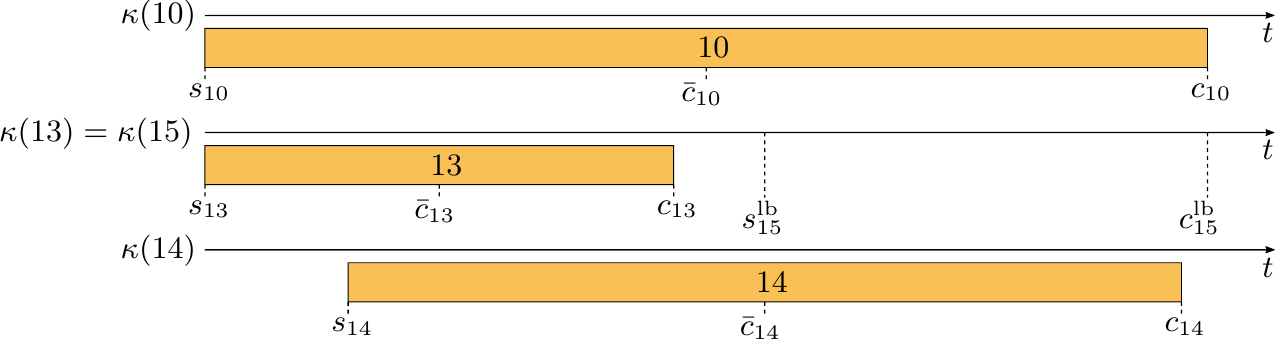}
\caption{According to the DAG in the right-hand-side of Figure~\ref{fig1}, we have $(10,15)$, $(13,15)$, and $(14,15) \in A$. This means that $s^{\lb}_{15} = \max\{ \bar c_{10}, \bar c_{13}, \bar c_{14} \}$ is a lower bound for the starting time~$s_{15}$; while $c^{\lb}_{15} = \max\{ c_{10}, c_{13}, c_{14} \}$ is a lower bound for the completion time~$c_{15}$. If $\kappa(15)=\kappa(13)$ and operation~$15$ is sequenced right after operation~$13$, then $c_{13} + \gamma^I_{13,15,\kappa(15)}$ is another lower bound for $s_{15}$, where $\gamma^I_{13,15,\kappa(15)}$ is the sequence-dependent setup time corresponding to the processing of operation~$13$ right before operation~$15$ on machine~$\kappa(15)$. In addition, $s_{15}$ must also satisfy $s_{15} \geq r_{15}$.}
\label{precedence}
\end{figure}

The precedence relations~$(i,j) \in A$ have a special meaning in the OPS scheduling problem. Each operation~$i$ has a constant~$\theta_i \in (0,1]$ associated with it. On the one hand, the precedence relation means that operation~$j$ can start to be processed after $\lceil \theta_i \times p_{ik} \rceil$ units of time of operation~$i$ have already been processed, where~$k \in F(i)$ is the machine to which operation~$i$ has been assigned. We assume that the given value of~$\theta_i$ is such that the ongoing processing of operation~$i$ does not prevent the regular processing of operation~$j$. (This assumption holds in the real-world instances of the OPS scheduling problem. However, aiming to increase the potential benefit of the overlapping, constants $\theta_i$ ($i \in V$) could be easily substituted with constants $\theta_{i,\kappa(i),j,\kappa(j)}$ ($(i,j) \in A$, $\kappa(i) \in F(i)$, $\kappa(j) \in F(j)$). On the other hand, the precedence relation imposes that operation~$j$ cannot be completed before the completion of operation~$i$. See Figure~\ref{precedence}. In the figure, for a generic operation~$h$ assigned to machine $\kappa(h)$, $\bar c_h$ denotes the instant at which $\lceil \theta_h \times p_{h,\kappa(h)} \rceil$ units of time of operation~$h$ have already been processed. Note that $\bar c_h$ could be larger than $s_h + \lceil \theta_h \times p_{h,\kappa(h)} \rceil$ due to the machines' periods of unavailability.

\begin{figure}[ht!]
\centering
\includegraphics{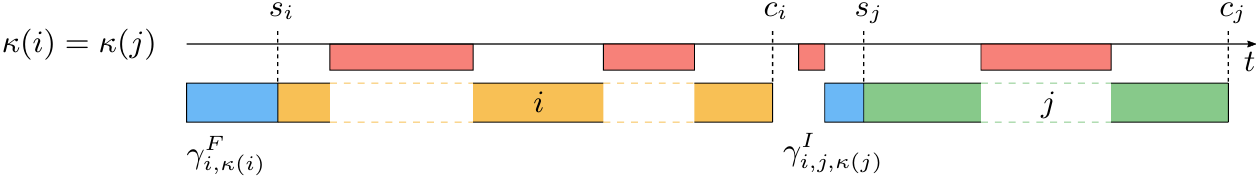}
\caption{Illustration of the fact that, unlike the processing of a regular operation, a setup operation cannot be interrupted by periods of unavailability of the machine to which the operation has been assigned. The picture also illustrates that the completion time of the setup operation must coincide with the starting time of the operation itself. In the picture, it is assumed that operation~$i$ is the first operation to be executed on machine~$\kappa(i)$; thus, the duration of its setup operation is given by $\gamma^F_{i,\kappa(i)}$.}
\label{setupmao}
\end{figure}

Operations have a sequence-dependent setup time associated with them. If the execution of operation~$j$ on machine~$k$ is immediately preceded by the execution of operation~$i$, then its associated setup time is given by~$\gamma^I_{ijk}$ (the super-index ``I'' stands for \textit{intermediate} or \textit{in between}); while, if operation~$j$ is the first operation to be executed on machine~$k$, the associated setup time is given by $\gamma^F_{jk}$ (the super-index ``F'' stands for \textit{first}). Of course, setup times of the form $\gamma^F_{jk}$ are defined if and only if $k \in F(j)$ while setup times of the form $\gamma^I_{ijk}$ are defined if and only if $k \in F(i) \cap F(j)$. Unlike the execution of an operation, the execution of a setup operation cannot be interrupted by periods of unavailability of the corresponding machine, i.e., setup operations are non-resumable. Moreover, the completion time of the setup operation must coincide with the starting time of the associated operation; see Figure~\ref{setupmao}.

Finally, the OPS scheduling problem may have some operations that were already assigned to a machine and for which the starting time has already been defined. These operations are known as \textit{fixed} operations. Note that the setup time of the operations is sequence-dependent. Then, the setup time of a fixed operation is unknown and it depends on which operation (if any) will precede the execution of the fixed operation in the machine to which it was assigned. Let $T \subseteq V$ be the set of indices of the fixed operations. Therefore, we assume that for $i \in T$, $s_i$ is given and that~$F(i)$ is a singleton, i.e., $F(i) = \{ k_i \}$ for some $k_i \in \{1,2,\dots,m\}$. Since a fixed operation~$i$ has already been assigned to a machine~$k_i$, its processing time~$p_i=p_{i,k_i}$ is known. Moreover, the instant $\bar c_i$ that is the instant at which $\lceil \theta_i \times p_i \rceil$ units of time of its execution has already been processed, its completion time~$c_i$, and the value~$u_i$ such that~$s_i + u_i + p_i = c_i$ can be easily computed taking the given starting time~$s_i$ and the periods of unavailability of machine~$k_i$ into account. It is assumed that, if $i \in T$ and $(j,i) \in A$, then $j \in T$, i.e., predecessors of fixed operations are fixed operations as well. This assumption is not present in the MILP formulation of the problem introduced in~\cite{ops1}. However, it is a valid assumption in practical instances of the problem; and assuming it holds eliminates the existence of infeasible instances and simplifies the development of a solution method. For further reference, we define $\bar o = |V| - |T|$, i.e., $\bar o$ is the number of non-fixed operations.

The problem, therefore, consists of assigning the non-fixed operations to the machines and sequencing all the operations while satisfying the given constraints. The objective is to minimize the makespan. Mixed integer linear programming and constraint programming models for the problem were given in~\cite{ops1}.

\section{Representation scheme and construction of a feasible solution} \label{repre}

In this section, we describe (a) the way the assignment of non-fixed operations to machines is represented, (b) the way the sequence of non-fixed operations assigned to each machine is represented and (c) the way a feasible solution is constructed from these two representations. From now on, we assume that all numbers that define an instance of the OPS scheduling problem are integer numbers. Namely, we assume that the processing times $p_{ik}$ ($i \in V$, $k \in F(i)$), the release times~$r_i$ ($i \in V$), the beginning~$\underline{u}^k_{\ell}$ and end~$\bar u^k_{\ell}$ of every period of unavailability of every machine ($k=1,\dots,m$, $\ell=1,\dots,q_k$), the setup times~$\gamma^F_{jk}$ ($j \in V$, $k \in F(j)$) and~$\gamma^I_{ijk}$ ($i,j \in V$, $k \in F(i) \cap F(j)$), and the starting times~$s_i$ of every fixed operation~$i \in T$ are integer values. It is very natural to assume that these constants are rational numbers; and the integrality can be easily obtained with a change of units.

\subsection{Representation of the assignment of non-fixed operations to machines}
\label{pisec}

Let $\{ i_1, i_2, \dots, i_{\bar o} \} = V \setminus T$, with $i_1 \leq i_2 \leq \dots \leq i_{\bar o}$, be the set of non-fixed operations. For each $i_j$, let $K_{i_j} = ( k_{i_j,1}, k_{i_j,2}, \dots, k_{i_j,|F(i_j)|} )$ be a permutation of~$F(i_j)$. Let ${\tilde \pi = ( \tilde \pi_j \in [0,1) : j \in \{1,\dots,\bar o\} )}$ be an array of real numbers that encodes the machine~$k_{i_j,\pi_j}$ to which each non-fixed operation~$i_j$ is assigned, where
\begin{equation}\label{eq:f}
\pi_j = \left\lfloor \tilde \pi_j |F(i_j)| + 1 \right\rfloor, 
\end{equation}  
for $j=1,\dots,\bar o$. For example, given $F(i_j) = \{1,4,7\}$, the permutation $K_{i_j} = ( 1, 4, 7 )$, and $\tilde \pi_j = 0.51$, we have $\pi_j = \lfloor 0.51 \times 3 + 1 \rfloor = 2$, and, thus, $k_{i_j,\pi_j} = k_{i_j,2} = 4$; implying that operation~$i_j$ is assigned to machine~4. For simplicity, we denote $\kappa(i_j) = \kappa_{i_j,\pi_j}$. Then, if we define $\kappa(i)$ as the only element in the singleton~$F(i)$ for the fixed operations~$i \in T$, it becomes clear that the array of real numbers $\tilde \pi = ( \tilde \pi_1, \dots, \tilde \pi_{\bar o} )$ defines a machine assignment $i \rightarrow \kappa(i)$ for $i=1,\dots,o$; see Figure~\ref{repassignment}.

\begin{figure}[ht!]
\centering
\footnotesize
\setlength{\tabcolsep}{3.0pt}
\renewcommand{\arraystretch}{1.25}
\begin{tabular}{c|c|c|c|c|c|c|c|c|c|c|c|c|c|c|}
\cline{2-15}
$j$ & 1 & 2 & 3 & 4 & 5 & 6 & 7 & 8 & 9 & 10 & 11 & 12 & 13 & 14 \\ \cline{2-15} 
$i_j$ & 2 & 3 & 4 & 5 & 6 & 7 & 8 & 9 & 10 & 12 & 13 & 14 & 15 & 16 \\ \cline{2-15} 
$K_{i_j}$ & $(1,2)$ & $(3,4)$ & $(2,4)$ & $(2,4)$ & $(1,2)$ & $(1,3)$ & $(3,4)$ & $(1,2)$ & $(3,4)$ & $(1,3)$ & $(1,3)$ & $(1,2)$ & $(2,4)$ & $(1,3)$ \\ \cline{2-15} 
$\tilde \pi_j$ & 0.05 & 0.79 & 0.48 & 0.26 & 0.17 & 0.53 & 0.99 & 0.09 & 0.95 & 0.63 & 0.52 & 0.02 & 0.31 & 0.62 \\ \cline{2-15} 
$\pi_j$ & 1 & 2 & 1 & 1 & 1 & 2 & 2 & 1 & 2 & 2 & 2 & 1 & 1 & 2 \\ \cline{2-15} 
$\kappa(i_j)$ & 1 & 4 & 2 & 2 & 1 & 3 & 4 & 1 & 4 & 3 & 3 & 1 & 2 & 3 \\ \cline{2-15} 
\end{tabular}
\caption{An arbitrary machine assignment array assuming that operations~$1$ and~$11$ are fixed operations with $F(1)=\{3\}$ and $F(11)=\{2\}$, so $\kappa(1)=3$ and $\kappa(11)=2$.}
\label{repassignment}
\end{figure}

\subsection{Representation of a sequencing of the non-fixed operations}
\label{sigmasec}

Let $\tilde \sigma = ( \tilde \sigma_j \in [0,1) : j \in \{1,\dots,\bar o\} )$ be an array of real numbers that encodes the order of execution of the non-fixed operations that are assigned to the same machine. Consider two non-fixed operations~$i_a$ and $i_b$ such that $\kappa(i_a)=\kappa(i_b)$, i.e., that were assigned to the same machine. If $\tilde \sigma_a < \tilde \sigma_b$ (or $\tilde \sigma_a = \tilde \sigma_b$ and $i_a<i_b$) and if there is no path from~$i_b$ to~$i_a$ in the DAG $(V, A)$, then operation~$i_a$ is executed before operation~$i_b$; otherwise $i_b$ is executed before~$i_a$.
    
Let $\sigma = ( \sigma_j : j \in \{1,\dots,\bar o\} )$ be a permutation of the set of non-fixed operations $\{i_1,\dots, i_{\bar o}\}$ such that, for every pair of non-fixed operations~$\sigma_{j_1}$ and $\sigma_{j_2}$ with $\kappa(\sigma_{j_1})=\kappa(\sigma_{j_2})$, we have that $j_1 < j_2$ if and only if~$\sigma_{j_1}$ is processed before~$\sigma_{j_2}$. The permutation~$\sigma$ can be computed from~$\tilde \sigma$ and the DAG $(V, A)$ as follows: (i) start with $\ell \leftarrow 0$; (ii) let $R \subseteq \{ i_1, i_2, \dots, i_{\bar o}\}$ be the set of non-fixed operations~$i_j$ such that $i_j \neq \sigma_s$ for $s=1,\dots,\ell$ and, in addition, for every arc $(i,i_j) \in A$ we have  $i \in V \setminus T$ and $i = \sigma_t$ for some $t=1,\dots,\ell$ or $i \in T$; (iii) take the operation~$i_j \in R$ with smallest $\tilde \sigma_j$ (in case of a tie, select the operation with the smallest index~$i_j$), set $\sigma_{\ell+1} = i_j$, and $\ell \leftarrow \ell + 1$; and (iv) if $\ell < \bar o$, return back to~(ii). See Figure~\ref{repseq}.

For further reference, for each machine~$k$ we define $\phi_k = ( \phi_{k,1}, \dots, \phi_{k,|\phi_k|} )$ as the subsequence of~$\sigma$ composed of the operations $\sigma_{\ell}$ such that~$\kappa(\sigma_{\ell}) = k$. Given the machine assignment $\tilde \pi$ as illustrated in Figure~\ref{repassignment} and the order of execution within each machine implied by $\tilde \sigma$ as illustrated in Figure~\ref{repseq}, we have $\phi_1 = ( 2, 14, 6, 9 ), \phi_2 = ( 5, 15, 4 ), \phi_3 = ( 12, 13, 7, 16 )$, and $\phi_4 = ( 10, 3, 8 )$. Note that fixed operations are not included. Moreover, we define $\Phi=(\phi_1,\dots,\phi_m)$.

\begin{figure}[ht!]
\centering 
\footnotesize
\setlength{\tabcolsep}{4.0pt}
\renewcommand{\arraystretch}{1.25}
\begin{tabular}{c|c|c|c|c|c|c|c|c|c|c|c|c|c|c|}
\cline{2-15}
$j$ & 1 & 2 & 3 & 4 & 5 & 6 & 7 & 8 & 9 & 10 & 11 & 12 & 13 & 14 \\ \cline{2-15} 
$i_j$ & 2 & 3 & 4 & 5 & 6 & 7 & 8 & 9 & 10 & 12 & 13 & 14 & 15 & 16 \\ \cline{2-15} 
\cline{2-15} 
$\tilde \sigma_j$ & 0.05 & 0.55 & 0.95 & 0.51 & 0.75 & 0.54 & 0.00 & 0.99 & 0.15 & 0.15 & 0.16 & 0.11 & 0.79 & 0.55 \\ 
\cline{2-15} 
$\sigma_j$ & 2 & 10 & 12 & 14 & 13 & 5 & 7 & 3 & 6 & 8 & 15 & 16 & 4 & 9 \\ 
\cline{2-15} 
\end{tabular}
\caption{An operations execution order sequence $\sigma$ produced by considering the values in $\tilde \sigma$ and the precedence relations given by the DAG represented in Figure~\ref{fig1}. Note, once again, that fixed operations~$1$ and~$11$ are unsequenced at this point.}
\label{repseq}
\end{figure}   

\subsection{Construction of a feasible solution and calculation of the makespan} \label{cmaxsec}

Let the machine assignment~$\tilde \pi$ and the execution order~$\tilde \sigma$ be given; and let $\pi$, $\sigma$, $\kappa$, and $\phi_k$ ($k=1,\dots,m$) be computed from~$\tilde \pi$ and~$\tilde \sigma$ as described in Sections~\ref{pisec} and~\ref{sigmasec}. Recall that, for all fixed operations $i \in T$, it is assumed that we already know the starting time~$s_i$, the processing time~$p_i$, the completion time~$c_i$, the value~$u_i$ such that $s_i + u_i + p_i = c_i$, and the ``partial completion time'' $\bar c_i$, that is the instant at which $\lceil \theta_i \times p_i \rceil$ units of time of operation~$i$ have already been processed. We now describe an algorithm to compute $s_i$, $\bar c_i$, $u_i$, $p_i$, and $c_i$ for all $i \in V \setminus T$ and to sequence the fixed operations $i \in T$ in order to construct a feasible schedule. The algorithm also determines for all the operations (fixed and non-fixed) the corresponding sequence-dependent setup time $\xi_i$  and some additional quantities ($d_i$, $s_i^{\lb}$, and $c_i^{\lb}$) whose meaning will be elucidated later. The algorithm processes one non-fixed operation $i \in V \setminus T$ at a time and schedules it as soon as possible (for the given~$\tilde \pi$ and~$\tilde \sigma$), constructing a semi-active schedule. This computation includes sequencing the fixed operations~$i \in T$.


Define $\pos(i)$ as the position of operation~$i$ in the sequence~$\phi_{\kappa(i)}$; i.e., for any non-fixed operation~$i$, we have that $1 \leq \pos(i) \leq |\phi_{\kappa(i)}|$. This means that, according to~$\tilde \pi$ and~$\tilde \sigma$ and ignoring the fixed operations, for a non-fixed operation~$i$, $\ant(i) = \phi_{\kappa(i),\pos(i)-1}$ is the operation that is processed immediately before~$i$ on machine~$\kappa(i)$; and $\ant(i)=0$ if $i$ is the first operation to be processed on the machine. For further reference, we also define $\suc(i) = \phi_{\kappa(i),\pos(i)+1}$ as the immediate successor of operation~$i$ on machine~$\kappa(i)$, if operation~$i$ is not the last operation to be processed on the machine; and $\suc(i)=o+1$, otherwise.

For $k=1,\dots,m$, define the $(o+1) \times o$ matrices $\Gamma^k$ of setup times, with row index starting at~$0$, given by $\Gamma^k_{0j} = \gamma^F_{jk}$ for $j=1,\dots,o$ and $\Gamma^k_{ij} = \gamma^I_{ijk}$ for $i,j=1,\dots,o$. Then we have that, according to~$\phi_k$ (that does not include the fixed operations yet), the setup time~$\xi_i$ of operation~$i$ is given by~$\xi_i = \Gamma^{\kappa(i)}_{\ant(i),i}$. Moreover, if we define $c_0=0$, we obtain $c_{\ant(i)} + \xi_i$ as a lower bound for the starting time~$s_i$ of operation~$i$ on machine~$\kappa(i)$.

The algorithm follows below. In the algorithm, $\size(\cdot)$ is a function that, if applied to an interval $[a,b]$, returns its size given by $b-a$ and, if applied to a set of non-overlapping intervals, returns the sum of the sizes of the intervals.\\[-2mm]

\noindent
\textbf{Algorithm~\ref{cmaxsec}.1.}\\ \vspace{-4mm}

\noindent
\textbf{Input:} $\sigma_i$, $\kappa_i$ ($i \in V$), $\phi_k$ ($k=1,\dots,m$), $s_i$, $u_i$, $p_i$, $\bar c$, $c_i$ ($i \in T$).\\ \vspace{-4mm}

\noindent
\textbf{Output:} $\phi_k$ ($k=1,\dots,m$), $s_i$, $u_i$, $p_i$, $\bar c$, $c_i$ ($i \in V \setminus T$), $\xi_i$, $d_i$, $s_i^{\lb}$, $c_i^{\lb}$ ($i \in V$), $C_{\max}$.\\ \vspace{-4mm}

\noindent
\textbf{For each $\ell=1,\dots, \bar o$, execute Steps~1 to~6. Then execute Step~7.}\\ \vspace{-4mm}
\begin{description}[noitemsep,topsep=0pt,leftmargin=1cm]

\item[\textbf{Step 1:}] Set $i \leftarrow \sigma_{\ell}$, $k \leftarrow \kappa(i)$, $p_i = p_{ik}$, $\bar p_i = \lceil \theta_i \times p_{ik} \rceil$, and $\delay_i \leftarrow 0$ and compute
\begin{equation} \label{sclb}
s_i^{\lb} = \max \left\{ \max_{\{ j \in V | (j,i) \in A\}} \left\{ \bar c_j \right\}, \; r_i \right\}
\; \mbox{ and } \;
c_i^{\lb} = \max_{\{ j \in V | (j,i) \in A\}} \left\{ c_j \right\}.
\end{equation}

\item[\textbf{Step 2:}] Set $\xi_i = \Gamma^k_{\ant(i),i}$, define
\begin{equation} \label{sibounds}
d_i = \max \left\{ s_i^{\lb}, \; c_{\ant(i)} + \xi_i \right\},
\end{equation}
and compute~$s_i \geq d_i + \delay_i$ as the earliest starting time such that the interval $(s_i - \xi_i, s_i]$ does not intersect any period of unavailability of machine~$k$, i.e.,
\begin{equation} \label{siconstr}
\left( \cup_{\ell=1}^{q_k} [\underline{u}_{\ell}^k,\bar u_{\ell}^k] \right) \cap (s_i - \xi_i, s_i] = \emptyset. 
\end{equation}

\item[\textbf{Step 3:}] Compute the completion time~$c_i \not\in (\underline{u}_{\ell}^k,\bar u_{\ell}^k]$, for $\ell=1,\dots,q_k$, such that
\begin{equation} \label{ciconstr}
\size([s_i,c_i]) - u_i = p_i,
\end{equation}
where
\begin{equation} \label{ui}
u_i = \size( [s_i,c_i] \cap ( \cup_{\ell=1}^{q_k} [\underline{u}_{\ell}^k,\bar u_{\ell}^k] ) )
\end{equation}
is the time machine~$k$ is unavailable in between~$s_i$ and~$c_i$.

\item[\textbf{Step 4:}] Let~$f \in T$ be an operation fixed at machine~$k$ such that 
\begin{equation} \label{fixed}
c_{\ant(i)} \leq s_f < c_i + \Gamma^k_{if}.
\end{equation}
If there is none, go to Step~5. If there is more than one, consider the one with the earliest starting time~$s_f$. Insert~$f$ in~$\phi_k$ in between operations~$\ant(i)$ and~$i$ and go to Step~2. (Note that this action automatically redefines $\ant(i)$ as~$f$.)

\item[\textbf{Step 5:}] If $c_i \not\geq c_i^{\lb}$ then set $\delay_i
  = \size([c_i,\hat c_i^{\lb}]) - \size([c_i,\hat c_i^{\lb}] \cap
  \left( \cup_{\ell=1}^{q_k} [\underline{u}_{\ell}^k,\bar u_{\ell}^k]
  \right) )$, where
\[
\hat c_i^{\lb} = 
\left\{
\begin{array}{ll}
c_i^{\lb}, & \mbox{if } c_i^{\lb} \not\in (\underline{u}_{\ell}^k,\bar u_{\ell}^k] \mbox{ for } \ell=1,\dots,q_k,\\[2mm]
\bar u_{\ell}^k + 1, & \mbox{if } c_i^{\lb} \in (\underline{u}_{\ell}^k,\bar u_{\ell}^k] \mbox{ for some } \ell \in \{ 1,\dots,q_k \},
\end{array}
\right.
\]
and go to Step~2. 

\item[\textbf{Step 6:}] Compute the ``partial completion time'' $\bar c_i \not\in (\underline{u}_{\ell}^k,\bar u_{\ell}^k]$, for $\ell=1,\dots,q_k$, such that 
$\size([s_i, \bar c_i]) - \bar u_i = \bar p_i$,
where
$\bar u_i = \size( [s_i, \bar c_i] \cap (\cup_{\ell=1}^{q_k} [\underline{u}_{\ell}^k,\bar u_{\ell}^k]))$.

\item[\textbf{Step 7:}] Compute $C_{\max} = \max_{i \in V} \{ c_i \}$. For each unsequenced operation~$f \in T$, sequence it according to its starting time~$s_f$, update $\phi_{\kappa(f)}$, compute $s_f^{\lb}$ and $c_f^{\lb}$ according to~(\ref{sclb}), $\xi_f = \Gamma_{\ant(f)}^{\kappa(f)}$, and $d_f$ as in~(\ref{sibounds}).

\end{description}


At Step~1, a lower bound~$s_i^{\lb}$ to~$s_i$ is computed based on the release time~$r_i$ and the partial completion times~$\bar c_j$ of the operations~$j$ such that $(j,i) \in A$ exists. In an analogous way, a lower bound~$c_i^{\lb}$ to~$c_i$ is computed, based on the completion times~$c_j$ of the operations~$j$ such that $(j,i) \in A$ exists. 

At Step~2, a tentative~$s_i$ is computed. At this point, it is assumed that the operation which is executed immediately before~$i$ on machine~$\kappa(i)$ is the one that appears right before it in~$\phi_k$ (namely~$\ant(i)$); and, for this reason, it is considered that the setup time of operation~$i$ is given by $\xi_i = \Gamma^k_{\ant(i),i}$. (This may not be the case if it is decided that a still-unsequenced fixed operation should be sequenced in between them.) The computed~$s_i$ is required by~(\ref{sibounds}) to be not smaller than (a) its lower bound~$s_i^{\lb}$ computed at Step~1 and (b) the completion time~$c_{\ant(i)}$ of operation~$\ant(i)$ plus the setup time~$\xi_i$. Note that if operation~$i$ is the first operation to be processed on machine~$\kappa(i)$ then~$\ant(i)=0$ and, by definition, $c_{\ant(i)}=c_0=0$. At this point, we assume that $\delay_i=0$. Its role will be elucidated soon. In addition to satisfying the lower bounds (a) and (b), $s_i$ is required in~(\ref{siconstr}) to be such that (i) it does not coincide with the beginning of a period of unavailability, (ii) there is enough time right before~$s_i$ to execute the setup operation, and (iii) the setup operation is not interrupted by periods of unavailability of the machine. We pick~$s_i$ as the smallest value that satisfies the lower bounds (a) and (b) and conditions (i), (ii), and (iii) mentioned above. Therefore, it becomes clear that there is only a finite number---in fact, a small number---of possibilities for~$s_i$ that depends on the imposed lower bounds and the periods of unavailability of the machine.

Once the tentative~$s_i$ has been computed in Step~2, Step~3 is devoted to the computation of its companion completion time~$c_i$. Basically, ignoring the possible existence of fixed operations on the machine, (\ref{ciconstr}) and~(\ref{ui}) indicate that~$c_i$ is such that between~$s_i$ and $c_i$ the time during which machine~$\kappa(i)$ is available is exactly the time required to process operation~$i$. In addition, $c_i \not\in (\underline{u}_{\ell}^k,\bar u_{\ell}^k]$, for $\ell=1,\dots,q_k$, says that, if the duration of the interval yields $c_i \in [\underline{u}_{\ell}^k,\bar u_{\ell}^k]$ for some $\ell \in \{1,\dots,q_k\}$, we must take $c_i=\underline{u}_{\ell}^k$, since any other choice would artificially increase the completion time of the operation. 

\begin{figure}[ht!]
\centering 
\includegraphics{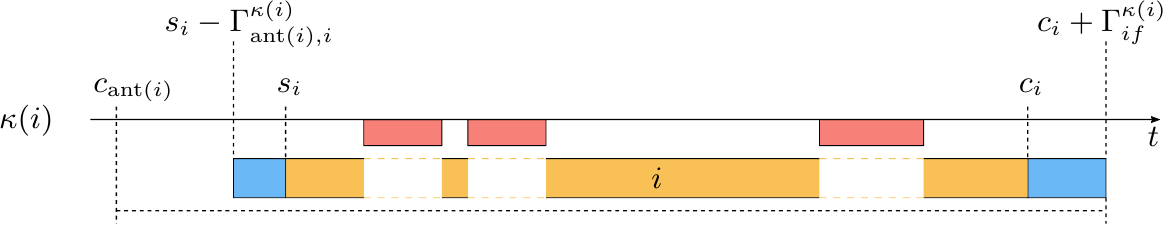}
\caption{If a fixed operation~$f$ on machine~$\kappa(i)$ exists such that $c_{\ant(i)} \leq s_f < c_i + \Gamma^{\kappa(i)}_{if}$, it means that there is not enough space for operation~$i$ after $\ant(i)$ and before~$f$. Thus, the unsequenced fixed operations~$f$ must be sequenced in between operations~$\ant(i)$ and~$i$.}
\label{fixedfig}
\end{figure}

In Step~4 it is checked whether the selected interval~$[s_i,c_i]$ is infeasible due to the existence of a fixed operation on the machine. If there is not a fixed operation~$f$ satisfying~(\ref{fixed}) then Step~4 is skipped. Note that~$c_{\ant(i)}$ is the completion time of the last operation scheduled on machine~$\kappa(i)$. This means that if a fixed operation~$f$ exists such that $s_f \geq c_{\ant(i)}$, the fixed operation~$f$ is still unsequenced. The non-existence of a fixed operation~$f$ satisfying~(\ref{fixed}) is related to exactly one of the following two cases:
\textbf{(a)} there are no fixed operations on machine~$\kappa(i)$ or all fixed operations on machine~$\kappa(i)$ have already been sequenced; and
\textbf{(b)} the starting time~$s_f$ of the closest unsequenced fixed operation~$f$ on machine~$\kappa(i)$ is such that operation~$i$ can be scheduled right after operation~$\ant(i)$, starting at~$s_i$, being completed at~$c_i$ and, after~$c_i$ and before~$s_f$ there is enough time to process the setup operation with duration~$\Gamma^{\kappa(i)}_{if}$.
Assume now that at least one fixed operation satisfying~(\ref{fixed}) exists and let~$f$ be the one with smallest~$s_f$. This means that to schedule operation~$i$ in the interval~$[s_i,c_i]$ is infeasible; see Figure~\ref{fixedfig}. Therefore, operation~$f$ must be sequenced right after~$\ant(i)$, by including it in $\phi_{\kappa(i)}$ in between~$\ant(i)$ and~$i$. This operation transforms~$f$ in a sequenced fixed operation that automatically becomes~$\ant(i)$, i.e., the operation sequenced on machine~$\kappa(i)$ right before operation~$i$. With the redefinition of~$\ant(i)$, the task of determining the starting and the completion times of operation~$i$ must be restarted. This task restarts returning to Step~2, where a new setup time for operation~$i$ is computed and a new~$c_{\ant(i)}$ is considered in~(\ref{sibounds}). Since the number of fixed operations is finite and the number of unsequenced fixed operations is reduced by one, this iterative process ends in a finite amount time.

\begin{figure}[ht!]
\centering
\begin{tabular}{cc} 
    \includegraphics{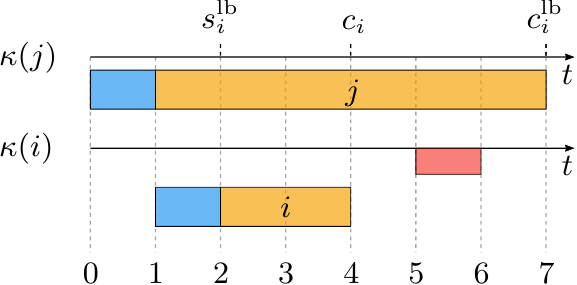} & \includegraphics{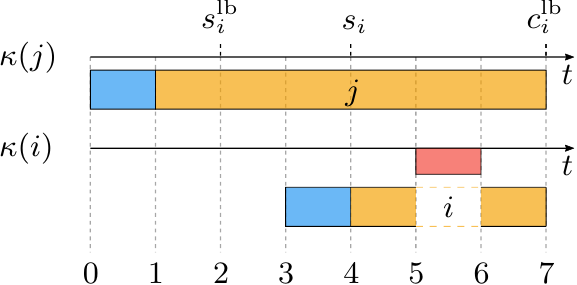} \\[2mm]
    \footnotesize{(a)} & \footnotesize{(b)} \\ [2mm]
    \includegraphics{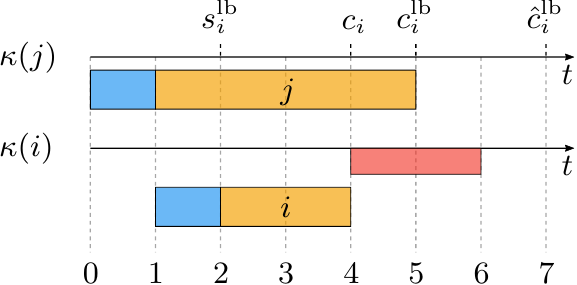} & \includegraphics{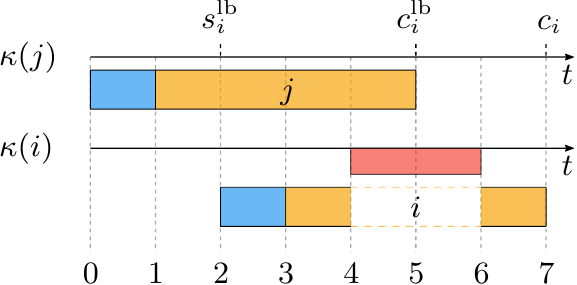} \\[2mm]
    \footnotesize{(c)} & \footnotesize{(d)}\\[-2mm]
\end{tabular}
\caption{Delay computation for the case in which $c_i \not\geq c_i^{\lb}$. In case~(a), $\hat c_i^{\lb}= c_i^{\lb}$ and machine~$\kappa(i)$ has two units of available time in between $c_i$ and $c_i^{\lb}$. Adding this delay to the lower bound of~$s_i$ results in the feasible schedule (of operation~$i$) depicted in~(b). In case~(c), $c_i^{\lb} \in (\underline{u}_{\ell}^{\kappa(i)},\bar u_{\ell}^{\kappa(i)}] \mbox{ for some } \ell \in \{ 1,\dots,q_{\kappa(i)} \}$. Thus, $\hat c_i^{\lb} = \bar u_{\ell}^{\kappa(i)} + 1$. Machine~$\kappa(i)$ has one unit of available time in between $c_i$ and $\hat c_i^{\lb}$. Adding this delay to the lower bound of~$s_i$ results in the feasible schedule (of operation~$i$) depicted in~(d).}
\label{delayfig}
\end{figure}

Step~5 is devoted to checking whether the computed completion time~$c_i$ is smaller than its lower bound~$c_i^{\lb}$, computed at Step~1, or not. If $c_i \geq c_i^{\lb}$, the algorithm proceeds to Step~6. In case $c_i < c_i^{\lb}$, the starting time of operation~$i$ must be delayed. This is the role of the variable~$\delay_i$ that was initialized with zero. If the extent of the delay is too short, the situation may repeat. If the extent is too long, the starting of the operation may be unnecessarily delayed.  Figure~\ref{delayfig} helps to visualize that the time during which machine~$\kappa(i)$ is available in between~$c_i$ and~$c_i^{\lb}$ is the minimum delay that is necessary to avoid the same situation when a new tentative~$s_i$ and its associated~$c_i$ are computed. So, the delay is computed and a new attempt is done by returning to Step~2; this time with a non-null~$\delay_i$.

When the algorithm arrives at Step~6, feasible values for $s_i$ and $c_i$ have been computed and we simply compute the partial completion time~$\bar c_i$ that will be used for computing the starting and completion times of the forthcoming operations.

While executing Steps~1--6 for $\ell=1,\dots,\bar o$, i.e., while scheduling the unfixed operations, some fixed operations have to be sequenced as well. However, when the last unfixed operation is scheduled, it may be the case that some fixed operations, that were scheduled ``far after'' the largest completion time of the unfixed operations, played no role in the scheduling process and thus remain unsequenced, i.e., these fixed operations are not in $\phi_k$ for any~$k$. These unsequenced fixed operations are sequenced in Step~7.

\section{Local search} \label{local}

Given an initial solution, a local search procedure is an iterative process that constructs a sequence of solutions in such a way that each solution in the sequence is in the \textit{neighborhood} of its predecessor in the sequence. The neighborhood of a solution is given by all solutions obtained by applying a \textit{movement} to the solution. A movement is a simple modification of a solution. In addition, the local search described in the current section is such that each solution in the sequence improves the objective function value of its predecessor. In the remainder of the current section, the neighbourhood and the movement introduced in~\cite{mastrolilli2000effective} for the FJS are extended to deal with the OPS scheduling problem.
    
The definition of the proposed movement is based on the representation of a solution by a digraph. Let~$\tilde \pi$, encoding the machine assignment of the non-fixed operations, and~$\tilde \sigma$, encoding the order of execution of the non-fixed operations within each machine, be given. Moreover, assume that, using Algorithm~\ref{cmaxsec}.1, $\xi_i$, $d_i$, $s_i$, $u_i$, $p_i$, $\bar c_i$, $c_i$, $s_i^{\lb}$, $c_i^{\lb}$, and $d_i$ have been computed for all $i=1,\dots,o$. From now on, $\varsigma(\tilde \pi, \tilde \sigma) = (\tilde \pi, \tilde \sigma, \pi, \sigma, \kappa, \Phi, \xi, d, s, u, p, \bar c, c, s^{\lb}, c^{\lb})$ represents a feasible solution. (Recall that $\pi$ is computed from $\tilde \pi$ as defined in~(\ref{eq:f}); $\sigma$ and $\Phi$ are computed from $\tilde \sigma$ as described in Section~\ref{sigmasec}; and $\kappa(i) = \kappa_{i,\pi_i}$.) Let $\suc(i) = \phi_{\kappa(i),\pos(i)+1}$ be the successor of operation~$i$ on machine~$\kappa(i)$, if operation~$i$ is not the last operation to be processed on the machine; and $\suc(i)=o+1$, otherwise. Recall that we already defined $\ant(i) = \phi_{\kappa(i),\pos(i)-1}$, if $i$ is \textit{not} the first operation to be processed on machine~$\kappa(i)$; while $\ant(i) = 0$, otherwise. This means that, for any $i \in V$, i.e., including non-fixed and fixed operations, $\ant(i)$ and $\suc(i)$ represent, respectively, the operations that are processed right before~$i$ (antecedent) and right after~$i$ (sucessor) on machine $ \kappa(i)$.

\begin{figure}[ht!]
\centering
\includegraphics[scale=0.7]{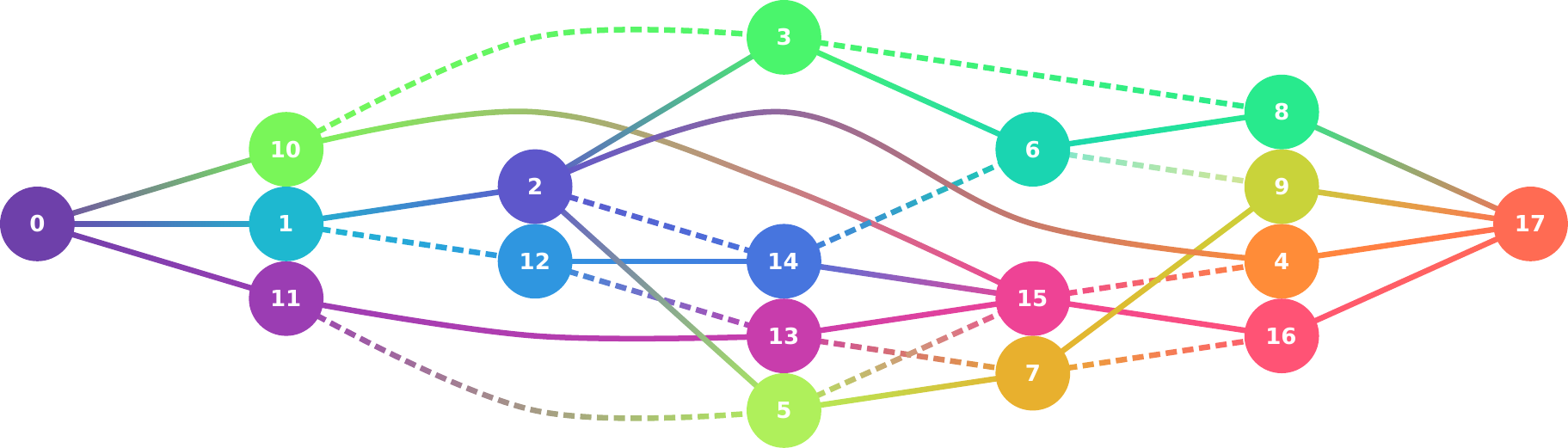}
\caption{Directed acyclic graph $D(\varsigma) = ( V \cup \{0,o+1\}, A \cup W \cup U)$ associated with the original precedence relations (in solid lines) illustrated in Figure~\ref{fig1} plus the precedence relations implied by the machine assignment~$\tilde \pi$ in Figure~\ref{repassignment} and the order of execution within each machine implied by~$\tilde \sigma$ in Figure~\ref{repseq} (dashed lines). Arcs are directed from left to right.}
\label{DAGwithW}
\end{figure}

The weighted augmented digraph that represents the feasible solution~$\varsigma$ is given by 
$D(\varsigma) = ( V \cup \{0,o+1\}, A \cup W \cup U)$,
where
$W = \left\{ (\phi_{k,\ell-1}, \phi_{k,\ell}) \; | \; k \in \{ 1,\dots,m \} \mbox{ and } \ell \in \{ 2,...,|\phi_k|\} \right\}$
and $U$ is the set of arcs of the form $(0,i)$ for every $i \in V$ such that $\ant(i)=0$ plus arcs of the form $(i,o+1)$ for every $i \in V$ such that $\suc(i)=o+1$; see Figure~\ref{DAGwithW}. The weights on the nodes and arcs of $D(\varsigma)$ are defined as follows:
\textbf{(a)} arcs $(j,i) \in A$ have weight $\bar c_j - c_j$;
\textbf{(b)} arcs $(\ant(i),i) \in W$ have weight $\xi_i$;
\textbf{(c)} arcs $(0,i) \in U$ have weight $\max\{r_i,\xi_i\}$;
\textbf{(d)} arcs $(i,o+1) \in U$ have null weight;
\textbf{(e)} each node $i \in V$ has weight $s_i - d_i + u_i + p_i$;
\textbf{(f)} nodes $0$ and $o+1$ have null weight.

Weights of nodes and arcs are defined in such a way that, if we define the weight of a path $i_1, i_2, \dots, i_q$ as the sum of the weights of nodes $i_2, i_3, \dots, i_q$ plus the sum of the weights of arcs $(i_1,i_2), \dots, (i_{q-1}, i_q)$, then the value of the completion time~$c_i$ of operation~$i$ is given by some longest path from node~$0$ to node~$i$. (If in between two nodes~$a$ and~$b$ there is more than one arc then the arc with the largest weight must be considered. This avoids naming the arcs explicitly when mentioning a path.) It follows that the weight of some longest path from $0$ to $o+1$ equals $C_{\max}$ and the nodes on this path are called critical nodes or \textit{critical operations}. We define~$t_i$ as the weight of a longest path from node~$i$ to node~$o+1$. The value $t_i$ (so-called tail time) gives a lower bound on the time elapsed between~$c_i$ and~$C_{\max}$. It is worth noticing that (a) if an operation~$i$ is critical then $c_i + t_i = C_{\max}$ and that (b) if there is a path from~$i$ to~$j$ then $t_i \geq t_j$.

Assume that $\sigma^{\mathrm{ifo}}$ (``ifo'' stands for ``including fixed operations'') is a permutation of $\{1, 2, \dots, o \}$ that represents the order in which operations (non-fixed and fixed) where scheduled by Algorithm~\ref{cmaxsec}.1. This means that non-fixed operations have in~$\sigma^{\mathrm{ifo}}$ the same relative order they have in $\sigma$ and that $\sigma^{\mathrm{ifo}}$ corresponds to~$\sigma$ with the fixed operations inserted in the appropriate places. Note that $\sigma^{\mathrm{ifo}}$ can be easily obtained with a simple modification of Algorithm~\ref{cmaxsec}.1: start with $\sigma^{\mathrm{ifo}}$ as an empty list and every time an operation (non-fixed or fixed) is scheduled, add $i$ to the end of the list. We now describe a simple way to compute $t_i$ for all $i \in V \cup \{ 0, o+1 \}$. Define $c_{o+1} = C_{\max}$ and $t_{o+1}=0$ and for $\ell=o,\dots,1$, i.e., in decreasing order, define $i = \sigma^{\mathrm{ifo}}_{\ell}$ and
\begin{equation} \label{ti}
t_i = \max \left\{ t_{\suc(i)} + \omega(\suc(i)) + \omega(i,\suc(i)), \max_{\{ j \in V | (i,j) \in A\}} \left\{ t_j + \omega(j) + \omega(i,j) \right\} \right\},
\end{equation}
where $\omega(\cdot)$ and $\omega(\cdot,\cdot)$ represent the weight of a node or an arc, respectively.
Finish defining
\begin{equation} \label{t0}
t_0 = \max_{\{ j \in V | (0,j) \in U\}} \left\{ t_j + \omega(j) + \omega(0,j) \right\}.
\end{equation}
In addition to the tail times, the local search strategy also requires identifying a longest (critical) path from node~$0$ to node~$o+1$, since operations on that path are the critical operations whose reallocation will be attempted. A critical path can be obtained as follows. Together with the computation of~(\ref{ti}), define $\mathrm{next}(i)$ as the index in $\{ \suc(i) \} \cup \{ j \; | \; (i,j) \in A \}$ such that $t_i = t_{\mathrm{next}(i)} + \omega(\mathrm{next}(i)) + \omega(i,\mathrm{next}(i))$, i.e., the one that realizes the maximum. Analogously, together with~(\ref{t0}) define $\mathrm{next}(0) = \argmax_{\{ j \in V | (0,j) \in A\}} \left\{ t_j + \omega(0,j) \right\}$. A longest path is then given by
$0$,
$\mathrm{next}(0)$,
$\mathrm{next}(\mathrm{next(0)})$,
$\mathrm{next}(\mathrm{next}(\mathrm{next(0)}))$,
$\dots,$
$o+1$.

\subsection{Movement: Reallocating operations}

Let~$i$ be a (non-fixed) operation to be removed and reallocated. It can be reallocated in the same machine~$\kappa(i)$, but in a different position in the sequence, or in a different machine $k \in F(i)$, $k \neq \kappa(i)$. Removing~$i$ from $\kappa(i)$ implies removing arcs $(\phi_{\kappa(i),\pos(i)-1},i)$ and $(i,\phi_{\kappa(i),\pos(i)+1})$ from $W \cup U$ and including the arc~$(\phi_{\kappa(i),\pos(i)-1}, \phi_{\kappa(i),\pos(i)+1})$ in~$W$ or $U$. (Whether the arcs to be removed or inserted belong to $W$ or $U$ depends on whether $\pos(i)-1=0$, $\pos(i)+1=o+1$, or none of these two cases occur.) In the same sense, reallocating~$i$ implies creating two new arcs and deleting an arc. Let $D(\varsigma)^{-i}$ be the digraph after the removal of the critical operation~$i$; and let $D(\varsigma)^{+i}$ be the digraph after its reallocation.

The relevant fact in the reallocation of operation~$i$ is avoiding the creation of a cycle in~$D(\varsigma)^{+i}$, i.e., the construction of a feasible solution. For each $k \in F(i)$, we define the sets of operations
$R_k = \{ j \in \phi_k \; | \; \bar c_j > s_i^{\lb} \}$
and
$L_k = \{ j \in \phi_k \; | \; t_j + u_j + p_j > C_{\max} - \bar c_i^{\ub} \}$, 
where
$\bar c_i^{\ub} = \min_{(i,j) \in A} \{ s_j \}$
is an upper bound for $\bar c_i$ and, thus, $C_{\max} - \bar c_i^{\ub}$ is a lower bound for the time between $\bar c_i$ and $C_{\max}$.
Properties of $R_k$ and $L_k$ follow:
\begin{description}[noitemsep,topsep=2mm]
\item[R1] If $j \in R_k$ then $\bar c_j > s_i^{\lb}$. Assume that there is a path from $j$ to $i$ in $D(\varsigma)^{-i}$. By the definition of $s_i^{\lb}$, $\bar c_j > s_i^{\lb}$ implies that $(j,i) \not\in A$. Then, in the path from $j$ to $i$, the immediate predecessor of~$i$ must be an operation~$j' \not\in R_k$ and such that $(j',i) \in A$, i.e., such that $\bar c_{j'} \leq s_i^{\lb}$. Therefore, we must have $\bar c_j \leq s_{j'} < \bar c_{j'} \leq s_i^{\lb}$. Thus, if $j \in R_k$ then  there is no path from $j$ to $i$ in $D(\varsigma)^{-i}$. 
\item[R2] If $j \in \phi_k \setminus R_k$ then $s_j < \bar c_j \leq s_i^{\lb} \leq s_i < \bar c_i$. Therefore, there is no path from $i$ to $j$ in $D(\varsigma)^{-i}$. 
\end{description}
\begin{description}[noitemsep,topsep=2mm]
\item[L1] If $j \in L_k$ then $t_j + u_j + p_j > C_{\max} - \bar c_i^{\ub}$. If there were a path from $i$ to $j$ in $D(\varsigma)^{-i}$ then $\bar c_i \leq s_j$ and, therefore, the lower bound on the distance between $\bar c_i$ and $C_{\max}$, given by $C_{\max} - \bar c_i^{\ub}$, should be greater than or equal to the lower bound of the distance between $s_j$ and $C_{\max}$, given by $t_j + u_j + p_j$. Therefore, if $j \in L_k$ then there is no path from $i$ to $j$ in $D(\varsigma)^{-i}$.

\item[L2] If $j \in \phi_k \setminus L_k$ then $C_{\max} - \bar c_i^{\ub} \geq t_j + u_j + p_j$. Assume that there is a path from $j$ to $i$ in $D(\varsigma)^{-i}$. Then, we must have $s_j < s_i$ and, since $\theta_i>0$ and, in consequence, $s_i < \bar c_i$, it follows that $s_j < \bar c_i$. This means that the distance between~$s_j$ and~$C_{\max}$ is greater than the distance between~$\bar c_i$ and~$C_{\max}$. The latter, by definition, is bounded from below by~$C_{\max} - \bar c_i^{\ub}$, i.e., $t_j + u_j + p_j > C_{\max} - \bar c_i^{\ub}$. Thus, if $j \in \phi_k \setminus L_k$ then there is no path from~$j$ to $i$ in $D(\varsigma)^{-i}$.
\end{description}
Properties R1, R2, L1, and L2 imply that if operation~$i$ is reallocated in the sequence of a machine~$k \in F(i)$ in a position such that all operations in $L_k \setminus R_k$ are to the left of~$i$ and all operations in $R_k \setminus L_k$ are to the right of~$i$, then this insertion defines a feasible solution, i.e., $D(\varsigma)^{+i}$ has no cycles. 



\subsection{Neighborhood} \label{neighborhood}

It is well known in the scheduling literature that removing and reallocating a non-critical operation does not reduce the makespan of the current solution. Therefore, in the present work, we define as neighborhood of a solution~$\varsigma$ the set of (feasible) solutions that are obtained when each critical operation~$i$ is removed and reallocated in all possible positions of the sequence of every machine $k \in F(i)$, as described in the previous section. This means that, for each critical operation~$i$, we proceed as follows: (i) operation~$i$ is removed from machine~$\kappa(i)$; (ii) for each $k \in F(i)$, (iia) the sets $R_k$ and $L_k$ are determined and (iib) operation~$i$ is reallocated in the sequence of machine~$k$ in every possible position such that all operations in $L_k \setminus R_k$ are to the left of~$i$ and all operations in $R_k \setminus L_k$ are to the right of~$i$. For further reference, the set of neighbours of~$\varsigma$ is named ${\cal N}(\varsigma)$.

\subsection{Estimation of the makespan of neighbor solutions} \label{estimatedmks}

Given the sequences $\tilde \pi$ and $\tilde \sigma$ of the current solution~$\varsigma$, computing the sequences $\tilde \pi'$ and $\tilde \sigma'$ (as well as $\pi'$, $\sigma'$, and $\kappa'$) associated with a neighbour solution $\varsigma' \in {\cal N}(\varsigma)$ is a trivial task. Computing the makespan (together with the quantities $\xi'$, $s'$, $u'$, $p'$, $\bar c'$, $c'$, $s^{\lb'}$, $c^{\lb'}$) associated with $\varsigma'$ is also simple, but it requires executing Algorithm~\ref{cmaxsec}.1, which might be considered an expensive task in this context. Therefore, the selection of a neighbor is based on the computation of an \textit{estimation} of its associated makespan. In fact, following~\cite{mastrolilli2000effective}, what is used as an estimation of the makespan is an estimation of the length of a longest path from node~$0$ to node~$o+1$ in $D(\varsigma')$ containing the operation that was reallocated to construct~$\varsigma'$ from~$\varsigma$. The exact length of this path is a lower bound on the makespan associated with~$\varsigma'$.
 
The estimation of the makespan of a neighbour solution $\varsigma' \in {\cal N}(\varsigma)$ obtained by removing and reallocating operation~$i$ somewhere in the sequence of machine~$k$ is determined as follows. If $L_k \cap R_k = \emptyset$ then the estimation of the makespan is given by $s_i^{\lb} + p_{ik} + C_{\max} - \bar c_i^{\ub}$. If $L_k \cap R_k \neq \emptyset$, consider the elements (operations) in $L_k \cap R_k$ sorted in increasing order of their starting times; and let $\tau : \{1,\dots,| L_k \cap R_k|\} \rightarrow L_k \cap R_k$ be such that $s_{\tau(1)} < s_{\tau(2)} < \dots < s_{\tau(| L_k \cap R_k|)}$ and, in consequence, $t_{\tau(1)} > t_{\tau(2)} > \dots > t_{\tau(| L_k \cap R_k|)}$. Let~$j$ be such that $j=0$ if operation~$i$ is being inserted before operation~$\tau(1)$ and $1 \leq j \leq | L_k \cap R_k|$ if operation~$i$ is being inserted right after operation $\tau(j)$. In this case, the estimation of the makespan is given by
\[
p_{ik} +
\left\{ 
\begin{array}{ll}
s_i^{\lb} + p_{\tau(1)} + u_{\tau(1)} + t_{\tau(1)}, & \mbox{if } j = 0, \\
s_{\tau(j)} + p_{\tau(j)} + u_{\tau(j)} + p_{\tau(j+1)} + u_{\tau(j+1)} + t_{\tau(j+1)}, & 
\mbox{if } 1 \leq j < | L_k \cap R_k|,  \\
s_{\tau(j)} + p_{\tau(j)} + u_{\tau(j)} + C_{\max} - \bar c_i^{\ub}, & \mbox{if } j = | L_k \cap R_k|. 
\end{array} 
\right.
\]
These estimations follow very closely those introduced by~\cite{mastrolilli2000effective} for the FJS, see~\cite[\S5]{mastrolilli2000effective} for details.

\subsection{Local search procedure}\label{localsec}

The local search procedure starts at a given solution. It identifies all critical operations (operations in the longest path from node~$0$ to node~$o+1$) and for each critical operation~$i$ and each~$k \in F(i)$ it computes the estimation of the makespan associated with removing and reallocating operation~$i$ in every possible position of the sequence of machine~$k$ (as described in the previous sections). The neighbor with the smallest estimation of the makespan is selected and its actual makespan is computed by applying Algorithm~\ref{cmaxsec}.1. In case this neighbor solution improves the makespan of the current solution, the neighbor solution is accepted as the new current solution and the iterative process continues. Otherwise, the local search stops.

\section{Metaheuristics}\label{sec:metaheuristic}

In this section, we briefly describe the four metaheuristics that we consider. Two of the metaheuristics, namely genetic algorithm (GA) and differential evolution (DE) are populational methods; while the other two, iterated local search (ILS) and tabu search (TS), are trajectory methods. GA and TS were chosen because they are the two most popular metaheuristics applied to the FJS scheduling problem (see~\cite[Table~4]{chaudhry2016research}). On the other hand, in the last decade DE has been successfully applied to a wide range of complex real-world problems (see for example \cite{damak2009differential}, \cite{wang2010novel}, \cite{ali2012efficient}, \cite{tsai2013optimized}, \cite{yuan2013flexible}), but its performance in the FJS scheduling problem with sequencing flexibility hasn't been tested yet. Another reason that reinforces the choice of DE is that preliminary experiments involving other well-known metaheuristics such as artificial bee colony, particle swarm optimization, and grey wolf optimizer showed that DE achieves much better results than the other methods that were tested~\citep{lunardi2020}. Finally, ILS is considered due to its simplicity of implementation and usage. All metaheuristics are based on the same representation scheme (described in Section~\ref{repre}) and use the same definition of the neighborhood (described in Section~\ref{local}).

In the current section, we define $\vec{x} \in \mathbb{R}^{2\bar o}$ as the concatenation of a machine assignment $\tilde \pi$ and an execution order $\tilde \sigma$. This means that $\vec{x}_1,\dots,\vec{x}_{\bar o}$ correspond to $\tilde \pi_1,\dots,\tilde \pi_{\bar o}$; while $\vec{x}_{\bar o + 1},\dots,\vec{x}_{2\bar o}$ correspond to $\tilde \sigma_1,\dots,\tilde \sigma_{\bar o}$. Given $\vec{x}$ (and the instance constants $s_i$, $u_i$, $p_i$, $\bar c_i$, and $c_i$ for $i \in T$), it is easy to compute $\pi_i$, $\sigma_i$ ($i \in V \setminus T$), $\kappa_i$ ($i \in V$), and $\phi_k$ ($k=1,\dots,m$) as described in Sections~\ref{pisec} and~\ref{sigmasec}; and then the associated makespan~$C_{\max}$ using Algorithm~\ref{cmaxsec}.1. In this section, given $\vec{x}$, we denote $f(\vec{x})=C_{\max}$. Additionally, in the algorithms, the short terms ``chosen'', ``random'' or ``randomly chosen'' should be interpreted as abbreviations of ``randomly chosen with uniform distribution''.

Initial solutions of all methods are constructed in the same way. For each operation $i \in V \setminus T$, the machine $k \in F(i)$ with the lowest processing time is chosen. (For operations $i \in T$, the machine that processes operation~$i$ is fixed by definition.) Then, a cost-based breadth-first search (CBFS) algorithm is used to sequence the operations. The \textit{costs} of each operation are given by a random number in $[0,1]$. At each iteration of the CBFS, a set of eligible operations $\mathcal E$ is defined. Operations in $\mathcal E$ are those for which their immediate predecessors have already been sequenced. If $|\mathcal E| > 1$, operations in $\mathcal E$ are sequenced in increasing order of their costs; if $|\mathcal E| = 1$ then the single operation in $\mathcal E$ is sequenced. The procedure ends when $\mathcal E = \emptyset$ which implies that all operations have been sequenced. In the following subsections, we briefly and schematically describe the main principles of each metaheuristic. 

\subsection{Differential Evolution}

Proposed by \cite{storn1997differential} (see also~\cite{price2006differential} for further references), DE disturbs the current population members, unlike traditional evolutionary algorithms, with a scaled difference of indiscriminately preferred and dissimilar population members. In the basic variant of the DE, at each iteration, a mutant $\vec{v}^{\, i}$ is generated for each solution $\vec{x}^{\, i}$ ($i=1,2,\dots,n_{\size})$ according to 
\begin{equation} \label{de1}
\vec{v}^{\, i} = \vec{x}^{\, r_{1}} + \zeta (\vec{x}^{\, r_{2}} - \vec{x}^{\, r_{3}})
\end{equation}
where $\zeta$ is a parameter in $(0,2]$, usually less than or equal to 1, and $r_1, r_2, r_3 \in \{1, 2, \dots, n_{\size}\} \setminus \{i\}$ are random indices. Note that $n_{\size} \geq 4$ must be fulfilled, since $r_1, r_2, r_3$ and $i$ must be mutually different. The parameter $\zeta$ controls the amplifications of the differential variation. The basic DE variant with the mutation scheme given by~(\ref{de1}) is named DE/rand/1. The second most often used DE variant, denoted DE/best/1 (see~\cite{qin2008differential}), is also based on~(\ref{de1}) but
$r_1 = \argmin_{i=1,\dots,n_{\size}} \left\{ f(\vec{x}^{\, i}) \right\}$,
i.e., $\vec{x}^{\, r_1}$ is the individual with the best fitness value in the population and $r_2, r_3 \in \{1, 2, \dots, n_{\size}\} \setminus \{i, r_1 \}$ are random indices. Once the mutant $\vec{v}^{\, i}$ is generated, a trial $\vec{u}^{\, i}$ is formed as
\[ 
\vec{u}^{\, i}_j = \left\{ 
\begin{array}{ll}
\vec{v}^{\, i}_j & \mbox{if a random value in } [0,1] \mbox{ is less than or equal to } p_{\cro} \mbox{ or if } j = R(i),\\
\vec{x}^{\, i}_j & \mbox{otherwise},
\end{array} \right. 
\]
where $p_{\cro} \in [0,1]$ is a given parameter and $R(i)$ is a randomly chosen index in $\{1, 2,\dots,2\bar o\}$, which ensures that at least one element of $\vec{v}^{\, i}$ is passed to $\vec{u}^{\, i}$. To decide whether  $\vec{u}^{\, i}$ should become a member of the next generation or not, it is compared with $\vec{x}^{\, i}$ using a greedy criterion. If $f(\vec{u}^{\, i}) < f(\vec{x}^{\, i})$, then $\vec{u}^{\, i}$ substitutes $\vec{x}^{\, i}$; otherwise $\vec{x}^{\, i}$ is retained. Algorithm~\ref{pseudo:de} shows the essential steps of the proposed DE algorithm.
    
\begin{algorithm}[ht!]
{\small
\begin{algorithmic}[1]
\State \textbf{Input parameters:} $n_{\size}$, $\zeta$, $p_{\cro}$, $\mathrm{variant}$, and $t$.
\State $\mathcal P \gets \emptyset$.
\For{$i \gets 1$ to $n_{\size}$}
    \State Compute a random array of costs $c \in [0,1]^o$ and, using CBFS, construct an initial solution $\vec{x}^{\, i}$.
    \State Let $\mathcal P \gets \mathcal P \cup \{\vec{x}^{\, i}\}$.
\EndFor
\While{time limit $t$ not reached}
    \For{$i \gets 1$ to $n_{\size}$}
        \If{$\mathrm{variant} = $ DE/rand/1}
            \State Compute random numbers $r_1 \neq r_2 \neq r_3 \in \{1, 2,..., n_{\size}\} \setminus \{i\}$.
        \ElsIf{$\mathrm{variant} = $ DE/best/1}
            \State Let $r_1 \gets \argmin_{\ell=1,\dots,n_{\size}} \left\{ f(\vec{x}^{\, \ell}) \right\}$
            \State Compute random numbers $r_2 \neq r_3 \in \{1, 2,..., n_{\size}\} \setminus \{i, r_1\}$.
        \EndIf
        \State Compute $\vec{v} \gets \max\left\{0, \min\left\{\vec{x}^{\, r_1} + \zeta (\vec{x}^{\, r_2} - \vec{x}^{\, r_3}), 1 - 10^{-16}\right\}\right\}$.
        \State Compute a random number $R(i) \in \{1,\dots,2\bar o\}$.
        \For{$j \gets 1$ to $2\bar o$}
            \State Compute a random number $\gamma \in [0,1]$.
            \If{$\gamma \leq p_{\cro}$ or $j = R(i)$}
                \State $\vec{u}^{\, i}_j \gets \vec{v}^{\, i}_j$
            \Else
                \State $\vec{u}^{\, i}_j \gets \vec{x}^{\, i}_j$
            \EndIf
        \EndFor 
        \State Perform a local search starting from $\vec{u}^{\, i}$ to obtain $\vec{w}^{\, i}$ and compute $f(\vec{w}^{\, i})$. 
        \If{$f(\vec{w}^{\, i}) < f(\vec{x}^{\, i})$}
            $\mathcal P \gets \mathcal P \setminus \{ \vec{x}^{\, i} \} \cup \{ \vec{w}^{\, i}\}$.
        \EndIf
    \EndFor
\EndWhile
\State $\vec{x}^{\, \best} \gets \argmin_{\vec{x} \in \mathcal P} \left\{ f(\vec{x}) \right\}$.
\State \textbf{Return} $\vec{x}^{\, \best}$.
\end{algorithmic}}
\caption{Differential Evolution algorithm}\label{pseudo:de}
\end{algorithm}   
    
\subsection{Genetic Algorithm}

Initiated by~\cite{holland1992adaptation} (see~\cite{goldberg1988genetic} and~\cite{reeves2002genetic} for further references), GA is inspired by Charles Darwin's theory of evolution through natural selection. In the proposed GA, tournament selection is used to select the individuals (solutions) that are recombined (crossover) to generate the offspring. During tournament selection, two pairs of individuals are randomly chosen from the population and the fittest individual of each pair takes part of the recombination using uniform crossover. Preliminary experiments with uniform crossover, two-point crossover and simulated binary crossover (see~\cite{deb1995simulated}), showed that uniform crossover achieves the best results. Therefore, during uniform crossover of two solutions $\vec{x}^{\, i_1}$ and $\vec{x}^{\, i_2}$, two new solutions  $\vec{x}^{\, j_1}$ and $\vec{x}^{\, j_2}$ are generated as follows. For each $k \in \{1, 2,\dots,2\bar o\}$, with probability $\frac{1}{2}$, $\vec{x}^{\, j_1}_k \gets \vec{x}^{\, i_1}_k$ and $\vec{x}^{\, j_2}_k \gets \vec{x}^{\, i_2}_k$; otherwise, $\vec{x}^{\, j_1}_k \gets \vec{x}^{\, i_2}_k$ and $\vec{x}^{\, j_2}_k \gets \vec{x}^{\, i_1}_k$. Preliminary experiments with uniform mutation, Gaussian mutation and polynomial mutation (see~\cite{deb1999niched}, \cite{deb2014analysing}), showed that uniform mutation achieves the best results. Therefore, following uniform crossover, each offspring solution $\vec{x}$ is mutated with probability $p_{\mut} \in [0,1]$. During mutation, a random integer value $j \in \{1,2,\dots,2\bar o\}$ is chosen; and the value $\vec{x}_{j}$ is set to a random number in $[0,1)$. Once the new population is finally built, an elitist strategy is used. If the best individual $\vec{x}^{\, \best}_{\mathrm{new}}$ of the new population is less fit than the best individual $\vec{x}^{\, \best}$ of the current population, i.e., if $f(\vec{x}^{\, \best}_{\mathrm{new}}) > f(\vec{x}^{\, \best})$, then the worst individual of the new population is replaced with~$\vec{x}^{\, \best}$. Algorithm~\ref{pseudo:ga} shows the essential steps of the proposed GA. 
    
\begin{algorithm}[ht!]
{\small
\begin{algorithmic}[1]
\State \textbf{Input parameters:} $n_{\size}$, $p_{\mut}$, and $t$.
\State $\mathcal P \gets \emptyset$.
\For{$i \gets 1$ to $n_{\size}$}
    \State Compute a random array of costs $c \in [0,1]^o$ and, using CBFS, construct an initial solution $\vec{x}^{\, i}$.
    \State Let $\mathcal P \gets \mathcal P \cup \{\vec{x}^{\, i}\}$.
\EndFor
\While{time limit $t$ not reached}  
    \State Let $\mathcal Q \gets \emptyset$.
    \For{$1$ to $n_{\size}/2$}
        \State Compute random numbers $r_1 \neq r_2 \neq r_3 \neq r_4 \in \{1, 2, \dots,n_{\size}\}$ and 
        \State let $\vec{x}^{\, i_1} \gets \argmin \{ f(\vec{x}^{\, r_1}), f(\vec{x}^{\, r_2}) \} \; \mbox{ and } \; 
        \vec{x}^{\, i_2} \gets \argmin \{ f(\vec{x}^{\, r_3}), f(\vec{x}^{\, r_4}) \}$. 
        \For{$\ell \gets 1$ to $2\bar o$}
            \State Compute a random number $\gamma \in [0,1]$.
            \If{$\gamma \leq \frac{1}{2}$}
                $\vec{x}^{\, j_1}_{\ell} \gets \vec{x}^{\, i_1}_{\ell}$ and
                $\vec{x}^{\, j_2}_{\ell} \gets \vec{x}^{\, i_2}_{\ell}$
            \Else
                $\; \vec{x}^{\, j_1}_{\ell} \gets \vec{x}^{\, i_2}_{\ell}$ and
                $\vec{x}^{\, j_2}_{\ell} \gets \vec{x}^{\, i_1}_{\ell}$
            \EndIf            
        \EndFor
        \For{$j \in \{j_1, j_2\}$}
            \State Compute a random number $\gamma \in [0,1]$.
            \If{$\gamma \leq p_{\mut}$} 
                \State Compute random numbers $r \in \{1,2,\dots,2\bar o\}$ and $\xi \in [0,1)$ and let $\vec{x}^{\, j}_{r} \gets \xi$.
            \EndIf
            \State Perform a local search starting from $\vec{x}^{\, j}$ to generate $\vec{x}^{\, k}$.
            \State Let $\mathcal Q \gets \mathcal Q \cup \{\vec{x}^{\, k}\}$.
        \EndFor    
    \EndFor
    \State Let $\vec{x}^{\, \best} \gets \argmin_{\vec{x} \in \mathcal P} \left\{ f(\vec{x}) \right\}$ and $\vec{x}^{\, \best}_{\mathrm{new}} \gets \argmin_{\vec{x} \in \mathcal Q} \left\{ f(\vec{x}) \right\}$.
    \If{$f(\vec{x}^{\, \best}_{\mathrm{new}}) > f(\vec{x}^{\, \best})$}
        \State $\vec{x}^{\, \mathrm{worst}}_{\mathrm{new}} \gets \argmax_{\vec{x} \in \mathcal Q} \left\{ f(\vec{x}) \right\}$ and $\mathcal Q \gets \mathcal Q \setminus \{ \vec{x}^{\, \mathrm{worst}}_{\mathrm{new}} \} \cup \{ \vec{x}^{\, \best} \}$.
    \EndIf
    \State Let $\mathcal P \gets \mathcal Q$.
\EndWhile
\State $\vec{x}^{\, \best} \gets \argmin_{\vec{x} \in \mathcal P} \left\{ f(\vec{x}) \right\}$.
\State \textbf{Return} $\vec{x}^{\, \best}$.
\end{algorithmic}}
\caption{Genetic Algorithm}\label{pseudo:ga}
\end{algorithm} 
    
\subsection{Iterated Local Search} 

ILS is a simple trajectory-based metaheuristic (see~\cite{lourencco2003iterated}) that generates a sequence of local minimizers as follows. Starting from a given initial solution or a perturbed local minimizer, it runs a local search to find a new local minimizer. If the new local minimizer is better than the current local minimizer, then it is accepted as the new current local minimizer. Otherwise, the current local minimizer is preserved. The perturbation must be sufficiently strong to allow the local search to explore new search spaces, but also weak enough so that not all the good information gained in the previous search is lost. In the ILS algorithm we implemented, the perturbation of the current solution $\vec{x}$ is governed by a perturbation strength $\hat p \in \{1,2,\dots,2\bar o\}$ that determines how many randomly chosen positions of a local minimizer must be perturbed. The perturbation of a position simply consists in attributing a random value to it in $[0,1)$. Algorithm~\ref{pseudo:ils} shows the essential steps of the ILS algorithm.

\begin{algorithm}[ht!]
{\small
\begin{algorithmic}[1]
\State \textbf{Input parameters:} $\hat p$ and $t$.    
\State Compute a random array of costs $c \in [0,1]^o$ and, using CBFS, construct an initial solution $\vec{x}$.
\State Let $\vec{x}^{\, \mathrm{pert}} \gets \vec{x}$.
\While{time limit $t$ not reached}
    \State Perform a local search starting from $\vec{x}^{\, \mathrm{pert}}$ to obtain $\vec{v}$.
    \If{$f(\vec{v}) \leq f(\vec{x})$}
        \State $\vec{x} \gets \vec{v}$
    \EndIf
    \State Compute a set $\mathcal R \subseteq \{1, 2, \dots, 2\bar o\}$, with $|\mathcal R|=\hat p$, of mutually exclusive random numbers.
    \For{$i \gets 1$ to $2\bar o$}        
        \If{$i \in \mathcal R$}
            compute a random number $\gamma \in [0,1]$ and let $\vec{x}^{\, \mathrm{pert}}_i \gets \gamma$
        \Else
            $\;$ let $\vec{x}^{\, \mathrm{pert}}_i \gets \vec{x}_i$. 
        \EndIf
    \EndFor
\EndWhile
\State \textbf{Return} $\vec{x}$.
\end{algorithmic}}
\caption{Iterated local search}\label{pseudo:ils}
\end{algorithm}  
    
\subsection{Tabu Search}

Tabu Search was introduced in~\cite{glover1986future}. A description of the method and its main components can be found in~\cite{glover1997tabu}. TS is among the most used metaheuristics for combinatorial optimization problems. TS contrasts with memoryless design, which relies heavily on semi-random processes, guiding local choices with the information collected during the optimization process. The use of a list of recent actions (\textit{tabu list}) prevents the method from returning to recently visited solutions. When an action is performed, it is considered \textit{tabu} for the forthcoming~$T$ iterations, where $T$ is the tabu tenure. A solution is forbidden if it is obtained by applying a tabu action to the current solution. In the considered TS, an action is composed of a couple $(i,k)$, where~$i$ is an operation being moved and~$k$ is the machine to which~$i$ was assigned before the move. We keep track of the actions with a matrix $\uptau = (\uptau_{ik})$ with $i=1,\dots,\bar o$ and $k=1,\dots,m$. In this way, we set $\uptau_{ik} = \mathit{iter} + T$ whenever we perform action $(i,k)$ at iteration $\mathit{iter}$, i.e. $\uptau_{ik} = \mathit{iter} + T$ whenever we move from the current solution $\vec{x}$ to another solution $\vec{x}\,' \in N(\vec{x})$ by assigning to machine~$k'$ an operation~$i$ currently assigned to machine~$k$. An action $(i,k)$ is tabu if $\uptau_{ik} > \mathit{iter}$. The tabu tenure $T$ is crucial to the success of the tabu search procedure. We define $T = T(\lambda) = \lceil \lambda \log_e(\bar o)^2 \rceil$, where $\lambda$ is a parameter in $[0,2]$. During the search, the next solution is randomly chosen among the two neighbors with the smallest estimated makespan (see Section~\ref{estimatedmks}) that are non-tabu. Note that the neighborhood is defined as in the local search described in Section~\ref{neighborhood}. If all neighbors are tabu, a neighbor whose associated action $(i,k)$ has the smallest $\uptau_{ik}$ is chosen. With this procedure, the generated sequence does not possess the property of exhibiting a non-increasing makespan. Thus, the best-visited solution must be saved to be returned when a stopping criterion is satisfied. Moreover, preliminary experiments showed that the chance of producing cycles, created by the use of an estimated makespan, is increased by the use of an aspiration criterion (also based on an estimate of the neighbors' makespan). This is the reason why the TS considered in this work lacks an aspiration criterion. With some abuse of notation, we are saying ``a neighbor is tabu or not'' depending on whether the action that transforms the current solution into the neighbor is tabu or not. Specifically, assume we are at iteration $\mathit{iter}$ and let $\vec{x}$ be the current solution. Let ${\cal N}(\vec{x})$ be its neighborhood and let $\vec{y} \in {\cal N}(\vec{x})$ be a neighbor. Moreover, assume that in $\vec{x}$ there is an operation~$i$ assigned to machine~$k$ and that the action that transforms $\vec{x}$ into $\vec{y}$ includes to remove $i$ from $k$ and to assign it to another machine $k'$. We say $\vec{y}$ is a tabu neighbor of $\vec{x}$ if $(i,k)$ is tabu, i.e., if $\uptau_{ik} > \mathit{iter}$. Otherwise, we say $\vec{y}$ is a non-tabu neighbor. Algorithm~\ref{pseudo:ts} shows the essential steps of the considered TS algorithm.

\begin{algorithm}[ht!]
{\small
\begin{algorithmic}[1]
\State \textbf{Input parameters:} $T(\lambda)$ and $t$.    
\State $\text{iter} \gets 0$ and $\uptau_{ik} \gets 0$ ($i=1,\dots,\bar o$, $k=1,\dots,m$).
\State Compute a random array of costs $c \in [0,1]^o$ and, using CBFS, construct an initial solution $\vec{x}$.
\State Initialize $\vec{x}^{\, \best} \gets \vec{x}$.
\While{time limit $t$ not reached}
\State $\text{iter} \gets \text{iter} + 1$.
    \If{there are at least two non-tabu neighbors in ${\cal N}(\vec{x})$}
        \State Let $\vec{v}, \vec{w} \in {\cal N}(\vec{x})$ the two non-tabu neighbour solutions with smallest estimated makespan, 
        \State and let $\vec{y} \in \{\vec{v}, \vec{w}\}$ be randomly chosen. Let $(i,k)$ be the action that transforms $\vec{x}$ into $\vec{y}$. 
    \ElsIf{there is a single non-tabu neighbor in ${\cal N}(\vec{x})$}
        \State Let $\vec{y} \in {\cal N}(\vec{x})$ be the single non-tabu neighbour and let $(i,k)$ be the action that transforms 
        \State $\vec{x}$ into $\vec{y}$. 
    \Else
        \State Let $\vec{y} \in {\cal N}(\vec{x})$ be a (tabu) neighbour whose associated action $(i,k)$ has minimum $\uptau_{ik}$.
    \EndIf
    \State Let $\uptau_{ik} \gets \text{iter} + T(\lambda)$. 
    \State Let $\vec{x} \gets \vec{y}$.
    \If{$f(\vec{x}) < f(\vec{x}^{\, \best})$} $\vec{x}^{\, \best} \gets \vec{x}$\EndIf
\EndWhile
\State \textbf{Return} $\vec{x}^{\, \best}$.
\end{algorithmic}}
\caption{Tabu search}\label{pseudo:ts}
\end{algorithm}

\section{Experimental verification and analysis}\label{sec:exp}

In this section, extensive numerical experiments with the proposed metaheuristics for the OPS scheduling problem are presented. In a first set of experiments, parameters of the proposed metaheuristics are calibrated with a reduced set of OPS instances. In a second set of experiments, considering the whole set of OPS instances, the calibrated methods are compared to each other and against the IBM ILOG CP Optimizer (CPO) considered in~\cite{ops1}. As a result of the analysis of the performance of the proposed methods, a combined metaheuristic approach is introduced. In a last set of experiments, the best performing approach is evaluated when applied to the FJS with sequencing flexibility and the classical FJS scheduling problems considering well-known benchmark sets from the literature.

Metaheuristics were implemented in C++. Numerical experiments were conducted using a single physical core on an Intel Xeon E5-2680 v4 2.4 GHz with 4GB memory (per core) running CentOS Linux~7.7 (in 64-bit mode), at the High-Performance Computing (HPC) facilities of the University of Luxembourg~\citep{VBCG_HPCS14}.

\subsection{Sets of instances}

As a whole, 20 medium-sized and~100 large-sized instances of the OPS scheduling problem were considered. The set of medium-sized instances, named MOPS from now on, corresponds to the instances described in~\cite[\S5.2.2, Table~4]{ops1}. The set of large-sized instances corresponds to the set with~50 instances described in~\cite[\S5.2.3, Table~7]{ops1}, named LOPS1 from now on, plus a set with~50 additional even larger instances, named LOPS2 from now on, generated with the random instance generator described in~\cite[\S5.1]{ops1}. The instance generator relies on six integer parameters, namely, the number of jobs~$n$, the minimum~$o_{\min}$ and maximum~$o_{\max}$ number of operations \textit{per} job, the minimum~$m_{\min}$ and the maximum~$m_{\max}$ number of machines, and the maximum number~$q$ of periods of unavailability \textit{per} machine. The LOPS2 set contains~50 instances numbered from~51 to~100, the $k$-th instance being generated with the following parameters: $n = 11 + \lceil \frac{k}{100} \times 189 \rceil$, $o_{\min} = 5$, $o_{\max} = 6 + \lceil \frac{k}{100} \times 14 \rceil$, $m_{\min} = 9 + \lceil \frac{k}{100} \times 20 \rceil$, $m_{\max} = 10 + \lceil \frac{k}{100} \times 90 \rceil$, and $q = 8$. The instance generator and all considered instances are freely available at \url{https://github.com/willtl/online-printing-shop}. Table~\ref{tab:lops2characteristics} describes the main features of the 50 instances in the set LOPS2. The union of LOPS1 and LOPS2 will be named LOPS from now on. It is worth noticing that, although random, the OPS instances possess the characteristics of real-world instances of the OPS scheduling problem. Moreover, large-sized instances are of the size of the instances that occur in practice. 

In addition to the OPS instances, instances of the FJS scheduling problem with sequencing flexibility as proposed in~\cite{birgin2014milp} and instances of the FJS scheduling problem as proposed in~\cite{brandimarte1993routing}, \cite{hurink1994tabu}, \cite{barnes1996flexible}, and \cite{dauzere1997integrated} were considered. The instances in~\cite{birgin2014milp} are divided into two sets named YFJS and DAFJS. The first set corresponds to instances with ``Y-jobs'' while the second set corresponds to instances in which the jobs' precedence constraints are given by certain types of directed acyclic graphs (see~\cite{birgin2014milp} for details.) The sets of instances of the FJS scheduling problem were named BR, HK, BC, and DP, respectively. The HK set consists of the well-known EData, RData, and Vdata sets, with varying degrees of routing flexibility. 

Table~\ref{tab:instances} shows the main features of each instance set. The first two columns of the table (``Set name'' and ``\#inst.'') identify the set and the number of instances in each set. In the remaining columns, characteristics of the instances in each set are given. Column~$m$ refers to the number of machines, $\hat q$ refers to the number of periods of unavailability \textit{per} machine, $n$ is the number of jobs, $\hat o$ refers to the number of operations \textit{per} job, $|V|$ is the total number of operations (i.e., $|V|=o$), $|A|$ is the total number of precedence constraints, $|T|$ is the number of fixed operations, ``\#overlap'' is the number of operations whose processing may overlap with the processing of a successor (i.e., $|\{ i \in V \;|\; \theta_i < 1\}|$), and ``\#release'' is the number of operations with an \textit{actual} release time (i.e., $|\{ i \in V \;|\; r_i > 0\}|$). For each of these quantities, the table shows the minimum ($\min$), the average ($\mathrm{avg}$), and the maximum ($\max$), in the form $\min$\textbar$\mathrm{avg}$\textbar$\max$, over the whole considered set. It is worth noticing that, as a whole, 348 instances of different sources and nature are being considered.

\begin{table}[ht!]
\centering 
\caption{Main features of the fifty large-sized OPS instances in the LOPS2 set.}
\label{tab:lops2characteristics}
\resizebox{\textwidth}{!}{
\begin{tabular}{cccccccccccccccccc}
\toprule
& \multicolumn{6}{c}{Main instance characteristics} & \multicolumn{2}{c}{CP Optimizer formulation} &
& \multicolumn{6}{c}{Main instance characteristics} & \multicolumn{2}{c}{CP Optimizer formulation} \\
\midrule
\multirow{2}{*}{Instance} & 
\multirow{2}{*}{$m$} & 
\multirow{2}{*}{$\sum_{k=1}^m q_k$} & 
\multirow{2}{*}{$n$} & 
\multirow{2}{*}{$o$} & 
\multirow{2}{*}{$|A|$} & 
\multirow{2}{*}{$|T|$} &
\# integer & 
\multirow{2}{*}{\# constraints} &
\multirow{2}{*}{Instance} & 
\multirow{2}{*}{$m$} & 
\multirow{2}{*}{$\sum_{k=1}^m q_k$} & 
\multirow{2}{*}{$n$} & 
\multirow{2}{*}{$o$} & 
\multirow{2}{*}{$|A|$} & 
\multirow{2}{*}{$|T|$} &
\# integer & 
\multirow{2}{*}{\# constraints} \\
  &    &    &    &    &    &    & variables &   &
  &    &    &    &    &    &    & variables &   \\
\midrule  
 51 & 49 & 219 & 108 & 1067 & 1812 & 1 &  85148 &  249492 &   76 & 58 & 247 & 155 & 1757 & 3054 & 0 & 161749 &  475363 \\
 52 & 41 & 170 & 110 & 1076 & 1772 & 0 &  71169 &  209792 &   77 & 71 & 292 & 157 & 1793 & 3137 & 1 & 201915 &  594077 \\
 53 & 20 &  88 & 112 & 1105 & 1807 & 4 &  35763 &  105440 &   78 & 79 & 367 & 159 & 1789 & 3105 & 0 & 223365 &  655201 \\
 54 & 54 & 262 & 114 & 1137 & 1917 & 0 &  98745 &  291090 &   79 & 74 & 324 & 161 & 1798 & 3121 & 2 & 212354 &  626118 \\
 55 & 40 & 203 & 115 & 1065 & 1720 & 1 &  67611 &  199468 &   80 & 80 & 355 & 163 & 1850 & 3202 & 1 & 231248 &  680674 \\
 56 & 31 & 143 & 117 & 1098 & 1745 & 1 &  55292 &  162385 &   81 & 49 & 207 & 165 & 2080 & 3785 & 2 & 164905 &  485323 \\
 57 & 46 & 177 & 119 & 1217 & 2013 & 2 &  88912 &  261270 &   82 & 29 & 139 & 166 & 2063 & 3763 & 0 &  98767 &  291587 \\
 58 & 51 & 233 & 121 & 1274 & 2122 & 3 & 103766 &  304923 &   83 & 49 & 207 & 168 & 2044 & 3565 & 2 & 158986 &  468564 \\
 59 & 26 & 124 & 123 & 1271 & 2181 & 0 &  54918 &  162302 &   84 & 78 & 357 & 170 & 2082 & 3753 & 0 & 256059 &  753223 \\
 60 & 48 & 212 & 125 & 1346 & 2339 & 0 & 103373 &  304605 &   85 & 61 & 251 & 172 & 2047 & 3700 & 4 & 202088 &  593823 \\
 61 & 50 & 228 & 127 & 1358 & 2381 & 4 & 106864 &  314187 &   86 & 67 & 301 & 174 & 2133 & 3827 & 6 & 226909 &  666635 \\
 62 & 32 & 130 & 129 & 1290 & 2133 & 0 &  68060 &  199865 &   87 & 56 & 273 & 176 & 2215 & 4006 & 0 & 198539 &  584436 \\
 63 & 41 & 144 & 131 & 1370 & 2297 & 1 &  90142 &  265633 &   88 & 27 & 130 & 178 & 2141 & 3953 & 1 &  95622 &  282029 \\
 64 & 54 & 257 & 132 & 1421 & 2442 & 3 & 122801 &  361440 &   89 & 45 & 188 & 180 & 2299 & 4187 & 3 & 166199 &  488842 \\
 65 & 55 & 264 & 134 & 1427 & 2384 & 1 & 125843 &  370335 &   90 & 51 & 255 & 182 & 2213 & 4020 & 1 & 181658 &  534436 \\
 66 & 63 & 281 & 136 & 1523 & 2627 & 0 & 152642 &  449811 &   91 & 72 & 341 & 183 & 2340 & 4276 & 1 & 266059 &  782641 \\
 67 & 64 & 304 & 138 & 1499 & 2621 & 2 & 153859 &  452214 &   92 & 56 & 246 & 185 & 2400 & 4418 & 0 & 215525 &  634439 \\
 68 & 38 & 158 & 140 & 1579 & 2750 & 2 &  97716 &  288436 &   93 & 85 & 374 & 187 & 2399 & 4386 & 0 & 320129 &  941218 \\
 69 & 40 & 171 & 142 & 1577 & 2739 & 1 &  99887 &  294099 &   94 & 38 & 153 & 189 & 2447 & 4431 & 0 & 150904 &  444416 \\
 70 & 37 & 147 & 144 & 1588 & 2755 & 4 &  95755 &  281382 &   95 & 73 & 337 & 191 & 2568 & 4721 & 1 & 299347 &  879356 \\
 71 & 53 & 247 & 146 & 1590 & 2734 & 0 & 136147 &  400311 &   96 & 60 & 310 & 193 & 2508 & 4565 & 2 & 237394 &  698041 \\
 72 & 70 & 354 & 148 & 1701 & 2952 & 4 & 185238 &  544518 &   97 & 70 & 324 & 195 & 2443 & 4530 & 1 & 268046 &  788064 \\
 73 & 32 & 132 & 149 & 1778 & 3174 & 3 &  92225 &  271844 &   98 & 32 & 173 & 197 & 2579 & 4667 & 2 & 134682 &  397669 \\
 74 & 29 & 125 & 151 & 1726 & 3000 & 1 &  82365 &  242813 &   99 & 97 & 433 & 199 & 2548 & 4649 & 2 & 390037 & 1148630 \\
 75 & 33 & 167 & 153 & 1744 & 3077 & 0 &  94906 &  278965 &  100 & 58 & 247 & 200 & 2661 & 5032 & 3 & 246960 &  727868 \\
\bottomrule
\end{tabular}}
\end{table}

\begin{table}[ht!]
\centering 
\caption{Main features of the considered sets of instances.}
\label{tab:instances}
\resizebox{\textwidth}{!}{\begin{tabular}{ccccccccccccc}
\toprule
Set name & \#inst. & $m$ & $\hat q$ & $n$ & $\hat o$ & $|V|$ & $|A|$ & $|T|$ & \#overlap & \#release \\
\midrule  
MOPS & 20 &
6\textbar10\textbar17 & 
25\textbar48\textbar75 & 
5\textbar8\textbar10 & 
6\textbar9\textbar14 & 
36\textbar67\textbar109 & 
54\textbar106\textbar207 & 
0\textbar1\textbar3 & 
0\textbar7\textbar16 & 
0\textbar1\textbar6 \\ 
\rowcolor{gray!10}
LOPS & 100 &
10\textbar37\textbar97 & 
44\textbar168\textbar433 & 
13\textbar106\textbar200 & 
5\textbar10\textbar22 & 
79\textbar1153\textbar2661 & 
95\textbar1985\textbar5032 & 
0\textbar1\textbar6 & 
7\textbar115\textbar270 & 
0\textbar28\textbar79 \\
YFJS & 20 &
7\textbar14\textbar26 & 
0 & 4\textbar10\textbar17 & 
4\textbar10\textbar17 & 
24\textbar115\textbar289 & 
18\textbar105\textbar272 & 0 & 0 & 0 \\ 
\rowcolor{gray!10}
DAFJS & 30 &
5\textbar7\textbar10 & 
0 & 4\textbar7\textbar12 & 
4\textbar9\textbar23 & 
25\textbar71\textbar120 & 
23\textbar66\textbar117 & 0 & 0 & 0 \\ 
BR & 10 &
4\textbar8\textbar15 & 
0 & 10\textbar15\textbar20 & 
3\textbar9\textbar15 & 
55\textbar141\textbar240 & 
45\textbar125\textbar220 & 0 & 0 & 0 \\ 
\rowcolor{gray!10}
HK & 129 &
5\textbar8\textbar15 & 
0 & 6\textbar16\textbar30 & 
5\textbar8\textbar15 & 
36\textbar145\textbar300 & 
30\textbar128\textbar270 & 0 & 0 & 0 \\ 
BC & 21 &
11\textbar13\textbar18 & 
0 & 10\textbar13\textbar15 & 
10\textbar11\textbar15 & 
100\textbar158\textbar225 & 
90\textbar145\textbar210 & 0 & 0 & 0 \\ 
\rowcolor{gray!10}
DP & 18 &
5\textbar7\textbar10 & 
0 & 10\textbar15\textbar20 & 
15\textbar19\textbar25 & 
196\textbar292\textbar387 & 
186\textbar277\textbar367 & 0 & 0 & 0 \\ 
\bottomrule
\end{tabular}}
\end{table}

\subsection{Parameters tuning}\label{sec:param_tuning}

In this section, we aim to evaluate the performance of the proposed metaheuristics under variations of their parameters. Thirty OPS instances were used to fine-tune each parameter of each metaheuristic. The set of instances was composed of the five most difficult instances from the MOPS set according to the numerical results presented in~\cite[Table~5]{ops1} plus twenty-five representative instances from the LOPS set, namely, instances $1, 5, 9, 13, \dots, 97$. Since methods whose parameters are being calibrated have a random component, each method was applied to each instance ten times for each desired combination of parameters. For each run, a CPU time limit of 1200 seconds was imposed.

Assume that the combinations of parameters $c_1, c_2, \dots, c_A$ for method~$M$ applied to the set of instances $\{p_1,p_2,\dots,p_B\}$ should be evaluated. Let $f(M(c_{\alpha}),p_{\beta})$ be the average makespan over the ten runs of method~$M$ with the combination of parameters $c_{\alpha}$ applied to instance~$p_{\beta}$ for $\alpha=1,\dots,A$ and $\beta=1,\dots,B$. Let
\[
f_{\mathrm{best}}(M,p_{\beta}) = \min_{\{\alpha=1,\dots,A\}} \left\{ f(M(c_{\alpha}),p_{\beta}) \right\}, \mbox{ for } \beta=1,\dots,B,
\]
\[
f_{\mathrm{worst}}(M,p_{\beta}) = \max_{\{\alpha=1,\dots,A\}} \left\{ f(M(c_{\alpha}),p_{\beta}) \right\}, \mbox{ for } \beta=1,\dots,B,
\]
and
\[
\mathrm{RDI}(M(c_{\alpha}),p_{\beta}) = \frac{f(M(c_{\alpha}),p_{\beta}) - f_{\mathrm{best}}(M,p_{\beta})}{f_{\mathrm{worst}}(M,p_{\beta}) - f_{\mathrm{best}}(M,p_{\beta})},
\mbox{ for } \alpha=1,\dots,A \mbox{ and } \beta=1,\dots,B,
\]
where RDI stands for ``relative deviation index''. Thus, for every $\alpha$ and $\beta$, $\mathrm{RDI}(M(c_{\alpha}),p_{\beta}) \in [0,1]$ indicates the performance of method $M$ with the combination of parameters $c_{\alpha}$ applied to instance~$p_{\beta}$ with respect to the performance of the same method with other combinations of parameters. The smaller the $\mathrm{RDI}(M(c_{\alpha}),p_{\beta})$, the better the performance. In particular, $\mathrm{RDI}(M(c_{\alpha}),p_{\beta}) = 0$ if and only if $f(M(c_{\alpha}),p_{\beta}) = f_{\mathrm{best}}(M,p_{\beta})$ and $\mathrm{RDI}(M(c_{\alpha}),p_{\beta}) = 1$ if and only if $f(M(c_{\alpha}),p_{\beta}) = f_{\mathrm{worst}}(M,p_{\beta})$. If we now define
\[
\mathrm{RDI}(M(c_{\alpha})) = \frac{1}{|B|} \sum_{\beta=1}^B \mathrm{RDI}(M(c_{\alpha}),p_{\beta}),
\mbox{ for } \alpha=1,\dots,A,
\]
then we can say that the combination of parameters~$c_{\alpha}$ with the smallest $\mathrm{RDI}(M(c_{\alpha}))$ is the one for which method~$M$ performed best.

\subsubsection{Differential Evolution}

\begin{figure}[ht!]
\centering 
\begin{tabular}{ccc}
\includegraphics{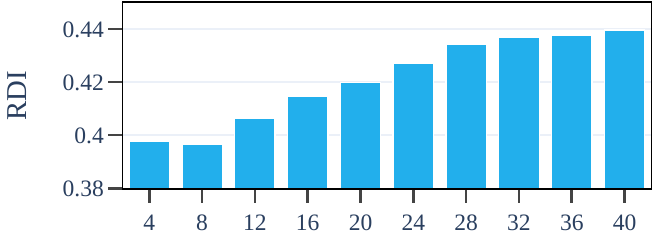} & \quad & \includegraphics{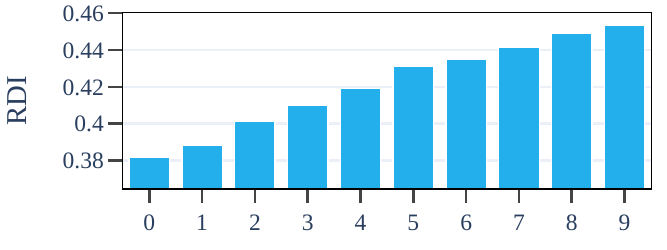} \\
(a) $n_{\size}$  & & (b) $p_{\cro} \times 10^3$ \\[2mm]
\includegraphics{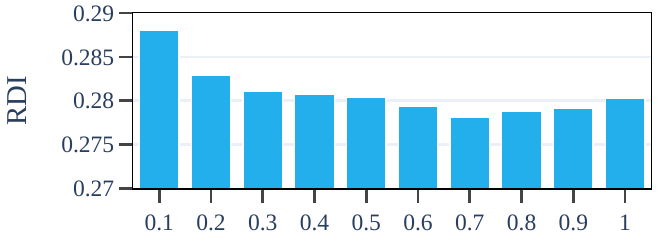} & \quad & \includegraphics{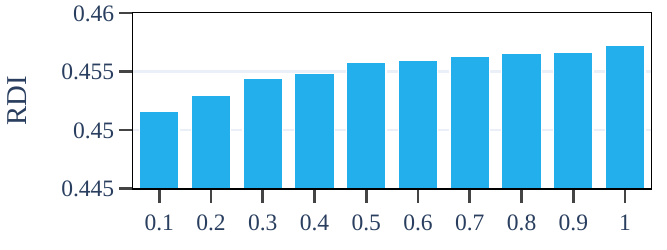} \\
(c) $\zeta$ (five MOPS instances)  & & (d) $\zeta$ (twenty five LOPS instances)
\end{tabular}
\caption{DE performance for different parameters' settings.}
\label{param_de}
\end{figure}

\begin{figure}[ht!]
\centering 
\begin{tabular}{ccc}
\includegraphics{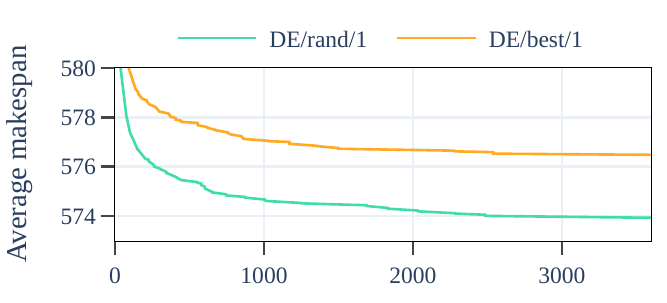} & \quad & \includegraphics{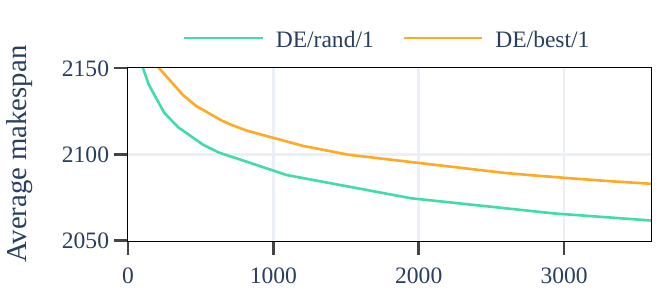} \\
(a) five MOPS instances  & & (b) twenty five LOPS instances \\ 
\end{tabular}
\caption{Evolution of the average makespan as a function of time obtained with DE (a) with $n_{\size}=8$, $p_{\cro}=0$, and $\zeta=0.7$ applied to the five selected instances from the MOPS set and (b) with $n_{\size}=8$, $p_{\cro}=0$, and $\zeta=0.1$ applied to the twenty five selected instances from the LOPS set.}
\label{fig:de_scheme}
\end{figure}

In DE there are four parameters to be calibrated, namely, $n_{\size}$, $p_{\cro}$, $\zeta$, and $\mathit{variant}$. Preliminary experiments indicated that varying these parameters within the ranges $n_{\size} \in [4,40]$, $p_{\cro} \in [0,0.01]$, $\zeta \in [0,1]$, and $\mathit{variant} \in \{ \mathrm{DE/rand/1}, \mathrm{DE/best/1} \}$ would provide acceptable results. Since testing all combinations in a grid would be very time consuming, we arbitrarily proceeded as follows. We first varied $n_{\size} \in \{ 4, 8, 12,\dots,40\}$ with  $p_{\cro} = 0.005$, $\zeta = 0.5$, and $\mathrm{variant}=\mathrm{DE/rand/1}$. Figure~\ref{param_de}a shows the RDI for the different values of $n_{\size}$. The figure shows that the method achieved its best performance at $n_{\size}=8$. In a second experiment, we fixed $n_{\size}=8$, $\zeta = 0.5$, $\mathit{variant}=\mathrm{DE/rand/1}$, and varied $p_{\cro} \in \{0, 10^{-3}, 2 \times 10^{-3}, \dots, 9 \times 10^{-3} \}$. Figure~\ref{param_de}b shows that the best performance was obtained with $p_{\cro}=0$. In a third experiment, we set $n_{\size}=8$, $p_{\cro}=0$, $\mathit{variant}=\mathrm{DE/rand/1}$, and varied $\zeta \in \{0.1, 0.2, \dots, 1 \}$. Figures~\ref{param_de}c and~\ref{param_de}d show the results for the five problems in the MOPS set and the twenty five instances in the LOPS set, respectively. The results demonstrate that the best performance is obtained for $\zeta=0.7$ and $\zeta=0.1$, respectively. It is worth noticing that the performance of the method varies smoothly as a function of its parameters as indicated by Figures~\ref{param_de}a--\ref{param_de}d. Finally, Figures~\ref{fig:de_scheme}a and~\ref{fig:de_scheme}b show the performance of the algorithm with $n_{\size}=8$, $p_{\cro}=0$, and $\zeta=0.7$ applied to the five instances from the MOPS set and with $n_{\size}=8$, $p_{\cro}=0$, and $\zeta=0.1$ applied to the twenty five instances from the LOPS set. In both cases, the figures compare the performance for variations of $\mathit{variant} \in \{ \mathrm{DE/rand/1}, \mathrm{DE/best/1} \}$. The considered mutation variants are the two most widely adopted ones in the literature. The main difference between both of them is that the former emphasizes exploration while the latter emphasizes exploitation. In this experiment, the time limit was extended to 1 hour. Figures~\ref{fig:de_scheme}a and~\ref{fig:de_scheme}b show the average makespan over the considered subsets of instances as a function of time. Both graphics show that a choice of $\mathit{variant} = \mathrm{DE/rand/1}$ is more efficient. 

\subsubsection{Genetic Algorithm}

\begin{figure}[ht!]
\centering
\resizebox{\textwidth}{!}{
\begin{tabular}{ccc}
\includegraphics{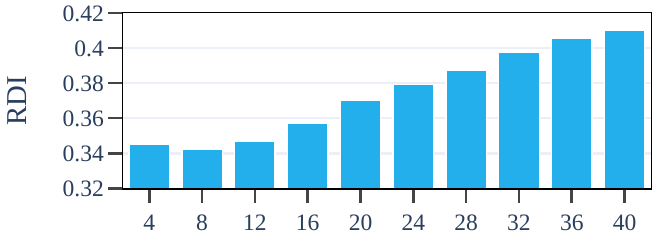} & 
\includegraphics{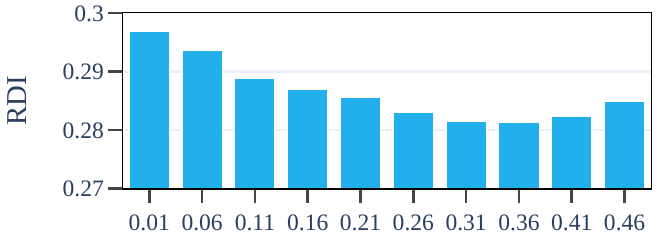} &
\includegraphics{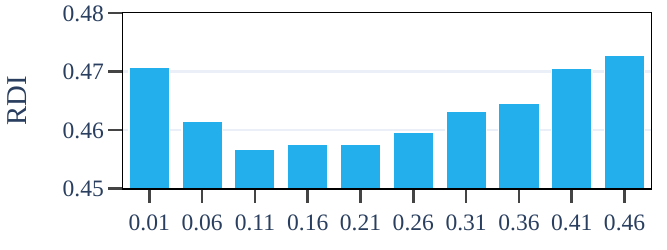} \\
(a) $n_{\size}$ &
(b) $p_{\mut}$ (five MOPS instances) &
(c) $p_{\mut}$ (twenty five LOPS instances) 
\end{tabular}}
\caption{Genetic Algorithm performance for different parameters' settings.}
\label{fig:param_ga}
\end{figure}

In GA there are two parameters to be calibrated, namely, $n_{\size}$ and $p_{\mut}$. Preliminary experiments indicated that varying these parameters within the ranges $n_{\size} \in [4,40]$ and $p_{\mut} \in [0.01,0.5]$ would provide acceptable results. In a first experiment, we varied $n_{\size} \in \{ 4, 8, \dots, 40\}$ with $p_{\mut}=0.25$. Figure~\ref{fig:param_ga}a shows that the best performance is obtained with $n_{\size} = 8$. In a second experiment, we fixed $n_{\size} = 8$ and varied $p_{\mut} \in \{0.01, 0.06,\dots,0.46\}$. Figures~\ref{fig:param_ga}b and~\ref{fig:param_ga}c show that the best performance is obtained with $p_{\mut} = 0.36$ when the method is applied to the five selected instances from the MOPS set; while its best performance is obtained with $p_{\mut} = 0.11$ when applied to the twenty five selected instances from the LOPS set. It can be observed that, as is happened with DE, the best population size is $n_{\size} = 8$ and it does not depend on the size of the instances. On the other hand, the same behavior is not observed for the mutation probability parameter $p_{\mut}$. Similar to the parameter $\zeta$ of DE that appears in its mutation scheme, a different behavior is observed when the method is applied to instances from the MOPS and the LOPS sets. At this point, it is important to stress that this should not be considered problematic. The goal of the present work is to develop an efficient and effective method to be applied to practical instances of the OPS scheduling problem, i.e., to a real-world problem; and these instances are very similar to the instances in the LOPS set. Numerical experimentation with the MOPS instances is carried out for assessment purposes, comparing the obtained results with the ones presented in~\cite{ops1}, which include numerical experiments with instances of the MOPS set.


\subsubsection{Iterated Local Search and Tabu Search}

\begin{figure}[ht!]
\centering
\begin{tabular}{cc}
\includegraphics{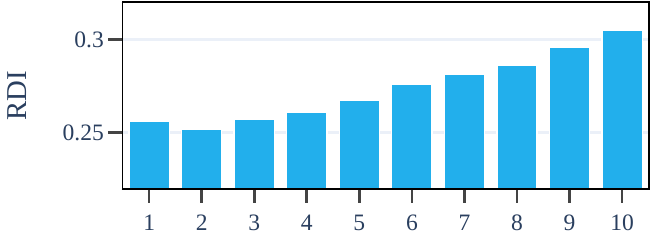} &
\includegraphics{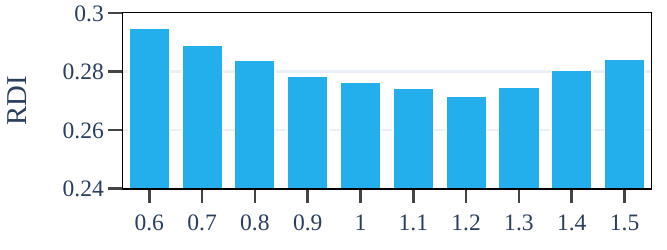} \\
(a) & (b)
\end{tabular}
\caption{Performances of (a) Iterated Local Search and (b) Tabu Search as a function of their parameters $\hat p$ and $\lambda$.}
\label{params_ilsts}
\end{figure}

ILS and TS have a single parameter to calibrate, namely $\hat p$ and $\lambda$, respectively. Preliminary experiments indicated that varying these parameters within the ranges $\hat p \in [1,10]$ and $\lambda \in [0.6,1.5]$ would provide acceptable results. Figures~\ref{params_ilsts}(a-b) show the results varying $\hat p \in \{1, 2, \dots, 10 \}$ and $\lambda \in \{0.6, 0.7,\dots,1.5\}$, respectively. They show that ILS performed best with $\hat p = 2$; while TS obtained the best results with $\lambda = 1.2$. It is worth noticing that, in both cases, the performance varies smoothly as a function of the parameters; thus similar performances are obtained for small variations of the parameters.

\subsection{Experiments with OPS instances}

This section presents numerical experiments with the four calibrated metaheuristics DE, GA, ILS, and TS. In addition, the performance of the IBM ILOG CP Optimizer (CPO)~\citep{Laborie2018}, version 12.9, is presented. CPO is a ``half-heuristic-half-exact'' solver specially designed to tackle scheduling problems. It has its own constraint programming (CP) modeling language to fully explore the structure of the underlying problem. In the experiments, the two-phase strategy ``Incomplete model + CP Model~4'' described in~\cite{ops1} is considered. This approach consists in first solving a simplified model and, in a second phase, using the solution obtained in the first phase as the initial solution to the full and more complex model. This is the approach that performed best among several alternative CP models and solution strategies considered in~\cite{ops1}. 

Numerical experiments consider the 20 instances in the MOPS set and the 100 instances in the LOPS set. Each metaheuristic was run 50 times in each instance of the MOPS set and 12 in each instance of the LOPS set. As described in Section~\ref{sec:param_tuning}, the \textit{average} over all runs is considered for comparison purposes. For each run, a CPU time limit of 2 hours was imposed. The metaheuristics being evaluated start from a feasible solution and generate a sequence of feasible solutions. Thus, it is possible to observe the evolution of the makespan over time. This is not the case of the strategy of the CPO being considered. In the two-phase strategy, 2/3 of the time budget is allocated to the solution of a relaxed or incomplete OPS formulation in which setup operations can be preempted and the setup of the first operation to be processed in each machine is considered to be null; while the remaining 1/3 of the time budget is allocated to the solution of the actual CP formulation of the OPS scheduling problem. Due to the two-phase strategy, it is not possible to track the evolution of the makespan over time, since in the first 2/3 of the time budget the incumbent solution is, with high probability, infeasible. Therefore, to compare the performance of the proposed methods against the CPO, CPO was run several times with increasing time budgets given by 5 minutes, 30 minutes, and 2 hours per instance.

\begin{figure}[ht!]
\centering 
\includegraphics{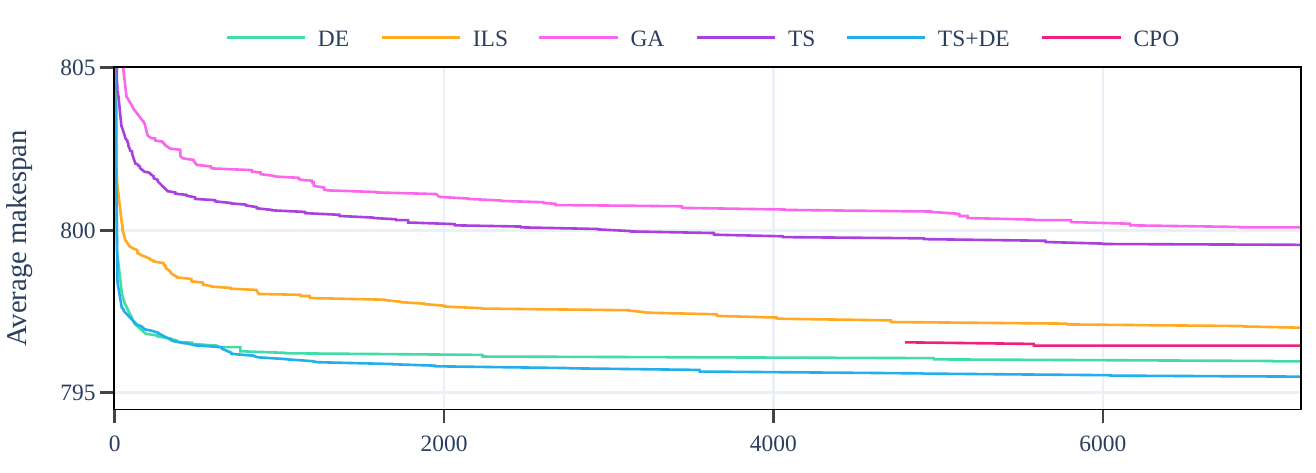}
\caption{Evolution of the average makespan over time of each proposed method and CPO applied to the MOPS set instances.} 
\label{fig:exp_mops}
\end{figure}

\begin{figure}[ht!]
\centering 
\includegraphics{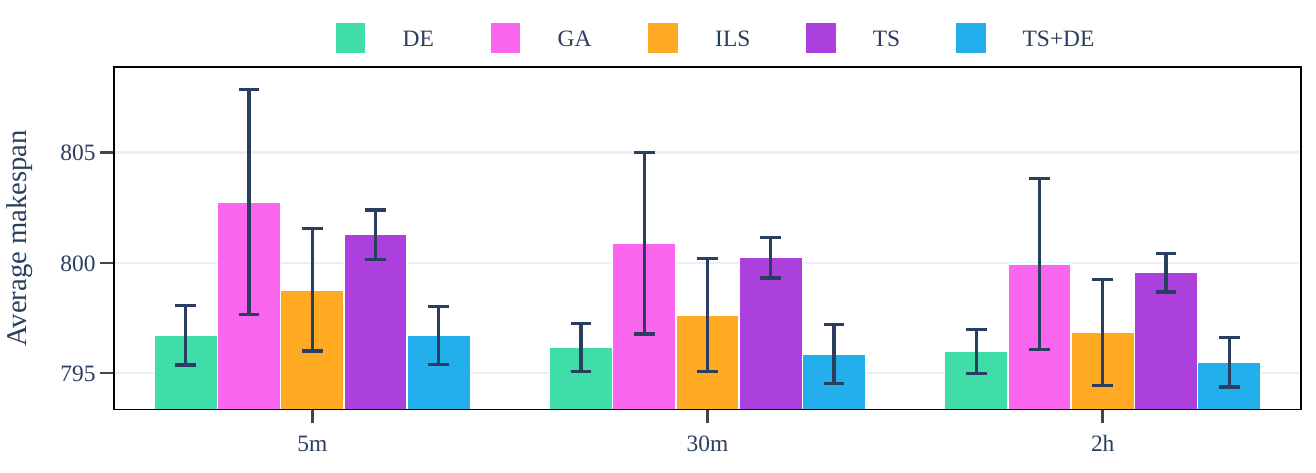}
\caption{Average makespans and pooled standard deviations that result from applying the proposed metaheuristic approaches fifty times to instances in the MOPS set with CPU time limits of 5 minutes, 30 minutes, and 2 hours.} 
\label{fig:pooled_std_dev_mops} 
\end{figure}

Figure~\ref{fig:exp_mops} shows the evolution of the average makespan (over the 50 runs and over all instances) when the five methods are applied to the instances in the MOPS set. Table~\ref{tab:exp_mops} presents the \textit{best} makespan (in the top half of the table) and the \textit{average} makespan (in the bottom half of the table) obtained by each metaheuristic method in each instance. The last line in each half of the table presents the average results. (Average of the best results in the first half and average of the average results in the second half.) In the second-half of the table, in which average results are being presented, an additional line exhibits the pooled standard deviation. For each instance, figures in bold represent the best result obtained by the methods under consideration. Average makespans and pooled standard deviations are graphically represented in Figure~\ref{fig:pooled_std_dev_mops}. Method TS+DE that appears in the figures and the table should be ignored at this time. The motivation for its definition as well as its presence in the experiments will be elucidated later in the current section. Table~\ref{tab:exp_mopsOPS} shows the results of applying CPO to instances in the MOPS set. In the table, ``UB'' corresponds to the best solution found (upper bound to the optimal solution); while ``LB'' corresponds to the computed lower bound when the CPU time limit is equal to two hours. A comparison between the lower and the upper bound shows that the optimal solution was found for instances 1--5, 7, 9, 10, 13, and 15--19; while a non-null gap is reported for instances 6, 8, 11, 12, 14, and 20. 

\begin{table}[ht!]
\centering
\caption{Results of applying the metaheuristic approaches to instances in the MOPS set.}
\label{tab:exp_mops}
\resizebox{\textwidth}{!}{\begin{tabular}{cccccc|ccccc|ccccc}
\toprule
\multirow{2}{*}{} & \multicolumn{15}{c}{Best makespan} \\
\cmidrule(lr){2-16}
\multirow{2}{*}{} & \multicolumn{5}{c}{CPU time limit: 5 minutes} & \multicolumn{5}{c}{CPU time limit: 30 minutes} & \multicolumn{5}{c}{CPU time limit: 2 hours} \\
\cmidrule(lr){2-16}
 & DE & GA & ILS & TS & TS+DE & DE & GA & ILS & TS & TS+DE & DE & GA & ILS & TS & TS+DE \\
\midrule
 1 & \textbf{344} & 351 & 346 & \textbf{344} & \textbf{344} & \textbf{344} & 350 & 346 & \textbf{344} & \textbf{344} & \textbf{344} & 350 & 346 & \textbf{344} & \textbf{344} \\  
2 & \textbf{357} & 358 & 358 & \textbf{357} & \textbf{357} & \textbf{357} & 358 & \textbf{357} & \textbf{357} & \textbf{357} & \textbf{357} & \textbf{357} & \textbf{357} & \textbf{357} & \textbf{357} \\  
3 & \textbf{405} & 409 & 407 & 409 & \textbf{405} & 405 & 409 & 407 & 409 & \textbf{404} & 405 & 409 & 405 & 408 & \textbf{404} \\  
4 & \textbf{458} & \textbf{458} & \textbf{458} & \textbf{458} & \textbf{458} & \textbf{458} & \textbf{458} & \textbf{458} & \textbf{458} & \textbf{458} & \textbf{458} & \textbf{458} & \textbf{458} & \textbf{458} & \textbf{458} \\  
5 & \textbf{507} & 516 & 510 & \textbf{507} & \textbf{507} & \textbf{507} & 516 & 510 & \textbf{507} & \textbf{507} & \textbf{507} & 516 & 509 & \textbf{507} & \textbf{507} \\  
6 & 435 & 447 & 436 & 437 & \textbf{432} & 435 & 446 & 436 & 434 & \textbf{432} & 435 & 442 & 436 & 433 & \textbf{432} \\  
7 & \textbf{2429} & \textbf{2429} & \textbf{2429} & \textbf{2429} & \textbf{2429} & \textbf{2429} & \textbf{2429} & \textbf{2429} & \textbf{2429} & \textbf{2429} & \textbf{2429} & \textbf{2429} & \textbf{2429} & \textbf{2429} & \textbf{2429} \\  
8 & \textbf{447} & 459 & 453 & 461 & 448 & \textbf{446} & 451 & 453 & 456 & 447 & \textbf{445} & 451 & 451 & 456 & 447 \\  
9 & \textbf{629} & 632 & 630 & 633 & \textbf{629} & \textbf{629} & 631 & 630 & 631 & \textbf{629} & \textbf{629} & 631 & \textbf{629} & 630 & \textbf{629} \\  
10 & \textbf{1184} & \textbf{1184} & \textbf{1184} & \textbf{1184} & \textbf{1184} & \textbf{1184} & \textbf{1184} & \textbf{1184} & \textbf{1184} & \textbf{1184} & \textbf{1184} & \textbf{1184} & \textbf{1184} & \textbf{1184} & \textbf{1184} \\  
11 & \textbf{413} & 427 & 419 & 433 & 414 & \textbf{413} & 426 & 414 & 433 & \textbf{413} & \textbf{413} & 423 & 414 & 430 & \textbf{413} \\  
12 & \textbf{491} & 500 & 496 & 511 & 492 & \textbf{489} & 492 & 492 & 511 & \textbf{489} & \textbf{489} & 492 & 492 & 507 & \textbf{489} \\  
13 & \textbf{347} & \textbf{347} & \textbf{347} & \textbf{347} & \textbf{347} & \textbf{347} & \textbf{347} & \textbf{347} & \textbf{347} & \textbf{347} & \textbf{347} & \textbf{347} & \textbf{347} & \textbf{347} & \textbf{347} \\  
14 & 392 & 404 & 396 & 412 & \textbf{389} & 391 & 404 & 393 & 408 & \textbf{389} & \textbf{389} & 400 & 391 & 408 & \textbf{389} \\  
15 & 320 & 320 & \textbf{319} & \textbf{319} & \textbf{319} & 320 & \textbf{319} & \textbf{319} & \textbf{319} & \textbf{319} & 320 & \textbf{319} & \textbf{319} & \textbf{319} & \textbf{319} \\  
16 & \textbf{543} & \textbf{543} & \textbf{543} & \textbf{543} & \textbf{543} & \textbf{543} & \textbf{543} & \textbf{543} & \textbf{543} & \textbf{543} & \textbf{543} & \textbf{543} & \textbf{543} & \textbf{543} & \textbf{543} \\  
17 & \textbf{1052} & \textbf{1052} & \textbf{1052} & \textbf{1052} & \textbf{1052} & \textbf{1052} & \textbf{1052} & \textbf{1052} & \textbf{1052} & \textbf{1052} & \textbf{1052} & \textbf{1052} & \textbf{1052} & \textbf{1052} & \textbf{1052} \\  
18 & \textbf{3184} & \textbf{3184} & \textbf{3184} & \textbf{3184} & \textbf{3184} & \textbf{3184} & \textbf{3184} & \textbf{3184} & \textbf{3184} & \textbf{3184} & \textbf{3184} & \textbf{3184} & \textbf{3184} & \textbf{3184} & \textbf{3184} \\  
19 & \textbf{1451} & \textbf{1451} & \textbf{1451} & \textbf{1451} & \textbf{1451} & \textbf{1451} & \textbf{1451} & \textbf{1451} & \textbf{1451} & \textbf{1451} & \textbf{1451} & \textbf{1451} & \textbf{1451} & \textbf{1451} & \textbf{1451} \\  
20 & \textbf{507} & 519 & 521 & 538 & 511 & \textbf{507} & 518 & 514 & 534 & \textbf{507} & \textbf{507} & 514 & 514 & 534 & \textbf{507} \\  
\midrule
 & \textbf{794.75} & 799.5 & 796.95 & 800.45 & \textbf{794.75} & 794.55 & 798.4 & 795.95 & 799.55 & \textbf{794.25} & 794.4 & 797.6 & 795.55 & 799.05 & \textbf{794.25} \\ 
\midrule
\multirow{2}{*}{} & \multicolumn{15}{c}{Average makespan} \\
\midrule
1 & 346.25 & 361.2 & 349.2 & \textbf{344} & 344.12 & 346 & 361 & 348 & \textbf{344} & 344.12 & 346 & 360.8 & 347.6 & \textbf{344} & 344.12 \\  
2 & 357.75 & 361.6 & 361.2 & \textbf{357.25} & 357.88 & 357.75 & 360.8 & 359.2 & \textbf{357} & 357.88 & 357.75 & 359 & 357.4 & \textbf{357} & 357.88 \\  
3 & 408.25 & 417.4 & 408.4 & 409.5 & \textbf{407.62} & 407 & 416 & 408.4 & 409 & \textbf{406.5} & 406.25 & 414.8 & 407.6 & 408.25 & \textbf{406.12} \\  
4 & \textbf{458} & 461.6 & \textbf{458} & \textbf{458} & \textbf{458} & \textbf{458} & 460 & \textbf{458} & \textbf{458} & \textbf{458} & \textbf{458} & 459 & \textbf{458} & \textbf{458} & \textbf{458} \\  
5 & 511 & 521.2 & 511.8 & 510 & \textbf{509.12} & 509.5 & 518.6 & 511.6 & \textbf{508} & 508.5 & \textbf{508} & 518.6 & 510.2 & \textbf{508} & 508.5 \\  
6 & 436.5 & 457.2 & 441.6 & 438 & \textbf{436.12} & 436.25 & 449 & 441 & \textbf{435.5} & 435.62 & 436.25 & 447.8 & 441 & \textbf{433.75} & 435.62 \\  
7 & \textbf{2429} & \textbf{2429} & \textbf{2429} & \textbf{2429} & \textbf{2429} & \textbf{2429} & \textbf{2429} & \textbf{2429} & \textbf{2429} & \textbf{2429} & \textbf{2429} & \textbf{2429} & \textbf{2429} & \textbf{2429} & \textbf{2429} \\  
8 & \textbf{451.25} & 463 & 459.2 & 462 & 452.5 & \textbf{450.5} & 461 & 456.2 & 458.75 & 451.12 & \textbf{450} & 460.6 & 453.6 & 456.5 & 450.62 \\  
9 & \textbf{630.5} & 638 & 630.8 & 637 & \textbf{630.5} & 630 & 632.6 & 630 & 633.5 & \textbf{629.88} & 629.75 & 631.8 & 629.6 & 632.25 & \textbf{629.5} \\  
10 & \textbf{1184} & \textbf{1184} & \textbf{1184} & \textbf{1184} & \textbf{1184} & \textbf{1184} & \textbf{1184} & \textbf{1184} & \textbf{1184} & \textbf{1184} & \textbf{1184} & \textbf{1184} & \textbf{1184} & \textbf{1184} & \textbf{1184} \\  
11 & 421.5 & 428.4 & 421.6 & 434.25 & \textbf{419.62} & 420 & 426.2 & 416.8 & 433.5 & \textbf{416.5} & 420 & 423.6 & 416.6 & 430.75 & \textbf{416} \\  
12 & \textbf{495} & 504.2 & 501.2 & 512 & 496.75 & 494.75 & 497.6 & 497.8 & 511.25 & \textbf{493.5} & 494 & 494.6 & 495.8 & 508.5 & \textbf{493} \\  
13 & \textbf{347} & \textbf{347} & \textbf{347} & \textbf{347} & \textbf{347} & \textbf{347} & \textbf{347} & \textbf{347} & \textbf{347} & \textbf{347} & \textbf{347} & \textbf{347} & \textbf{347} & \textbf{347} & \textbf{347} \\  
14 & 396.5 & 408.2 & 401.4 & 414.25 & \textbf{395.5} & 394.25 & 404.4 & 397.6 & 410.5 & \textbf{393.88} & 394 & 400.8 & 394.8 & 409 & \textbf{393.12} \\  
15 & 320 & 320 & 319.6 & \textbf{319} & 319.5 & 320 & 319.6 & \textbf{319} & \textbf{319} & 319.5 & 320 & 319.2 & \textbf{319} & \textbf{319} & 319.5 \\  
16 & \textbf{543} & \textbf{543} & \textbf{543} & \textbf{543} & \textbf{543} & \textbf{543} & \textbf{543} & \textbf{543} & \textbf{543} & \textbf{543} & \textbf{543} & \textbf{543} & \textbf{543} & \textbf{543} & \textbf{543} \\  
17 & \textbf{1052} & \textbf{1052} & \textbf{1052} & \textbf{1052} & \textbf{1052} & \textbf{1052} & \textbf{1052} & \textbf{1052} & \textbf{1052} & \textbf{1052} & \textbf{1052} & \textbf{1052} & \textbf{1052} & \textbf{1052} & \textbf{1052} \\  
18 & \textbf{3184} & \textbf{3184} & \textbf{3184} & \textbf{3184} & \textbf{3184} & \textbf{3184} & \textbf{3184} & \textbf{3184} & \textbf{3184} & \textbf{3184} & \textbf{3184} & \textbf{3184} & \textbf{3184} & \textbf{3184} & \textbf{3184} \\  
19 & \textbf{1451} & \textbf{1451} & \textbf{1451} & \textbf{1451} & \textbf{1451} & \textbf{1451} & \textbf{1451} & \textbf{1451} & \textbf{1451} & \textbf{1451} & \textbf{1451} & \textbf{1451} & \textbf{1451} & \textbf{1451} & \textbf{1451} \\  
20 & \textbf{511.75} & 523 & 521.4 & 540.25 & 516.75 & \textbf{509.25} & 520.8 & 519 & 536.75 & 512 & 509.25 & 518.2 & 515.4 & 536 & \textbf{507.88} \\  
\midrule
Avg. 1--20 &  796.71 & 802.75 & 798.77 & 801.27 & \textbf{796.7} & 796.16 & 800.88 & 797.63 & 800.24 & \textbf{795.85} & 795.96 & 799.94 & 796.83 & 799.55 & \textbf{795.49} \\ 
Pooled SD & 1.35 & 5.10 & 2.77 & 1.12 & 1.31 & 1.09 & 4.11 & 2.56 & 0.92 & 1.34 & 1.00 & 3.88 & 2.41 & 0.87 & 1.13 \\
\bottomrule
\end{tabular}}
\end{table}

\begin{table}[!ht]
\centering
\caption{Results of applying CPO to instances in the MOPS set.}
\label{tab:exp_mopsOPS}
\resizebox{\textwidth}{!}{\begin{tabular}{cccccccccccccccccccc}
\toprule
\multirow{2}{*}{Inst.} & 5 min. & 30 min. & \multicolumn{2}{c}{2 hours} & \multirow{2}{*}{Inst.} & 5 min. & 30 min. & \multicolumn{2}{c}{2 hours} & \multirow{2}{*}{Inst.} & 5 min. & 30 min. & \multicolumn{2}{c}{2 hours} & \multirow{2}{*}{Inst.} & 5 min. & 30 min. & \multicolumn{2}{c}{2 hours} \\
\cmidrule(lr){2-5}
\cmidrule(lr){7-10}
\cmidrule(lr){12-15}
\cmidrule(lr){17-20}
  & UB & UB & LB & UB & & UB & UB & LB & UB &  & UB & UB & LB & UB & & UB & UB & LB & UB \\ 
\midrule
1 & 344 & 344 & 344 & 344 & 6 & 441 & 441 & 335 & 441 & 11 & 418 & 418 & 406 & 418 & 16 & 543 & 543 & 543 & 543 \\ 
2 & 357 & 357 & 357 & 357 & 7 & 2429 & 2429 & 2429 & 2429 & 12 & 506 & 497 & 457 & 499 & 17 & 1080 & 1052 & 1052 & 1052 \\ 
3 & 404 & 404 & 404 & 404 & 8 & 456 & 450 & 360 & 450 & 13 & 347 & 347 & 347 & 347 & 18 & 3184 & 3184 & 3184 & 3184 \\ 
4 & 458 & 458 & 458 & 458 & 9 & 632 & 629 & 629 & 629 & 14 & 402 & 402 & 320 & 394 & 19 & 1451 & 1451 & 1451 & 1451 \\ 
5 & 506 & 506 & 506 & 506 & 10 & 1184 & 1184 & 1184 & 1184 & 15 & 319 & 319 & 319 & 319 & 20 & 522 & 522 & 417 & 520 \\  
\midrule
&&&&&&&&&&&&&&& Avg. 1--20 & 799.1 & 796.9 & & 796.5 \\
\bottomrule
\end{tabular}}
\end{table}

The results presented in Figure~\ref{fig:exp_mops} show that DE outperforms any other method at any instant in time if the average makespan is considered. Recalling that CPO does not produce feasible solutions in the first 2/3 of the time budget, the comparison of DE with CPO requires the analysis of the results in Tables~\ref{tab:exp_mops} and~\ref{tab:exp_mopsOPS}. The results in the tables show that DE outperforms CPO when the CPU time limit is 5 minutes, 30 minutes, or 2 hours. Results in the tables show that DE outperforms CPO also when the performance measure is the best makespan instead of the average makespan. The method that ranks in second place depends on the time limit and the performance measure (average or best makespan). Depending on the choice, CPO or ILS achieve second best result. The second place belongs to CPO when the average makespan is considered or when the CPU time limit is 2 hours. If the performance measure is the best makespan and the CPU time limit is 5 minutes or 30 minutes, the second place belongs to ILS. Concerning the best makespan and considering a CPU time limit of 2 hours, DE, GA, ILS, TS, and CPO obtained the best makespan 17, 10, 12, 12, and 13 times, respectively. Note that these numbers are slightly influenced by the presence of the method TS+DE that should be ignored. This is because TS+DE was the only method to find the best makespan in instance~6; so this instance is not computed for TS, that was the only method that found the second-best makespan for this instance. In any case, considering the average makespan, it is worth noting that, depending on whether the CPU time limit is 5 minutes, 30 minutes, or 2 hours, the difference between the methods that rank in first and last places is not larger than 0.8\%, 0.7\%, or 0.6\%, respectively.

\begin{figure}[ht!]
\centering 
\includegraphics{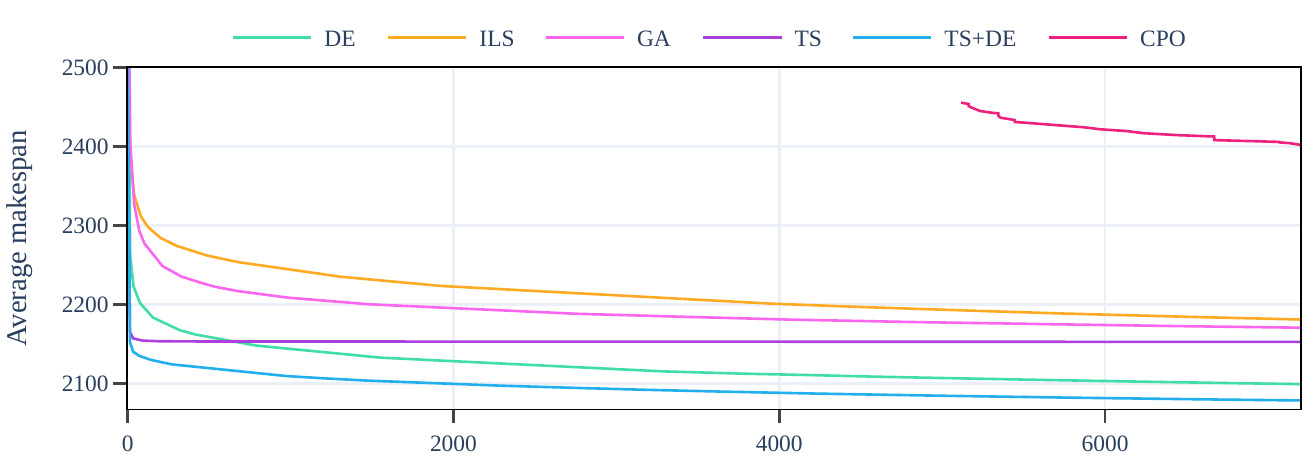}
\caption{Evolution of the average makespan over time of each proposed method and CPO applied to the LOPS set instances.} 
\label{fig:exp_lops} 
\end{figure}

\begin{figure}[ht!]
\centering 
\includegraphics{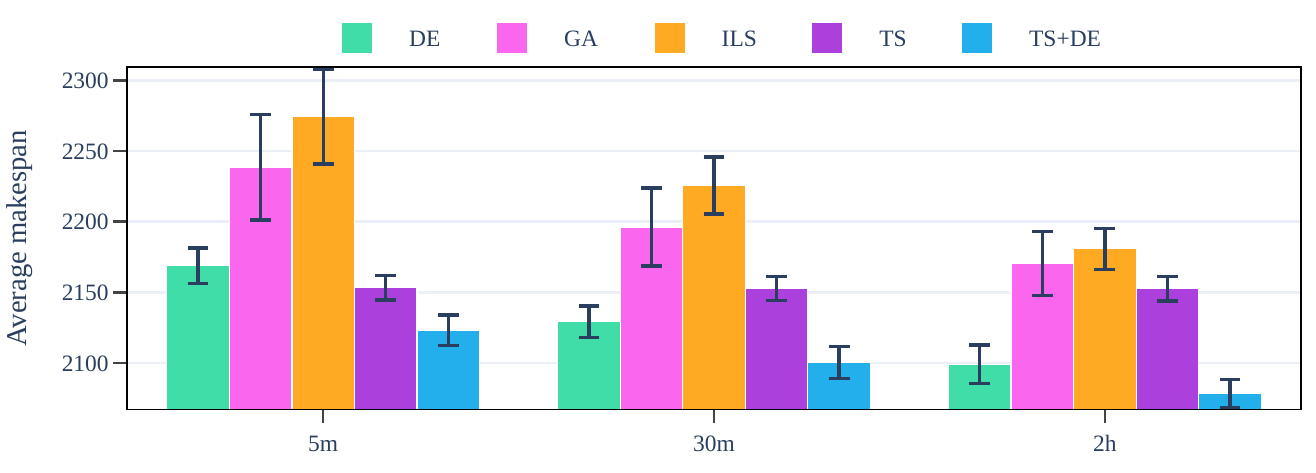}
\caption{Average makespans and pooled standard deviations that result from applying the proposed metaheuristic approaches twelve times to instances in the LOPS set with CPU time limits of 5 minutes, 30 minutes, and 2 hours.} 
\label{fig:pooled_std_dev_lops} 
\end{figure}

Figure~\ref{fig:exp_lops} shows the evolution of the average makespan (over the 12 runs and over all instances) when the five methods are applied to the instances in the LOPS set. Tables~\ref{tab:lops1} and~\ref{tab:lops2} present the best makespan while Tables~\ref{tab:lops3} and~\ref{tab:lops4} present the average makespan obtained by each metaheuristic method in each instance when the CPU time limit is 5 minutes, 30 minutes, or 2 hours. For each instance, numbers in bold represent the best results obtained by the methods under consideration. Method TS+DE should still be ignored. At the end of Tables~\ref{tab:lops1}--\ref{tab:lops4}, ``Avg. 1--50'' and ``Avg. 51--100'' correspond to the average of the instances contained in the table; while in Tables~\ref{tab:lops2} and~\ref{tab:lops4}, ``Avg. 1--100'' corresponds to the average over the whole LOPS set. In Table~\ref{tab:lops4}, and additional line exhibits the pooled standard deviation. Average makespans and pooled standard deviations are graphically represented in Figure~\ref{fig:pooled_std_dev_lops}. Table~\ref{tab:lops5} shows the results of applying CPO to the instances in the LOPS set. The symbol `` | '' means that CPO was not able to find a feasible solution within the time budget. 

\begin{table}[!ht]
\centering
\caption{Best makespan that results from applying the metaheuristics to the first-half of the instances in the LOPS set.}
\label{tab:lops1}
\resizebox{\textwidth}{!}{\begin{tabular}{cccccc|ccccc|ccccc}
\toprule
\multirow{2}{*}{Instance} & \multicolumn{5}{c}{CPU time limit: 5 minutes} & \multicolumn{5}{c}{CPU time limit: 30 minutes} & \multicolumn{5}{c}{CPU time limit: 2 hours} \\
\cmidrule(lr){2-6}
\cmidrule(lr){7-11} 
\cmidrule(lr){12-16}
 & DE & GA & ILS & TS & TS+DE & DE & GA & ILS & TS & TS+DE & DE & GA & ILS & TS & TS+DE \\
\midrule
1 & \textbf{516} & 525 & \textbf{516} & 527 & \textbf{516} & \textbf{516} & 522 & \textbf{516} & 526 & \textbf{516} & \textbf{513} & 520 & 516 & 524 & 514 \\  
2 & \textbf{641} & 655 & 647 & 660 & 642 & \textbf{641} & 651 & 647 & 659 & \textbf{641} & \textbf{638} & 648 & 642 & 658 & 639 \\  
3 & 620 & 623 & 622 & 644 & \textbf{617} & 615 & 622 & 616 & 644 & \textbf{614} & 613 & 618 & \textbf{612} & 640 & 613 \\  
4 & 741 & 750 & \textbf{736} & 769 & 742 & \textbf{736} & 743 & \textbf{736} & 767 & 738 & \textbf{736} & 741 & \textbf{736} & 767 & 737 \\  
5 & 826 & 836 & 829 & 857 & \textbf{824} & \textbf{820} & 831 & 821 & 856 & 821 & \textbf{820} & 824 & \textbf{820} & 854 & \textbf{820} \\  
6 & 689 & 691 & 690 & 726 & \textbf{683} & 681 & 683 & 679 & 724 & \textbf{678} & 676 & 679 & \textbf{674} & 724 & 675 \\  
7 & \textbf{896} & 903 & \textbf{896} & 935 & 897 & 890 & 895 & 892 & 935 & \textbf{888} & \textbf{886} & 894 & 890 & 935 & \textbf{886} \\  
8 & \textbf{1007} & 1014 & 1013 & 1049 & 1010 & 1001 & 1005 & 1004 & 1049 & \textbf{1000} & 999 & 1002 & 999 & 1049 & \textbf{997} \\  
9 & 919 & 921 & 920 & 969 & \textbf{915} & \textbf{905} & 914 & 911 & 969 & 906 & 902 & 910 & 906 & 969 & \textbf{900} \\  
10 & 765 & 775 & 768 & 829 & \textbf{762} & \textbf{748} & 764 & 754 & 829 & 752 & 745 & 755 & 749 & 829 & \textbf{744} \\  
11 & 1182 & 1191 & 1182 & 1217 & \textbf{1174} & 1165 & 1183 & 1167 & 1217 & \textbf{1162} & 1160 & 1164 & 1161 & 1217 & \textbf{1153} \\  
12 & 1168 & 1183 & 1170 & 1227 & \textbf{1164} & \textbf{1146} & 1162 & 1150 & 1227 & 1149 & \textbf{1138} & 1155 & \textbf{1138} & 1227 & 1140 \\  
13 & \textbf{988} & 1001 & 994 & 1055 & 990 & \textbf{971} & 986 & 976 & 1055 & 974 & \textbf{961} & 980 & 965 & 1055 & 965 \\  
14 & 1443 & 1450 & 1443 & 1498 & \textbf{1436} & 1430 & 1443 & 1428 & 1498 & \textbf{1427} & 1421 & 1431 & \textbf{1419} & 1498 & \textbf{1419} \\  
15 & 1386 & 1398 & 1384 & 1454 & \textbf{1380} & 1366 & 1384 & \textbf{1360} & 1454 & 1362 & 1355 & 1373 & 1356 & 1454 & \textbf{1352} \\  
16 & 1311 & 1327 & \textbf{1306} & 1366 & 1312 & 1293 & 1308 & \textbf{1288} & 1366 & 1293 & 1284 & 1301 & \textbf{1280} & 1366 & 1282 \\  
17 & \textbf{1041} & 1061 & \textbf{1041} & 1085 & 1045 & \textbf{1028} & 1046 & 1030 & 1085 & 1029 & 1019 & 1030 & \textbf{1016} & 1085 & 1021 \\  
18 & 1885 & 1898 & \textbf{1880} & 1956 & 1885 & 1862 & 1875 & \textbf{1855} & 1956 & 1859 & 1848 & 1858 & \textbf{1840} & 1956 & 1843 \\  
19 & 990 & 1007 & 997 & 1025 & \textbf{989} & 978 & 992 & 980 & 1025 & \textbf{974} & \textbf{962} & 985 & 964 & 1025 & 968 \\  
20 & \textbf{965} & 988 & 971 & 1013 & 967 & \textbf{948} & 969 & 952 & 1013 & 949 & \textbf{932} & 955 & 934 & 1013 & 934 \\  
21 & 1879 & 1894 & 1881 & 1948 & \textbf{1878} & \textbf{1852} & 1866 & 1853 & 1948 & 1854 & 1837 & 1850 & \textbf{1834} & 1948 & 1835 \\  
22 & 1417 & 1424 & 1442 & 1477 & \textbf{1402} & 1380 & 1404 & 1406 & 1477 & \textbf{1359} & 1361 & 1381 & 1383 & 1477 & \textbf{1349} \\  
23 & \textbf{1070} & 1083 & 1074 & 1105 & 1074 & \textbf{1050} & 1062 & 1056 & 1105 & 1059 & 1038 & 1050 & \textbf{1037} & 1105 & 1040 \\  
24 & \textbf{1914} & 1935 & 1921 & 1974 & 1919 & 1884 & 1905 & 1894 & 1974 & \textbf{1882} & 1859 & 1887 & 1870 & 1974 & \textbf{1857} \\  
25 & 1227 & 1245 & 1238 & 1272 & \textbf{1222} & \textbf{1204} & 1223 & 1209 & 1272 & 1205 & \textbf{1189} & 1208 & 1194 & 1272 & 1191 \\  
26 & 1281 & 1306 & 1293 & 1309 & \textbf{1279} & \textbf{1256} & 1284 & 1259 & 1309 & 1261 & \textbf{1237} & 1264 & 1238 & 1309 & 1243 \\  
27 & 1698 & 1718 & 1715 & 1753 & \textbf{1696} & \textbf{1670} & 1683 & 1677 & 1753 & \textbf{1670} & 1652 & 1674 & 1651 & 1753 & \textbf{1648} \\  
28 & 1929 & 1944 & 1953 & 1988 & \textbf{1926} & 1885 & 1925 & 1901 & 1988 & \textbf{1877} & 1858 & 1892 & 1879 & 1988 & \textbf{1853} \\  
29 & 2011 & 2072 & 2097 & 2098 & \textbf{1977} & 1950 & 2017 & 2009 & 2098 & \textbf{1943} & 1909 & 1986 & 1958 & 2098 & \textbf{1905} \\  
30 & 1557 & 1571 & 1576 & 1586 & \textbf{1548} & 1521 & 1546 & 1535 & 1586 & \textbf{1516} & 1496 & 1527 & 1510 & 1586 & \textbf{1487} \\  
31 & 1164 & 1185 & 1221 & 1179 & \textbf{1133} & 1128 & 1155 & 1167 & 1179 & \textbf{1103} & 1100 & 1136 & 1127 & 1179 & \textbf{1089} \\  
32 & 1062 & 1079 & 1094 & 1086 & \textbf{1058} & 1050 & 1062 & 1057 & 1086 & \textbf{1043} & 1034 & 1053 & 1039 & 1086 & \textbf{1030} \\  
33 & 2095 & 2114 & 2145 & 2151 & \textbf{2092} & 2058 & 2094 & 2086 & 2151 & \textbf{2052} & 2033 & 2075 & 2053 & 2151 & \textbf{2025} \\  
34 & 1438 & 1429 & 1465 & 1437 & \textbf{1390} & 1391 & 1405 & 1419 & 1437 & \textbf{1361} & 1356 & 1390 & 1388 & 1437 & \textbf{1342} \\  
35 & \textbf{2772} & 2835 & 2877 & 2895 & 2795 & \textbf{2732} & 2789 & 2795 & 2895 & 2740 & \textbf{2689} & 2754 & 2726 & 2895 & 2694 \\  
36 & 2482 & 2504 & 2549 & 2544 & \textbf{2478} & 2446 & 2482 & 2492 & 2544 & \textbf{2445} & 2419 & 2463 & 2443 & 2544 & \textbf{2417} \\  
37 & 1275 & 1299 & 1307 & 1287 & \textbf{1271} & 1253 & 1278 & 1281 & 1287 & \textbf{1247} & \textbf{1236} & 1263 & 1263 & 1287 & 1238 \\  
38 & 1159 & 1164 & 1182 & 1169 & \textbf{1145} & 1145 & 1142 & 1154 & 1169 & \textbf{1135} & 1125 & 1134 & 1133 & 1169 & \textbf{1119} \\  
39 & 1756 & 1754 & 1787 & 1760 & \textbf{1733} & 1721 & 1739 & 1758 & 1760 & \textbf{1706} & 1692 & 1714 & 1728 & 1760 & \textbf{1688} \\  
40 & 2204 & 2220 & 2261 & 2226 & \textbf{2181} & 2174 & 2200 & 2213 & 2226 & \textbf{2154} & 2142 & 2186 & 2163 & 2226 & \textbf{2131} \\  
41 & 2316 & 2344 & 2345 & 2304 & \textbf{2251} & 2268 & 2307 & 2281 & 2304 & \textbf{2229} & 2231 & 2275 & 2246 & 2304 & \textbf{2194} \\  
42 & 1582 & 1605 & 1655 & 1559 & \textbf{1539} & 1546 & 1565 & 1591 & 1559 & \textbf{1521} & 1520 & 1551 & 1551 & 1559 & \textbf{1502} \\  
43 & 2523 & 2540 & 2574 & 2548 & \textbf{2507} & 2490 & 2535 & 2526 & 2548 & \textbf{2486} & 2457 & 2500 & 2496 & 2548 & \textbf{2449} \\  
44 & 3678 & 3776 & 3839 & 3695 & \textbf{3638} & 3645 & 3715 & 3771 & 3695 & \textbf{3571} & 3578 & 3663 & 3702 & 3695 & \textbf{3516} \\  
45 & 2060 & 2065 & 2143 & 2069 & \textbf{2051} & 2043 & 2054 & 2095 & 2069 & \textbf{2033} & 2021 & 2044 & 2055 & 2069 & \textbf{2014} \\  
46 & 2185 & 2199 & 2236 & 2220 & \textbf{2180} & 2153 & 2185 & 2187 & 2220 & \textbf{2150} & \textbf{2122} & 2163 & 2150 & 2220 & 2123 \\  
47 & 3413 & 3539 & 3689 & 3438 & \textbf{3363} & 3356 & 3491 & 3602 & 3438 & \textbf{3330} & 3297 & 3454 & 3515 & 3438 & \textbf{3273} \\  
48 & 1272 & 1287 & 1313 & 1260 & \textbf{1242} & 1250 & 1274 & 1276 & 1260 & \textbf{1231} & 1231 & 1255 & 1256 & 1260 & \textbf{1218} \\  
49 & 2862 & 2893 & 2933 & 2889 & \textbf{2851} & 2836 & 2876 & 2886 & 2889 & \textbf{2829} & 2816 & 2844 & 2844 & 2889 & \textbf{2804} \\  
50 & 1374 & 1382 & 1403 & 1369 & \textbf{1347} & 1349 & 1369 & 1380 & 1369 & \textbf{1333} & 1331 & 1356 & 1359 & 1369 & \textbf{1321} \\  
\midrule
Avg. 1--50 &  1532.7 & 1552.0 & 1564.3 & 1569.1 & \textbf{1522.3} & 1508.5 & 1532.2 & 1531.6 & 1569.0 & \textbf{1501.1} & 1489.5 & 1516.3 & 1508.2 & 1568.8 & \textbf{1483.9} \\   
\bottomrule
\end{tabular}}
\end{table}

\begin{table}[!ht]
\centering
\caption{Best makespan that results from applying the metaheuristics to the second-half of the instances in the LOPS set.}
\label{tab:lops2}
\resizebox{\textwidth}{!}{\begin{tabular}{cccccc|ccccc|ccccc}
\toprule
\multirow{2}{*}{Instance} & \multicolumn{5}{c}{CPU time limit: 5 minutes} & \multicolumn{5}{c}{CPU time limit: 30 minutes} & \multicolumn{5}{c}{CPU time limit: 2 hours} \\
\cmidrule(lr){2-6}
\cmidrule(lr){7-11} 
\cmidrule(lr){12-16}
 & DE & GA & ILS & TS & TS+DE & DE & GA & ILS & TS & TS+DE & DE & GA & ILS & TS & TS+DE \\
\midrule
51 & 1567 & 1590 & 1649 & 1595 & \textbf{1557} & 1549 & 1573 & 1608 & 1595 & \textbf{1543} & 1530 & 1554 & 1573 & 1595 & \textbf{1527} \\  
52 & 1956 & 2034 & 2099 & 1952 & \textbf{1913} & 1920 & 1984 & 2040 & 1952 & \textbf{1888} & 1886 & 1952 & 1985 & 1952 & \textbf{1868} \\  
53 & 3629 & 3639 & 3678 & 3668 & \textbf{3623} & 3590 & 3614 & 3643 & 3668 & \textbf{3588} & 3555 & 3585 & 3599 & 3668 & \textbf{3549} \\  
54 & 1553 & 1578 & 1596 & 1542 & \textbf{1535} & 1535 & 1554 & 1563 & 1542 & \textbf{1525} & 1526 & 1542 & 1545 & 1542 & \textbf{1514} \\  
55 & 2001 & 2021 & 2082 & 1998 & \textbf{1986} & 1975 & 1999 & 2036 & 1998 & \textbf{1970} & 1953 & 1984 & 1993 & 1998 & \textbf{1944} \\  
56 & 2582 & 2584 & 2635 & 2577 & \textbf{2554} & 2552 & 2562 & 2602 & 2577 & \textbf{2535} & 2528 & 2537 & 2567 & 2577 & \textbf{2509} \\  
57 & 1917 & 1947 & 1989 & 1892 & \textbf{1878} & 1888 & 1936 & 1960 & 1892 & \textbf{1865} & 1873 & 1892 & 1907 & 1892 & \textbf{1857} \\  
58 & 1814 & 1841 & 1890 & 1807 & \textbf{1779} & 1790 & 1822 & 1853 & 1807 & \textbf{1764} & 1769 & 1795 & 1802 & 1807 & \textbf{1739} \\  
59 & 3415 & 3440 & 3485 & 3432 & \textbf{3411} & 3393 & 3418 & 3448 & 3432 & \textbf{3388} & 3356 & 3394 & 3408 & 3432 & \textbf{3355} \\  
60 & 1994 & 2044 & 2056 & 2004 & \textbf{1981} & 1970 & 2015 & 2031 & 2004 & \textbf{1966} & 1948 & 1998 & 2001 & 2004 & \textbf{1945} \\  
61 & 1988 & 2049 & 2127 & 2014 & \textbf{1980} & \textbf{1950} & 2034 & 2078 & 2014 & 1953 & \textbf{1930} & 2003 & 2032 & 2014 & 1934 \\  
62 & 3224 & 3289 & 3346 & 3138 & \textbf{3113} & 3137 & 3238 & 3284 & 3138 & \textbf{3108} & 3107 & 3200 & 3226 & 3138 & \textbf{3075} \\  
63 & 2512 & 2623 & 2727 & 2522 & \textbf{2495} & 2451 & 2574 & 2657 & 2522 & \textbf{2448} & \textbf{2412} & 2561 & 2571 & 2522 & 2416 \\  
64 & 1875 & 1933 & 1973 & 1911 & \textbf{1866} & 1858 & 1914 & 1948 & 1911 & \textbf{1857} & \textbf{1842} & 1897 & 1915 & 1911 & 1844 \\  
65 & 1909 & 1964 & 2047 & 1935 & \textbf{1893} & 1880 & 1951 & 2030 & 1935 & \textbf{1874} & 1860 & 1939 & 1994 & 1935 & \textbf{1852} \\  
66 & 1773 & 1813 & 1892 & 1761 & \textbf{1748} & 1750 & 1802 & 1878 & 1761 & \textbf{1747} & 1743 & 1794 & 1837 & 1761 & \textbf{1734} \\  
67 & 1687 & 1734 & 1755 & 1678 & \textbf{1671} & 1670 & 1714 & 1741 & 1678 & \textbf{1666} & 1658 & 1682 & 1712 & 1678 & \textbf{1653} \\  
68 & 3259 & 3385 & 3472 & 3074 & \textbf{3023} & 3081 & 3315 & 3425 & 3068 & \textbf{2967} & 2967 & 3272 & 3337 & 3068 & \textbf{2922} \\  
69 & 3037 & 3406 & 3479 & 2921 & \textbf{2891} & 2902 & 3296 & 3346 & 2921 & \textbf{2863} & 2844 & 3223 & 3220 & 2921 & \textbf{2831} \\  
70 & 3240 & 3269 & 3508 & 3161 & \textbf{3123} & 3121 & 3248 & 3461 & 3161 & \textbf{3053} & 2997 & 3227 & 3376 & 3161 & \textbf{2973} \\  
71 & 2179 & 2247 & 2280 & 2189 & \textbf{2173} & 2167 & 2228 & 2252 & 2189 & \textbf{2158} & 2145 & 2217 & 2227 & 2189 & \textbf{2142} \\  
72 & 1957 & 2031 & 2073 & 1802 & \textbf{1790} & 1880 & 1986 & 2024 & 1802 & \textbf{1787} & 1802 & 1964 & 1961 & 1802 & \textbf{1763} \\  
73 & 3967 & 3992 & 4086 & 3940 & \textbf{3915} & 3907 & 3981 & 4062 & 3940 & \textbf{3902} & \textbf{3867} & 3964 & 4013 & 3940 & 3869 \\  
74 & 4342 & 4355 & 4389 & 4260 & \textbf{4238} & 4228 & 4341 & 4354 & 4260 & \textbf{4199} & 4171 & 4290 & 4313 & 4260 & \textbf{4157} \\  
75 & 3635 & 3667 & 3712 & 3655 & \textbf{3632} & \textbf{3607} & 3642 & 3684 & 3655 & 3612 & \textbf{3583} & 3629 & 3653 & 3655 & 3585 \\  
76 & 2444 & 2570 & 2638 & \textbf{2264} & 2266 & 2362 & 2519 & 2585 & 2252 & \textbf{2221} & 2250 & 2481 & 2522 & 2252 & \textbf{2184} \\  
77 & 1799 & 1864 & 1904 & 1792 & \textbf{1777} & 1770 & 1846 & 1879 & 1792 & \textbf{1764} & 1751 & 1829 & 1837 & 1792 & \textbf{1750} \\  
78 & 1671 & 1694 & 1748 & 1669 & \textbf{1659} & \textbf{1650} & 1685 & 1716 & 1669 & \textbf{1650} & \textbf{1634} & 1671 & 1686 & 1669 & \textbf{1634} \\  
79 & 1750 & 1799 & 1841 & 1751 & \textbf{1738} & 1736 & 1786 & 1803 & 1751 & \textbf{1733} & 1726 & 1759 & 1777 & 1751 & \textbf{1722} \\  
80 & 1788 & 1891 & 1893 & 1739 & \textbf{1732} & 1745 & 1832 & 1864 & 1739 & \textbf{1723} & 1711 & 1804 & 1819 & 1739 & \textbf{1697} \\  
81 & 3253 & 3260 & 3375 & 3145 & \textbf{3140} & 3171 & 3189 & 3354 & 3145 & \textbf{3122} & 3140 & 3171 & 3327 & 3145 & \textbf{3086} \\  
82 & 4691 & 4742 & 4784 & 4693 & \textbf{4683} & 4665 & 4706 & 4757 & 4693 & \textbf{4659} & 4635 & 4687 & 4732 & 4693 & \textbf{4634} \\  
83 & 3122 & 3175 & 3192 & 3125 & \textbf{3091} & 3088 & 3160 & 3174 & 3125 & \textbf{3072} & 3062 & 3136 & 3144 & 3125 & \textbf{3050} \\  
84 & 2020 & 2056 & 2121 & 1960 & \textbf{1951} & 1961 & 2026 & 2076 & 1960 & \textbf{1942} & 1940 & 2008 & 2042 & 1960 & \textbf{1931} \\  
85 & 2400 & 3132 & 3196 & 2379 & \textbf{2367} & 2369 & 2919 & 2972 & 2379 & \textbf{2344} & 2344 & 2819 & 2756 & 2379 & \textbf{2332} \\  
86 & 2330 & 2786 & 2966 & 2296 & \textbf{2267} & 2282 & 2507 & 2760 & 2296 & \textbf{2248} & 2246 & 2425 & 2521 & 2296 & \textbf{2237} \\  
87 & 3230 & 3315 & 3455 & 2962 & \textbf{2938} & 3137 & 3243 & 3390 & 2962 & \textbf{2909} & 3055 & 3203 & 3311 & 2962 & \textbf{2876} \\  
88 & 5401 & 5481 & 5523 & 5405 & \textbf{5382} & 5358 & 5399 & 5490 & 5405 & \textbf{5351} & 5316 & 5372 & 5455 & 5405 & \textbf{5315} \\  
89 & 3863 & 3943 & 3942 & 3760 & \textbf{3737} & 3760 & 3858 & 3893 & 3760 & \textbf{3720} & 3716 & 3844 & 3864 & 3760 & \textbf{3681} \\  
90 & 3350 & 3438 & 3491 & 3328 & \textbf{3310} & 3344 & 3389 & 3476 & 3328 & \textbf{3280} & 3308 & 3380 & 3439 & 3328 & \textbf{3257} \\  
91 & 2456 & 2598 & 2637 & 2418 & \textbf{2401} & 2421 & 2523 & 2585 & 2418 & \textbf{2376} & 2380 & 2506 & 2550 & 2418 & \textbf{2357} \\  
92 & 3579 & 3659 & 3724 & \textbf{3212} & 3248 & 3540 & 3622 & 3662 & \textbf{3167} & 3212 & 3423 & 3560 & 3610 & \textbf{3167} & 3205 \\  
93 & 2194 & 2260 & 2372 & 2144 & \textbf{2127} & 2155 & 2213 & 2311 & 2144 & \textbf{2116} & 2137 & 2201 & 2263 & 2144 & \textbf{2111} \\  
94 & 4458 & 4543 & 4616 & 4430 & \textbf{4407} & 4422 & 4503 & 4581 & 4430 & \textbf{4393} & 4378 & 4471 & 4557 & 4430 & \textbf{4370} \\  
95 & 2647 & 2763 & 2827 & 2607 & \textbf{2580} & 2601 & 2675 & 2767 & 2607 & \textbf{2570} & 2571 & 2648 & 2746 & 2607 & \textbf{2548} \\  
96 & 3437 & 3588 & 3622 & \textbf{3112} & 3120 & 3384 & 3533 & 3600 & 3112 & \textbf{3087} & 3296 & 3505 & 3559 & 3112 & \textbf{3046} \\  
97 & 2538 & 2837 & 3204 & 2547 & \textbf{2518} & 2517 & 2708 & 2981 & 2547 & \textbf{2506} & 2491 & 2646 & 2763 & 2547 & \textbf{2487} \\  
98 & 5566 & 5637 & 5692 & 5534 & \textbf{5511} & 5506 & 5601 & 5664 & 5534 & \textbf{5484} & 5458 & 5562 & 5632 & 5534 & \textbf{5455} \\  
99 & 2146 & 2236 & 2227 & \textbf{1972} & 1989 & 2105 & 2192 & 2196 & \textbf{1972} & 1983 & 2065 & 2169 & 2179 & 1972 & \textbf{1969} \\  
100 & 3340 & 3399 & 3426 & 3306 & \textbf{3294} & 3310 & 3356 & 3397 & 3306 & \textbf{3291} & 3270 & 3343 & 3384 & 3306 & \textbf{3265} \\  
\midrule
Avg. 51--100 &  2769.7 & 2862.8 & 2928.8 & 2719.6 & \textbf{2700.0} & 2722.2 & 2814.6 & 2878.8 & 2718.3 & \textbf{2679.6} & 2683.1 & 2785.9 & 2824.8 & 2718.3 & \textbf{2655.1} \\   
\midrule
Avg. 1--100 & 2151.2 & 2207.4 & 2246.6 & 2144.4 & \textbf{2111.2} & 2115.4 & 2173.4 & 2205.2 & 2143.7 & \textbf{2090.4} & 2086.3 & 2151.1 & 2166.5 & 2143.6 & \textbf{2069.5} \\
\bottomrule
\end{tabular}}
\end{table}

\begin{table}[!ht]
\centering
\caption{Average makespan that results from applying the metaheuristics to the first-half of the instances in the LOPS set.}
\label{tab:lops3}
\resizebox{\textwidth}{!}{\begin{tabular}{cccccc|ccccc|ccccc}
\toprule
\multirow{2}{*}{Instance} & \multicolumn{5}{c}{CPU time limit: 5 minutes} & \multicolumn{5}{c}{CPU time limit: 30 minutes} & \multicolumn{5}{c}{CPU time limit: 2 hours} \\
\cmidrule(lr){2-6}
\cmidrule(lr){7-11} 
\cmidrule(lr){12-16}
 & DE & GA & ILS & TS & TS+DE & DE & GA & ILS & TS & TS+DE & DE & GA & ILS & TS & TS+DE \\
\midrule
1 & \textbf{519} & 528.5 & 520.5 & 528.5 & 521 & \textbf{517} & 524.2 & 518.8 & 526.5 & 518.9 & \textbf{515.8} & 523.8 & 518.5 & 524.8 & 518.1 \\  
2 & 650 & 657.2 & 650.8 & 661.5 & \textbf{647.6} & 647.5 & 653 & 648 & 659.5 & \textbf{644} & 644.2 & 651.5 & 644.2 & 659.2 & \textbf{641.4} \\  
3 & \textbf{621} & 625.8 & 623.5 & 647.5 & 621.2 & 618 & 622.8 & 621.5 & 645.5 & \textbf{617.6} & 616.5 & 620.8 & 618.5 & 642.8 & \textbf{615.4} \\  
4 & \textbf{743.8} & 752.2 & 745.5 & 771.2 & 745.6 & \textbf{738} & 745 & 741 & 769.2 & 741.6 & \textbf{737.5} & 743 & 739.8 & 768.5 & 739.2 \\  
5 & \textbf{827.8} & 839.5 & 833.5 & 861.5 & 828.2 & \textbf{824.5} & 832 & 827.8 & 857.2 & 824.6 & \textbf{821.2} & 828.5 & 822.8 & 856.5 & 822 \\  
6 & 692.8 & 702.8 & 695.5 & 728.8 & \textbf{690.5} & 683 & 690 & 685 & 726.5 & \textbf{681.4} & \textbf{676.8} & 685.8 & 679.5 & 724.8 & 677.6 \\  
7 & \textbf{897.2} & 910.8 & 902 & 942 & 898.4 & 893.2 & 899.8 & 895 & 940.5 & \textbf{892.4} & 889.5 & 896 & 892.8 & 939.2 & \textbf{888.8} \\  
8 & \textbf{1012} & 1020.2 & 1019 & 1052 & 1013.5 & \textbf{1004.8} & 1008.5 & 1006.5 & 1052 & 1005 & \textbf{999.8} & 1005.2 & 1003.2 & 1052 & 1000.8 \\  
9 & 922 & 935 & 925.2 & 977.8 & \textbf{921.8} & 911.2 & 925.2 & 912.2 & 977.8 & \textbf{910.2} & 905.2 & 912.5 & 907.8 & 977.8 & \textbf{904.9} \\  
10 & \textbf{766.8} & 785.5 & 773.8 & 831.5 & 768.6 & \textbf{751.8} & 768.2 & 761 & 831.5 & 754.9 & \textbf{746.8} & 760.5 & 750.5 & 831.2 & 748.6 \\  
11 & 1188.2 & 1201.8 & 1190.2 & 1227.5 & \textbf{1179.1} & 1174.5 & 1185.5 & 1171.8 & 1227.5 & \textbf{1166.4} & 1163.2 & 1174.8 & 1162.5 & 1227.5 & \textbf{1158.4} \\  
12 & 1172 & 1191.8 & 1184 & 1229 & \textbf{1170.9} & 1156.8 & 1177.5 & \textbf{1153.8} & 1228.2 & 1154.4 & 1146.8 & 1164.8 & \textbf{1141.5} & 1228.2 & 1145 \\  
13 & 997.5 & 1006.5 & 1004.8 & 1064.5 & \textbf{997} & \textbf{976} & 994 & 983 & 1064.5 & 980.9 & \textbf{967} & 983 & 970.8 & 1064.5 & 968.4 \\  
14 & \textbf{1446.5} & 1458.8 & 1447.5 & 1504.5 & 1447.4 & 1433.5 & 1449.2 & 1433.8 & 1504.5 & \textbf{1432.6} & 1426.2 & 1436.2 & \textbf{1423.8} & 1504.5 & 1424.8 \\  
15 & 1394.8 & 1407.8 & 1394.8 & 1468.2 & \textbf{1386.9} & 1370.5 & 1390.2 & 1369 & 1466.2 & \textbf{1367.6} & 1358.5 & 1376 & 1359.5 & 1465.2 & \textbf{1355.2} \\  
16 & 1317.5 & 1331.5 & 1320.5 & 1371 & \textbf{1315.8} & \textbf{1296.8} & 1314 & 1301.2 & 1371 & 1297 & \textbf{1286.8} & 1305 & \textbf{1286.8} & 1371 & 1287.4 \\  
17 & \textbf{1047} & 1067.8 & \textbf{1047} & 1087.5 & 1049.5 & \textbf{1032.8} & 1047.2 & 1034.5 & 1087.5 & 1033.8 & 1024 & 1032.5 & \textbf{1020} & 1087.5 & 1024.2 \\  
18 & 1889.2 & 1901 & \textbf{1886} & 1966.2 & 1892.4 & 1866.2 & 1878.8 & \textbf{1863} & 1966.2 & 1863.4 & 1855.2 & 1863.2 & \textbf{1844.8} & 1966.2 & 1850.4 \\  
19 & 995.2 & 1010.2 & 1000.8 & 1028.2 & \textbf{994.2} & 981.5 & 995.8 & 984.5 & 1028.2 & \textbf{980.6} & 970.8 & 988 & 970.8 & 1028.2 & \textbf{970.6} \\  
20 & \textbf{977.5} & 990.5 & 977.8 & 1019.5 & 977.6 & 955.8 & 980.5 & \textbf{954.8} & 1019.5 & 956.8 & 941.8 & 964.8 & \textbf{935.2} & 1019.5 & 939.6 \\  
21 & 1889.2 & 1911.8 & 1894.5 & 1957 & \textbf{1888.8} & 1859 & 1874.5 & 1860.2 & 1957 & \textbf{1858.6} & 1845.8 & 1859.8 & \textbf{1841.5} & 1957 & 1841.8 \\  
22 & 1421.5 & 1435.5 & 1469.2 & 1487.2 & \textbf{1408} & 1385.2 & 1410.2 & 1414 & 1487.2 & \textbf{1375.6} & 1364 & 1393 & 1392.8 & 1487.2 & \textbf{1357.9} \\  
23 & \textbf{1076.8} & 1090 & 1087 & 1109.5 & 1081.8 & \textbf{1055.5} & 1076.8 & 1067 & 1109.5 & 1062.1 & \textbf{1043} & 1061.5 & 1046.2 & 1109.5 & 1048.5 \\  
24 & 1923.8 & 1946 & 1938.5 & 1978 & \textbf{1922.9} & 1889.5 & 1918 & 1906 & 1978 & \textbf{1886.4} & 1870.8 & 1902 & 1879.2 & 1978 & \textbf{1866.2} \\  
25 & 1232.5 & 1254.2 & 1244.8 & 1281 & \textbf{1228.8} & 1208.5 & 1229.8 & 1215.5 & 1281 & \textbf{1208} & \textbf{1193.5} & 1212.5 & 1199.5 & 1281 & 1194.1 \\  
26 & \textbf{1286.8} & 1310.2 & 1307.8 & 1326.2 & 1287.6 & \textbf{1263.8} & 1292.5 & 1271.2 & 1326.2 & 1270.2 & \textbf{1242} & 1279.2 & 1244 & 1326.2 & 1249.2 \\  
27 & \textbf{1704.5} & 1727.8 & 1732.8 & 1761.8 & 1705.1 & 1676 & 1707.5 & 1690.8 & 1761.8 & \textbf{1675.6} & \textbf{1653} & 1686.2 & 1663.8 & 1761.8 & 1654.9 \\  
28 & 1935.2 & 1956.2 & 1993 & 1999.2 & \textbf{1931.9} & 1898.2 & 1931 & 1931.5 & 1999.2 & \textbf{1884.6} & 1872.5 & 1899.2 & 1894 & 1999.2 & \textbf{1861.8} \\  
29 & 2019.5 & 2075.8 & 2113.2 & 2108.2 & \textbf{1999.8} & 1968.8 & 2032.5 & 2029.5 & 2108.2 & \textbf{1953.2} & 1930.2 & 1989.5 & 1983 & 2108.2 & \textbf{1918.4} \\  
30 & 1565 & 1573.2 & 1588.5 & 1595.8 & \textbf{1557.5} & 1529.2 & 1551.2 & 1550.5 & 1595.8 & \textbf{1521.5} & 1503.5 & 1534 & 1516 & 1595.8 & \textbf{1496.8} \\  
31 & 1170 & 1207 & 1231.8 & 1182.8 & \textbf{1141.5} & 1131 & 1176 & 1178.2 & 1182.8 & \textbf{1111.4} & 1105.5 & 1151.5 & 1137 & 1182.8 & \textbf{1092.2} \\  
32 & 1074.2 & 1084 & 1102.8 & 1091 & \textbf{1061.4} & 1054.2 & 1067.5 & 1066.5 & 1091 & \textbf{1046.2} & 1039.2 & 1056.8 & 1044.5 & 1091 & \textbf{1034} \\  
33 & 2100.2 & 2123 & 2161.2 & 2164 & \textbf{2099.1} & 2068.8 & 2101.5 & 2096 & 2164 & \textbf{2064.2} & 2036.2 & 2087 & 2062.5 & 2164 & \textbf{2034.5} \\  
34 & 1454.5 & 1451.5 & 1493.8 & 1443.5 & \textbf{1404.4} & 1406.5 & 1427 & 1435.5 & 1443.5 & \textbf{1368.9} & 1369.2 & 1399 & 1401 & 1443.5 & \textbf{1345.2} \\  
35 & 2818.8 & 2867 & 2922 & 2907.2 & \textbf{2816} & \textbf{2744.8} & 2817.5 & 2827.8 & 2907.2 & 2748.5 & \textbf{2700.5} & 2786.8 & 2752.8 & 2907.2 & 2705.8 \\  
36 & 2494.8 & 2522.2 & 2604.8 & 2548.8 & \textbf{2483.1} & 2456.5 & 2486.8 & 2524.2 & 2548.8 & \textbf{2451} & 2425 & 2468.8 & 2467 & 2548.8 & \textbf{2420.1} \\  
37 & 1281.8 & 1304.5 & 1316.2 & 1290.8 & \textbf{1274.4} & 1263.2 & 1287.5 & 1290.2 & 1290.8 & \textbf{1256.1} & 1241.2 & 1270.2 & 1267 & 1290.8 & \textbf{1241.1} \\  
38 & 1173.5 & 1177.8 & 1192.8 & 1173.8 & \textbf{1150.4} & 1151.8 & 1156 & 1159.8 & 1173.8 & \textbf{1137.1} & 1131.5 & 1143.8 & 1137.2 & 1173.8 & \textbf{1122.8} \\  
39 & 1768.5 & 1794.2 & 1800 & 1767 & \textbf{1739.1} & 1734.8 & 1762 & 1764 & 1767 & \textbf{1716} & 1706 & 1738.8 & 1736.8 & 1767 & \textbf{1693.4} \\  
40 & 2217.2 & 2235.2 & 2289.5 & 2236.2 & \textbf{2189.6} & 2180.8 & 2210.5 & 2234.2 & 2236.2 & \textbf{2162.9} & 2148.8 & 2192.8 & 2184.2 & 2236.2 & \textbf{2135.9} \\  
41 & 2332 & 2386.5 & 2355.8 & 2323 & \textbf{2278.5} & 2278.5 & 2342.5 & 2294.8 & 2323 & \textbf{2247.8} & 2243 & 2296.8 & 2261.2 & 2323 & \textbf{2209.9} \\  
42 & 1601.2 & 1618 & 1668.2 & 1570.2 & \textbf{1542.9} & 1559 & 1585.2 & 1602.2 & 1570.2 & \textbf{1527.6} & 1530 & 1561.8 & 1558.2 & 1570.2 & \textbf{1506.1} \\  
43 & 2530.8 & 2567.5 & 2601.8 & 2552 & \textbf{2522.4} & 2498.8 & 2545.5 & 2557.8 & 2552 & \textbf{2491.1} & 2460.2 & 2529.2 & 2515.2 & 2552 & \textbf{2458.6} \\  
44 & 3748 & 3835.2 & 3886.8 & 3716 & \textbf{3673.6} & 3680.8 & 3774.5 & 3812.5 & 3716 & \textbf{3614.1} & 3629.5 & 3746.8 & 3744.8 & 3716 & \textbf{3545} \\  
45 & 2082.5 & 2116 & 2151.8 & 2077 & \textbf{2054.1} & 2053.2 & 2086.8 & 2100.5 & 2077 & \textbf{2038} & 2028 & 2065.8 & 2059.2 & 2077 & \textbf{2020} \\  
46 & 2190.2 & 2215 & 2258 & 2225.5 & \textbf{2186} & 2160.2 & 2192 & 2198.8 & 2225.5 & \textbf{2155.6} & 2133.8 & 2168.8 & 2154.8 & 2225.5 & \textbf{2128.6} \\  
47 & 3434.5 & 3753.5 & 3822.2 & 3461.8 & \textbf{3389.8} & 3369 & 3671.5 & 3720.2 & 3461.8 & \textbf{3336.8} & 3313 & 3586.2 & 3618.5 & 3461.8 & \textbf{3285.6} \\  
48 & 1280.8 & 1293.8 & 1342.5 & 1263.2 & \textbf{1247.9} & 1259 & 1285 & 1296 & 1263.2 & \textbf{1236.1} & 1236 & 1264 & 1264.8 & 1263.2 & \textbf{1222.9} \\  
49 & 2885.5 & 2927 & 2947.2 & 2896.8 & \textbf{2864.5} & 2847.8 & 2898.5 & 2895.2 & 2896.8 & \textbf{2838.6} & 2822.8 & 2866.2 & 2853.5 & 2896.8 & \textbf{2811.4} \\  
50 & 1378.5 & 1389.8 & 1417.5 & 1377.8 & \textbf{1353.8} & 1354.2 & 1373.5 & 1394.8 & 1377.8 & \textbf{1339} & 1335 & 1360 & 1366.8 & 1377.8 & \textbf{1326.1} \\  
\midrule
Avg. 1--50 &  1543.0 & 1569.5 & 1581.6 & 1576.8 & \textbf{1531.0} & 1516.4 & 1547.1 & 1545.0 & 1576.4 & \textbf{1508.1} & 1496.9 & 1529.5 & 1518.8 & 1576.2 & \textbf{1490.2} \\  
\bottomrule
\end{tabular}}
\end{table}

\begin{table}[!ht]
\centering
\caption{Average makespan that results from applying the metaheuristics to the second-half of the instances in the LOPS set.}
\label{tab:lops4}
\resizebox{\textwidth}{!}{\begin{tabular}{cccccc|ccccc|ccccc}
\toprule
\multirow{2}{*}{Instance} & \multicolumn{5}{c}{CPU time limit: 5 minutes} & \multicolumn{5}{c}{CPU time limit: 30 minutes} & \multicolumn{5}{c}{CPU time limit: 2 hours} \\
\cmidrule(lr){2-6}
\cmidrule(lr){7-11} 
\cmidrule(lr){12-16}
 & DE & GA & ILS & TS & TS+DE & DE & GA & ILS & TS & TS+DE & DE & GA & ILS & TS & TS+DE \\
\midrule
51 & 1585.8 & 1611.5 & 1668.5 & 1598.2 & \textbf{1564.5} & 1561 & 1591.2 & 1621.8 & 1598.2 & \textbf{1546.8} & 1539.5 & 1571 & 1584.5 & 1598.2 & \textbf{1530.9} \\  
52 & 1979 & 2060.2 & 2159 & 1964 & \textbf{1920.4} & 1933.8 & 2007.5 & 2078 & 1964 & \textbf{1895.2} & 1896.5 & 1971 & 2001.8 & 1964 & \textbf{1873.9} \\  
53 & 3637.5 & 3645.8 & 3683.5 & 3672 & \textbf{3628.6} & 3598.5 & 3618 & 3645 & 3672 & \textbf{3592.4} & 3563.2 & 3590 & 3601 & 3672 & \textbf{3556.4} \\  
54 & 1566 & 1578.8 & 1606 & 1555.5 & \textbf{1538.4} & 1546.2 & 1561.2 & 1575 & 1555.5 & \textbf{1529.9} & 1530.2 & 1547.2 & 1553.5 & 1555.5 & \textbf{1516.6} \\  
55 & 2009 & 2058 & 2096 & 2013 & \textbf{1993.1} & 1982.8 & 2022.8 & 2048.8 & 2013 & \textbf{1972.4} & 1962 & 1997.5 & 1999.5 & 2013 & \textbf{1951.2} \\  
56 & 2597 & 2610.8 & 2647.8 & 2586.8 & \textbf{2565.5} & 2563.5 & 2589 & 2612 & 2586.8 & \textbf{2542.6} & 2535.8 & 2569.2 & 2574.2 & 2586.8 & \textbf{2515.9} \\  
57 & 1955.2 & 1957.2 & 2055 & 1906.5 & \textbf{1884.6} & 1928.2 & 1948.5 & 1999 & 1906.5 & \textbf{1872.8} & 1883.5 & 1920.5 & 1925.5 & 1906.5 & \textbf{1859} \\  
58 & 1822.2 & 1847.5 & 1897 & 1812.2 & \textbf{1786.8} & 1799 & 1830.2 & 1858.5 & 1812.2 & \textbf{1773.5} & 1780.8 & 1811 & 1814.5 & 1812.2 & \textbf{1754.1} \\  
59 & 3429 & 3454.8 & 3494.2 & 3442.2 & \textbf{3416.9} & 3400.2 & 3430 & 3459 & 3442.2 & \textbf{3393.4} & 3365.2 & 3408 & 3421.8 & 3442.2 & \textbf{3363.4} \\  
60 & 1998.8 & 2050.2 & 2076 & 2006.5 & \textbf{1987.9} & 1976.2 & 2024 & 2039.8 & 2006.5 & \textbf{1970.9} & \textbf{1953.2} & 2005 & 2007.2 & 2006.5 & \textbf{1953.2} \\  
61 & 2001 & 2070.5 & 2139.2 & 2025.5 & \textbf{1989.8} & \textbf{1962.8} & 2046.5 & 2081.8 & 2025.5 & 1963 & \textbf{1932.2} & 2014.5 & 2035.2 & 2025.5 & 1937.9 \\  
62 & 3280.5 & 3350.8 & 3366.2 & 3158 & \textbf{3121.2} & 3187.8 & 3273.5 & 3311.8 & 3158 & \textbf{3111.5} & 3116.5 & 3232.5 & 3246.8 & 3158 & \textbf{3104.6} \\  
63 & 2538.5 & 2648.2 & 2742.8 & 2540.8 & \textbf{2506.1} & 2465.5 & 2605.5 & 2686.2 & 2540.8 & \textbf{2459.6} & 2424.8 & 2569 & 2596 & 2540.8 & \textbf{2422.6} \\  
64 & 1883.2 & 1952.5 & 2005.5 & 1914.8 & \textbf{1879.6} & 1863 & 1925 & 1981 & 1914.8 & \textbf{1860.5} & 1845.8 & 1903.8 & 1937 & 1914.8 & \textbf{1845.1} \\  
65 & 1945.2 & 1992.5 & 2100.2 & 1939.5 & \textbf{1906} & 1892.2 & 1976 & 2049.2 & 1939.5 & \textbf{1884.2} & 1866 & 1957 & 2003 & 1939.5 & \textbf{1862.6} \\  
66 & 1791.5 & 1850.5 & 1927.8 & 1767.2 & \textbf{1752.4} & 1755.8 & 1827.5 & 1891 & 1767.2 & \textbf{1749} & 1746.5 & 1808 & 1846 & 1767.2 & \textbf{1743.8} \\  
67 & 1698 & 1766.2 & 1794.2 & 1685.2 & \textbf{1677.2} & 1678.5 & 1728.5 & 1765.2 & 1685.2 & \textbf{1670.2} & 1662.5 & 1704 & 1727.8 & 1685.2 & \textbf{1658.4} \\  
68 & 3386.5 & 3444.8 & 3488.8 & 3082 & \textbf{3073.4} & 3234 & 3358 & 3440.2 & 3080 & \textbf{2985.2} & 3052.8 & 3324 & 3364.5 & 3080 & \textbf{2930.1} \\  
69 & 3087.5 & 3443.5 & 3522 & 2945.2 & \textbf{2904} & 2924 & 3309.5 & 3388.8 & 2945.2 & \textbf{2871.9} & 2859 & 3239.2 & 3261.8 & 2945.2 & \textbf{2835.9} \\  
70 & 3246.5 & 3342.2 & 3544 & 3178 & \textbf{3141.5} & 3158.5 & 3308.5 & 3474.8 & 3178 & \textbf{3078.6} & 3069.5 & 3274 & 3395.2 & 3178 & \textbf{2989.8} \\  
71 & 2211.5 & 2270.2 & 2315 & 2196.5 & \textbf{2181.6} & 2179.2 & 2251 & 2284.2 & 2196.5 & \textbf{2166.4} & 2151.2 & 2229.2 & 2245 & 2196.5 & \textbf{2147.5} \\  
72 & 1979.5 & 2070.5 & 2094.8 & 1815.5 & \textbf{1801.1} & 1891.8 & 2023 & 2039.5 & 1815.5 & \textbf{1793} & 1811.8 & 1987.8 & 1970.2 & 1815.5 & \textbf{1774.5} \\  
73 & 3982 & 4048.5 & 4114.8 & 3951.5 & \textbf{3925.2} & 3929.5 & 4006.2 & 4078.8 & 3951.5 & \textbf{3911.1} & 3890.2 & 3983 & 4033.8 & 3951.5 & \textbf{3879.8} \\  
74 & 4398 & 4416.2 & 4418.8 & 4286.2 & \textbf{4265.5} & 4291.5 & 4348.5 & 4373.2 & 4286.2 & \textbf{4232.4} & 4229.8 & 4306.2 & 4329.8 & 4286.2 & \textbf{4176.5} \\  
75 & 3639.5 & 3688.5 & 3717.5 & 3662.2 & \textbf{3635.5} & \textbf{3610.8} & 3663.5 & 3684.8 & 3662.2 & 3615.1 & \textbf{3588.2} & 3642.5 & 3657.5 & 3662.2 & 3589 \\  
76 & 2514.2 & 2619.2 & 2664 & \textbf{2267} & 2343.6 & 2391 & 2560 & 2597.5 & \textbf{2261.2} & 2280 & 2300.2 & 2518.5 & 2535.8 & 2261.2 & \textbf{2221.2} \\  
77 & 1806.2 & 1889.2 & 1919.2 & 1798.2 & \textbf{1781.2} & 1774.8 & 1854.2 & 1887.8 & 1798.2 & \textbf{1770.9} & 1760.2 & 1834.2 & 1843.8 & 1798.2 & \textbf{1760} \\  
78 & 1681.8 & 1724.5 & 1773.2 & 1676.2 & \textbf{1664} & \textbf{1653.8} & 1699.2 & 1734 & 1676.2 & 1654.8 & \textbf{1636} & 1685.5 & 1695.8 & 1676.2 & 1636.9 \\  
79 & 1759.8 & 1829 & 1844.5 & 1756.5 & \textbf{1745.6} & 1743.2 & 1793.8 & 1811.8 & 1756.5 & \textbf{1738.6} & 1727.5 & 1770.5 & 1781 & 1756.5 & \textbf{1726.2} \\  
80 & 1801.2 & 1900 & 1907.5 & 1746.2 & \textbf{1741.8} & 1760.2 & 1842 & 1875.2 & 1746.2 & \textbf{1731.6} & 1731.8 & 1821 & 1827.8 & 1746.2 & \textbf{1705.6} \\  
81 & 3260 & 3309.5 & 3387.8 & 3149.5 & \textbf{3144.2} & 3206 & 3243 & 3363 & 3149.5 & \textbf{3136.8} & 3150.8 & 3216.8 & 3330.2 & 3149.5 & \textbf{3107.5} \\  
82 & 4712.5 & 4761.8 & 4797.8 & 4705 & \textbf{4689.5} & 4682.2 & 4724.2 & 4773 & 4705 & \textbf{4666.6} & 4646.5 & 4703.2 & 4749.5 & 4705 & \textbf{4638.6} \\  
83 & 3141.8 & 3200.8 & 3224.5 & 3134.8 & \textbf{3102.4} & 3097.2 & 3169 & 3198.5 & 3134.8 & \textbf{3081.4} & 3063.2 & 3151.8 & 3170.2 & 3134.8 & \textbf{3056.9} \\  
84 & 2038 & 2103 & 2163 & 1961.2 & \textbf{1960.6} & 1974.5 & 2067.2 & 2108.2 & 1961.2 & \textbf{1951.9} & 1957.8 & 2041.5 & 2061.2 & 1961.2 & \textbf{1937.6} \\  
85 & 2473.8 & 3222 & 3236.2 & 2395.5 & \textbf{2375.5} & 2398.2 & 3010 & 3027.2 & 2395.5 & \textbf{2357.5} & 2359.2 & 2885 & 2791.8 & 2395.5 & \textbf{2340.2} \\  
86 & 2345.8 & 2875.8 & 3005.5 & 2310.8 & \textbf{2282.8} & 2303 & 2629 & 2774 & 2310.8 & \textbf{2265.2} & 2258 & 2530.5 & 2565.8 & 2310.8 & \textbf{2242.2} \\  
87 & 3253.8 & 3375.5 & 3535 & \textbf{2975.2} & 2990.2 & 3157.8 & 3292.5 & 3424.5 & 2975.2 & \textbf{2949.8} & 3097.8 & 3246.8 & 3330 & 2975.2 & \textbf{2908.5} \\  
88 & 5413 & 5495.2 & 5534.5 & 5415.5 & \textbf{5395.2} & 5369.8 & 5432.2 & 5504 & 5415.5 & \textbf{5362.2} & 5324.5 & 5400.2 & 5464.5 & 5415.5 & \textbf{5323} \\  
89 & 3884 & 3998.5 & 4004.8 & 3765.8 & \textbf{3748} & 3818 & 3917 & 3951.2 & 3765.8 & \textbf{3729.5} & 3762.8 & 3869.5 & 3906.5 & 3765.8 & \textbf{3693.4} \\  
90 & 3369.5 & 3504 & 3514.5 & 3334.2 & \textbf{3330.1} & 3345 & 3434.2 & 3490.8 & 3334.2 & \textbf{3310.4} & 3329.2 & 3411.8 & 3455.8 & 3334.2 & \textbf{3288.5} \\  
91 & 2501.8 & 2622 & 2823 & 2428.2 & \textbf{2408.8} & 2431.2 & 2559.2 & 2656.2 & 2428.2 & \textbf{2388.5} & 2400.8 & 2526 & 2590 & 2428.2 & \textbf{2362} \\  
92 & 3609 & 3743.2 & 3776 & \textbf{3224.5} & 3274.8 & 3559.5 & 3654 & 3719.5 & \textbf{3199.8} & 3237.4 & 3499.5 & 3603.2 & 3642.5 & \textbf{3197} & 3210.1 \\  
93 & 2202.2 & 2320.5 & 2422.8 & 2148 & \textbf{2145.8} & 2168.5 & 2261.8 & 2325.5 & 2148 & \textbf{2138.9} & 2148.5 & 2227.2 & 2269.8 & 2148 & \textbf{2133} \\  
94 & 4474.5 & 4579.2 & 4639.5 & 4441 & \textbf{4427.2} & 4433.2 & 4525 & 4597.8 & 4441 & \textbf{4409.8} & 4389.8 & 4494.8 & 4568.8 & 4441 & \textbf{4380} \\  
95 & 2702.5 & 2834.8 & 2872.5 & 2609.8 & \textbf{2594.1} & 2616.2 & 2748.2 & 2806.8 & 2609.8 & \textbf{2578.9} & 2581.8 & 2713 & 2762.2 & 2609.8 & \textbf{2558.1} \\  
96 & 3487.5 & 3658.5 & 3683.8 & \textbf{3123.2} & 3192.8 & 3418.2 & 3572 & 3628.5 & \textbf{3123.2} & 3153.1 & 3336.2 & 3529.8 & 3588.5 & 3123.2 & \textbf{3108.4} \\  
97 & 2560.8 & 3185.2 & 3427.5 & 2556.8 & \textbf{2532.2} & 2527.8 & 2776 & 3197 & 2556.8 & \textbf{2511.4} & 2496.2 & 2682 & 2813 & 2556.8 & \textbf{2490.5} \\  
98 & 5589 & 5698.8 & 5714.5 & 5548 & \textbf{5516.1} & 5524.8 & 5631 & 5682 & 5548 & \textbf{5496.2} & 5477.5 & 5584.8 & 5646.5 & 5548 & \textbf{5461} \\  
99 & 2157.2 & 2265.2 & 2322.5 & \textbf{1986.5} & 2013.8 & 2114.2 & 2204 & 2246 & \textbf{1986.5} & 2003.4 & 2071.8 & 2177 & 2194.2 & 1986.5 & \textbf{1985.8} \\  
100 & 3349 & 3434.5 & 3494.5 & 3313.5 & \textbf{3309} & 3317.8 & 3403.5 & 3454.2 & 3313.5 & \textbf{3295.6} & 3281.8 & 3388.5 & 3431.5 & 3313.5 & \textbf{3271.6} \\  
\midrule
Avg. 51--100 &  2794.7 & 2907.6 & 2967.3 & 2729.5 & \textbf{2715.1} & 2742.0 & 2845.5 & 2906.0 & 2728.9 & \textbf{2692.8} & 2700.9 & 2811.6 & 2843.0 & 2728.8 & \textbf{2666.4} \\ 
\midrule
Avg. 1--100 & 2168.9 & 2238.6 & 2274.5 & 2153.2 & \textbf{2123.1} & 2129.2 & 2196.3 & 2225.5 & 2152.7 & \textbf{2100.5} & 2098.9 & 2170.6 & 2180.9 & 2152.5 & \textbf{2078.3} \\ 
\midrule
Pooled SD & 12.43 & 37.38 & 33.54 & 8.59 & 10.93 & 11.13 & 27.58 & 20.18 & 8.61 & 11.33 & 13.71 & 22.70 & 14.53 & 8.58 & 9.97 \\
\bottomrule
\end{tabular}}
\end{table}

\begin{table}[!ht]
\centering
\caption{Results of applying CPO to the instances in the LOPS set.}
\label{tab:lops5}
\resizebox{\textwidth}{!}{\begin{tabular}{cccccccccccccccccccc}
\toprule
\multirow{2}{*}{Inst.} & 5 min. & 30 min. & \multicolumn{2}{c}{2 hours} & \multirow{2}{*}{Inst.} & 5 min. & 30 min. & \multicolumn{2}{c}{2 hours} & \multirow{2}{*}{Inst.} & 5 min. & 30 min. & \multicolumn{2}{c}{2 hours} & \multirow{2}{*}{Inst.} & 5 min. & 30 min. & \multicolumn{2}{c}{2 hours} \\
\cmidrule(lr){2-5}
\cmidrule(lr){7-10}
\cmidrule(lr){12-15}
\cmidrule(lr){17-20}
  & UB & UB & LB & UB & & UB & UB & LB & UB &  & UB & UB & LB & UB & & UB & UB & LB & UB \\ 
\midrule
1 & 538 & 530 & 387 & 527 & 26 & 1581 & 1486 & 880 & 1362 & 51 & 2130 & 1991 & 521 & 1734 & 76 & --- & --- & 639 & 2637 \\ 
2 & 663 & 654 & 494 & 650 & 27 & 1833 & 1791 & 1246 & 1790 & 52 & 2540 & 2236 & 553 & 2091 & 77 & --- & --- & 604 & 2010 \\ 
3 & 653 & 635 & 452 & 633 & 28 & 2185 & 2171 & 1396 & 2089 & 53 & 4711 & 3836 & 523 & 3860 & 78 & --- & --- & 636 & 1960 \\ 
4 & 780 & 755 & 562 & 756 & 29 & 2256 & 2298 & 1452 & 2199 & 54 & 3074 & 2086 & 533 & 1661 & 79 & --- & --- & 560 & 2379 \\ 
5 & 860 & 837 & 625 & 828 & 30 & 2473 & 1590 & 1116 & 1769 & 55 & 2416 & 2386 & 497 & 2295 & 80 & --- & --- & 654 & 2392 \\ 
6 & 724 & 718 & 490 & 715 & 31 & 1296 & 1295 & 822 & 1218 & 56 & 2876 & 2939 & 510 & 2706 & 81 & --- & --- & 693 & 3377 \\ 
7 & 964 & 938 & 659 & 923 & 32 & 1214 & 1208 & 776 & 1151 & 57 & 2413 & 2683 & 584 & 2169 & 82 & --- & 5339 & 624 & 5276 \\ 
8 & 1091 & 1044 & 739 & 1039 & 33 & 2698 & 2398 & 1469 & 2276 & 58 & 2520 & 2045 & 823 & 1993 & 83 & --- & 3346 & 701 & 3355 \\ 
9 & 1019 & 966 & 654 & 980 & 34 & 1545 & 1604 & 951 & 1483 & 59 & 3800 & 3998 & 511 & 3764 & 84 & --- & --- & 672 & 2231 \\ 
10 & 902 & 900 & 547 & 790 & 35 & 3145 & 3099 & 1932 & 3049 & 60 & 2375 & 2236 & 587 & 2577 & 85 & --- & --- & 818 & 3786 \\ 
11 & 1290 & 1231 & 857 & 1230 & 36 & 3167 & 2819 & 1788 & 2826 & 61 & --- & --- & 582 & 2845 & 86 & --- & --- & 918 & 3065 \\ 
12 & 1257 & 1223 & 838 & 1180 & 37 & 1463 & 1671 & 924 & 1468 & 62 & 3845 & 3352 & 658 & 3341 & 87 & --- & --- & 696 & 3388 \\ 
13 & 1084 & 1010 & 699 & 1011 & 38 & 1264 & 1200 & 832 & 1192 & 63 & 3169 & 2934 & 1392 & 2782 & 88 & --- & 5963 & 671 & 5956 \\ 
14 & 1557 & 1514 & 1052 & 1486 & 39 & 1880 & 1870 & 1214 & 1845 & 64 & 2960 & 2123 & 914 & 2104 & 89 & --- & --- & 1374 & 4068 \\ 
15 & 1542 & 1476 & 975 & 1445 & 40 & 2463 & 2509 & 1552 & 2439 & 65 & --- & --- & 584 & 2061 & 90 & --- & --- & 646 & 3616 \\ 
16 & 1516 & 1420 & 924 & 1382 & 41 & 2550 & 2522 & 1587 & 2524 & 66 & --- & --- & 611 & 1924 & 91 & --- & --- & 728 & 3449 \\ 
17 & 1131 & 1080 & 763 & 1117 & 42 & 1732 & 1688 & 1111 & 1715 & 67 & --- & --- & 804 & 1930 & 92 & --- & --- & 712 & 3769 \\ 
18 & 2014 & 1918 & 1415 & 1897 & 43 & 2784 & 2779 & 1737 & 2767 & 68 & 3858 & 3343 & 642 & 3238 & 93 & --- & --- & 751 & 2392 \\ 
19 & 1236 & 1046 & 737 & 1028 & 44 & 4030 & 4063 & 2587 & 3908 & 69 & 3787 & 3346 & 615 & 3518 & 94 & --- & --- & 780 & 4883 \\ 
20 & 1135 & 1132 & 671 & 1053 & 45 & 2378 & 2466 & 1446 & 2301 & 70 & 4037 & 4340 & 629 & 4172 & 95 & --- & --- & 726 & 3051 \\ 
21 & 2104 & 1992 & 1378 & 1956 & 46 & 2439 & 2497 & 1539 & 2446 & 71 & 3771 & 2417 & 598 & 2435 & 96 & --- & --- & 1555 & 3662 \\ 
22 & 1639 & 1642 & 985 & 1496 & 47 & 4195 & 3757 & 2518 & 4040 & 72 & --- & --- & 943 & 1949 & 97 & --- & --- & 1650 & 3976 \\ 
23 & 1336 & 1128 & 762 & 1146 & 48 & 1377 & 1506 & 896 & 1473 & 73 & 4946 & 4304 & 624 & 4376 & 98 & --- & --- & 716 & 6117 \\ 
24 & 2135 & 2073 & 1377 & 2010 & 49 & 4065 & 3012 & 2108 & 3233 & 74 & 5062 & 4668 & 713 & 4602 & 99 & --- & --- & 690 & 2825 \\ 
25 & 1524 & 1442 & 892 & 1367 & 50 & 2244 & 1570 & 931 & 1505 & 75 & 4185 & 4177 & 711 & 4046 & 100 & --- & --- & 682 & 3717 \\ 
\midrule
&&&&&&&&&&&&&&& Avg. 1--100 & 2263.4 & 2195.4 & & 2402.2\\
\bottomrule
\end{tabular}}
\end{table}

The results in Figure~\ref{fig:exp_lops} show that, differently from the previous experiments with the medium-sized OPS instances, in the large-sized instances no method obtains the smallest average makespan regardless of the considered time instant. TS outperforms all the other methods for any instant $t \leq 650$ seconds while DE outperforms all the other methods for any instant $t \geq 650$ seconds. Another difference concerning the medium-sized instances is that CPO was outperformed by all introduced metaheuristic approaches. The numerical values in Tables~\ref{tab:lops1}--\ref{tab:lops4} reflect the results already observed in Figure~\ref{fig:exp_lops}. TS found the best results for small-time limits while DE found the best results for large time limits. From the average results at the end of Tables~\ref{tab:lops2} and~\ref{tab:lops4} we can see that the methods rank (a) TS, DE, GA, ILS, and CPO; (b) DE, TS, GA, ILS, and CPO; and (c) DE, TS, GA, ILS, and CPO, when the CPU time limit is 5 minutes, 30 minutes, and 2 hours, respectively, independently of whether we consider the \textit{best} or the \textit{average} makespan as a performance measure.

The observations described in the paragraph above led us to consider a combined approach, named TS+DE, that uses TS to construct an initial population for DE. The combined approach has three phases. In the first phase, TS is used to obtain a solution. Instead of running the method until it reaches the CPU time limit, the search is stopped if the incumbent solution is not updated during a period of $\log_{10}(o)$ seconds of CPU time, recalling that~$o$ is the number of operations of an instance. In the second phase, a population is constructed by running the local search procedure starting from $n_{\size}-1$ perturbations of the TS solution. The perturbation procedure is the one described for the ILS algorithm in Section~\ref{sec:metaheuristic}. The solution of the TS plus the $n_{\size}-1$ solutions found with the local search constitute the initial population of DE. Running DE with this initial population is the third phase of the strategy. The three-phase strategy is interrupted at any time if the CPU time limit is reached. In this strategy, parameters of TS, DE, and the perturbation procedure of ILS were set as already calibrated for each individual method. 

Figure~\ref{fig:exp_mops} and Table~\ref{tab:exp_mops} show the performance of the combined approach when applied to the OPS instances in the MOPS set while Figure~\ref{fig:exp_lops} and Tables~\ref{tab:lops1}--\ref{tab:lops4} show the performance of the combined approach when applied to the OPS instances in the LOPS set. Specifically for the instances in the MOPS set, TS+DE (with a CPU time limit of at least~30 minutes) finds the optimal solutions in the~14 instances with the known optimal solution and improves the solutions found by CPO in the~6 instances with a non-null gap. Figures and tables show that TS+DE is the most successful approach. It always found the lowest average makespan in the MOPS and LOPS sets independent of the CPU time limit imposed. It found the lowest best and average makespans and it found the largest number of best solutions among all considered methods, outperforming CPO by a large extent. It is worth noting that, in the LOPS set, considering the average makespan, the difference between the metaheuristics that rank in first and last places is not larger than 7\%, 6\%, or 5\%, depending on whether the CPU time limit is 5 minutes, 30 minutes, or 2 hours, respectively. This result is not surprising since the four metaheuristic approaches share the representation scheme and the definition of the neighborhood in the local search strategy. On the other hand, the difference between TS+DE and CPO, with a CPU time limit of~2 hours, is 16\%. (With CPU time limits of~5 and~30 minutes, CPO failed in obtaining feasible solutions in~30 and~27 instances, respectively.)

\subsection{Experiment with FJS and FJS with sequencing flexibility scheduling problems}

In this section, in order to asses the performance of the TS+DE method with respect to the state-of-the-art in the literature, numerical experiments with classical instances of the FJS and FJS with sequencing flexibility scheduling problems are conducted. Instances, whose main characteristics are shown in Table~\ref{tab:instances}, correspond to the instances introduced in~\cite{brandimarte1993routing}, \cite{hurink1994tabu}, \cite{barnes1996flexible}, \cite{dauzere1997integrated}, and \cite{birgin2014milp}. TS+DE was run 50 times on the instances in sets YFJS, DAFJS, BR, BC, and DP  and 12 times in the instances in set HK. A CPU time limit of 2 hours was imposed. The performances of TS+DE and its competitors are reported in these experiments through the relative error (RE) of the best makespan $mks(M,p)$ that method ``$M$'' found when applied to instance~$p$, with respect to a known lower bound $mks_{\mathrm{LB}}(p)$, given by
\[
RE(M,p) = 100\% \times \frac{mks(M,p) - mks_{\mathrm{LB}}(p)}{mks_{\mathrm{LB}}(p)}.
\]
Lower bounds for instances~$p$ in the sets BR, BC, DP, and HK were taken from~\cite{FJSinstances}. Lower bounds for instances~$p$ in the sets YFJS and DAFJS were computed running CPO with a CPU time limit of 2 hours. TS+DE was compared with ten different methods from the literature that reported results in at least one of the considered sets, namely: (GRASP) GRASP with a multi-level evolutionary local search proposed in~\cite{kemmoe2017}; (HA) hybrid GA and TS proposed in \cite{li2016effective}; (HDE-N$_2$) hybrid DE with local search proposed in \cite{yuan2013flexible}; (HGTS) hybrid GA and TS proposed in \cite{palacios2015genetic}; (HGVNA) hybrid GA and variable neighborhood descent algorithm proposed in \cite{gao2008hybrid}; (BS) Beam Search algorithm introduced in~\cite{birgin2015list}; (KCSA) Knowledge-based Cuckoo Search Algorithm proposed in~\cite{cao2019b}; (ICA+TS) hybrid Imperialist Competitive Algorithm and TS introduced in~\cite{lunardi2019}; (PBGA) priority-based GA introduced in~\cite{cinar2016}; and (SSPR) Scatter search with path relinking introduced in~\cite{gonza2015}. The used lower bounds and the best solutions obtained by TS+DE and its competitors were gathered in tables and can be found in~\citep{ops2supl}. Tables~\ref{tab:re_fjs} and~\ref{tab:re_efjs} show the results. In the tables, for each method~$M$ and each instances' set~${\cal S}$, we report
\[
\frac{1}{|{\cal S}|} \sum_{p \in {\cal S}} RE(M,p).
\]
Besides, the tables also report how often each method found the best solution (among the solutions found by all the methods). The numerical values in both tables show that TS+DE, although developed to address the OPS scheduling problem, achieves a competitive performance in all sets. It is worth noticing that the goal of this comparison is to analyse the effectiveness of the proposed approach. Efficiency is being neglected in the comparison, since methods being compared were run under different environments and with different stopping criteria.

\begin{table}[!ht]
\centering
\caption{Comparison of TS+DE against other methods from the literature on classical instances of the FJS scheduling problem.}
\label{tab:re_fjs}
\resizebox{\textwidth}{!}{
\begin{tabular}{lcccccccccccccccccc} 
\toprule
\multirow{2}{*}{Set} & \multirow{2}{*}{\#inst.} &
\multicolumn{2}{c}{GRASP} &
\multicolumn{2}{c}{HA} &
\multicolumn{2}{c}{HDE-N$_2$} &
\multicolumn{2}{c}{HGTS} &
\multicolumn{2}{c}{HGVNA} &
\multicolumn{2}{c}{PBGA} &
\multicolumn{2}{c}{SSPR} &
\multicolumn{2}{c}{TS+DE} \\
\cmidrule{3-18}
& & RE & \#best & RE & \#best & RE & \#best & RE & \#best & RE & \#best & RE & \#best & RE & \#best & RE & \#best \\
\midrule 
BR     & 10 & 14.916 &  8 & 14.613 &  9 & 14.674 &  9 & 14.674 &  9 & 14.916 & 8 & 17.982 & 5 & 14.553 & 10 & 14.613 &  9 \\
BC     & 21 & 22.321 & 21 & 22.383 & 15 & 22.386 & 14 & 22.388 & 14 & 22.612 & 9 & 22.508 & 9 & 22.358 & 18 & 22.321 & 21 \\
DP     & 18 &  1.885 &  3 &  1.823 &  2 &    |   &  | &  1.730 &  5 &  2.124 & 0 &    |   & | &  1.567 & 11 &  1.594 &  7 \\
HK (E) & 43 &  2.017 & 38 &  2.125 & 32 &    |   &  | &    |   &  | &     |  & | &    |   & | &  2.035 & 36 &  1.983 & 43 \\ 
HK (R) & 43 &  0.998 & 36 &  1.162 & 22 &    |   &  | &    |   &  | &     |  & | &    |   & | &  1.029 & 34 &  1.082 & 30 \\ 
HK (V) & 43 &  0.082 & 30 &  0.073 & 34 &    |   &  | &    |   &  | &     |  & | &    |   & | &  0.035 & 38 &  0.022 & 43 \\
\bottomrule
\end{tabular}}
\end{table}

\begin{table}[!ht]
\centering
\caption{Comparison of TS+DE against other methods from the literature on the FJS with sequencing flexibility instances introduced in~\cite{birgin2014milp}.}
\label{tab:re_efjs}
{\footnotesize
\begin{tabular}{lcrcrcrcrcrc} 
\toprule
\multirow{2}{*}{Set} & \multirow{2}{*}{\#inst.} & \multicolumn{2}{c}{CPO} & \multicolumn{2}{c}{BS} & \multicolumn{2}{c}{KCSA} & \multicolumn{2}{c}{ICA+TS} & \multicolumn{2}{c}{TS+DE} \\
\cmidrule{3-12}
& & RE & \#best & RE & \#best  & RE & \#best & RE & \#best \\
\midrule
YFJS  & 20 &  0.000 &20& 12.300 &0& 16.939 & 0 &  0.107 &18&  0.000 &20\\
DAFJS & 30 & 30.348 &12& 38.997 &2& 49.681 & 1 & 34.047 & 5& 29.378 &30\\
\bottomrule
\end{tabular}}
\end{table}

\section{Conclusions and future work}

We tackled a challenging real-world scheduling problem named Online Printing Shop (OPS) scheduling problem. The problem was formally defined through mixed integer linear programming and constraint programming formulations in~\cite{ops1}, where the possibility of using the CP Optimizer in practice was analyzed. In the present work, metaheuristic approaches to the problem were proposed. All proposed methods rely on a common representation scheme and a neighborhood adapted from the classical local search introduced in~\cite{mastrolilli2000effective} for the FJS scheduling problem. While considering the sequencing flexibility in the local search is somehow immediate, this is definitely not the case for fixed operations, machines' downtimes, and resumable operations. Two populational and two trajectory metaheuristics were considered and, finally, a combined approach was the one that presented the best performance. The resulting method outperformed by a large extent the results obtained with the CP Optimizer. When applied to classical instances of the FJS scheduling problem and FJS scheduling problem with sequencing flexibility from the literature, the approach introduced in this paper proved to have a competitive performance. The problem addressed in the present work is a real-world problem from the printing industry in Europe. The introduced approach recently started to be tested in practice with a partner company.

\section*{Acknowledgement}

\noindent
This work has been partially supported by FAPESP (grants 2013/07375-0, 2016/01860-1, and 2018/24293-0) and CNPq (grants 306083/2016-7 and 302682/2019-8). The experiments presented in this paper were carried out using the HPC facilities of the University of Luxembourg~\citep{VBCG_HPCS14} --- see \url{https://hpc.uni.lu}. 

\bibliographystyle{plainnat}
\setlength{\bibsep}{0.0pt}
{\small
\bibliography{ops2}

\begin{thebibliography}{50}
\providecommand{\natexlab}[1]{#1}
\providecommand{\url}[1]{\texttt{#1}}
\expandafter\ifx\csname urlstyle\endcsname\relax
  \providecommand{\doi}[1]{doi: #1}\else
  \providecommand{\doi}{doi: \begingroup \urlstyle{rm}\Url}\fi

\bibitem[Ali et~al.(2012)Ali, Siarry, and Pant]{ali2012efficient}
M.~Ali, P.~Siarry, and M.~Pant.
\newblock An efficient differential evolution based algorithm for solving
  multi-objective optimization problems.
\newblock \emph{European Journal of Operational Research}, 217\penalty0
  (2):\penalty0 404--416, 2012.
\newblock \doi{10.1016/j.ejor.2011.09.025}.

\bibitem[Alvarez-Vald{\'e}s et~al.(2005)Alvarez-Vald{\'e}s, Fuertes, Tamarit,
  Gim{\'e}nez, and Ramos]{alvarez2005heuristic}
R.~Alvarez-Vald{\'e}s, A.~Fuertes, J.~M. Tamarit, G.~Gim{\'e}nez, and R.~Ramos.
\newblock A heuristic to schedule flexible job-shop in a glass factory.
\newblock \emph{European Journal of Operational Research}, 165\penalty0
  (2):\penalty0 525--534, 2005.
\newblock \doi{10.1016/j.ejor.2004.04.020}.

\bibitem[Andrade-Pineda et~al.(2020)Andrade-Pineda, Canca, Gonzalez-R, and
  Calle]{andradescheduling}
J.~L. Andrade-Pineda, D.~Canca, P.~L. Gonzalez-R, and M.~Calle.
\newblock Scheduling a dual-resource flexible job shop with makespan and due
  date-related criteria.
\newblock \emph{Annals of Operations Research}, 291:\penalty0 5--35, 2020.
\newblock \doi{10.1007/s10479-019-03196-0}.

\bibitem[Barnes and Chambers(1996)]{barnes1996flexible}
J.~W. Barnes and J.~B. Chambers.
\newblock Flexible job shop scheduling by tabu search.
\newblock Technical Report ORP96-09, Graduate Program in Operations and
  Industrial Engineering, The University of Texas at Austin, Austin, TX, 1996.

\bibitem[Birgin et~al.(2014)Birgin, Feofiloff, Fernandes, De~Melo, Oshiro, and
  Ronconi]{birgin2014milp}
E.~G. Birgin, P.~Feofiloff, C.~G. Fernandes, E.~L. De~Melo, M.~T.~I. Oshiro,
  and D.~P. Ronconi.
\newblock A milp model for an extended version of the flexible job shop
  problem.
\newblock \emph{Optimization Letters}, 8\penalty0 (4):\penalty0 1417--1431,
  2014.
\newblock \doi{10.1007/s11590-013-0669-7}.

\bibitem[Birgin et~al.(2015)Birgin, Ferreira, and Ronconi]{birgin2015list}
E.~G. Birgin, J.~E. Ferreira, and D.~P. Ronconi.
\newblock List scheduling and beam search methods for the flexible job shop
  scheduling problem with sequencing flexibility.
\newblock \emph{European Journal of Operational Research}, 247\penalty0
  (2):\penalty0 421--440, 2015.
\newblock \doi{10.1016/j.ejor.2015.06.023}.

\bibitem[Brandimarte(1993)]{brandimarte1993routing}
P.~Brandimarte.
\newblock Routing and scheduling in a flexible job shop by tabu search.
\newblock \emph{Annals of Operations Research}, 41\penalty0 (3):\penalty0
  157--183, 1993.
\newblock \doi{10.1007/BF02023073}.

\bibitem[Cao et~al.(2019)Cao, Lin, and Zhou]{cao2019b}
Z.~Cao, C.~Lin, and M.~Zhou.
\newblock A knowledge-based cuckoo search algorithm to schedule a flexible job
  shop with sequencing flexibility.
\newblock \emph{IEEE Transactions on Automation Science and Engineering}, pages
  1--14, 2019.
\newblock \doi{10.1109/TASE.2019.2945717}.

\bibitem[Chaudhry and Khan(2016)]{chaudhry2016research}
I.~A. Chaudhry and A.~A. Khan.
\newblock A research survey: review of flexible job shop scheduling techniques.
\newblock \emph{International Transactions in Operational Research},
  23\penalty0 (3):\penalty0 551--591, 2016.
\newblock \doi{10.1111/itor.12199}.

\bibitem[Cinar et~al.(2015)Cinar, Topcu, and Oliveira]{cinar2015}
D.~Cinar, Y.~I. Topcu, and J.~A. Oliveira.
\newblock A taxonomy for the flexible job shop scheduling problem.
\newblock In A.~Migdalas and A.~Karakitsiou, editors, \emph{Optimization,
  Control, and Applications in the Information Age}, pages 17--37, Cham, 2015.
  Springer International Publishing.
\newblock \doi{10.1007/978-3-319-18567-5_2}.

\bibitem[Cinar et~al.(2016)Cinar, Oliveira, Topcu, and Pardalos]{cinar2016}
D.~Cinar, J.~A. Oliveira, Y.~I. Topcu, and P.~M. Pardalos.
\newblock A priority-based genetic algorithm for a flexible job shop scheduling
  problem.
\newblock \emph{Journal of Industrial and Management Optimization},
  12:\penalty0 1391--1415, 2016.
\newblock \doi{10.3934/jimo.2016.12.1391}.

\bibitem[Damak et~al.(2009)Damak, Jarboui, Siarry, and
  Loukil]{damak2009differential}
N.~Damak, B.~Jarboui, P.~Siarry, and T.~Loukil.
\newblock Differential evolution for solving multi-mode resource-constrained
  project scheduling problems.
\newblock \emph{Computers \& Operations Research}, 36\penalty0 (9):\penalty0
  2653--2659, 2009.
\newblock \doi{10.1016/j.cor.2008.11.010}.

\bibitem[Dauz{\`e}re-P{\'e}r{\`e}s and Paulli(1997)]{dauzere1997integrated}
S.~Dauz{\`e}re-P{\'e}r{\`e}s and J.~Paulli.
\newblock An integrated approach for modeling and solving the general
  multiprocessor job-shop scheduling problem using tabu search.
\newblock \emph{Annals of Operations Research}, 70:\penalty0 281--306, 1997.
\newblock \doi{10.1023/A:1018930406487}.

\bibitem[Deb and Agrawal(1995)]{deb1995simulated}
K.~Deb and R.~B. Agrawal.
\newblock Simulated binary crossover for continuous search space.
\newblock \emph{Complex Systems}, 9\penalty0 (2):\penalty0 115--148, 1995.

\bibitem[Deb and Agrawal(1999)]{deb1999niched}
K.~Deb and S.~Agrawal.
\newblock A niched-penalty approach for constraint handling in genetic
  algorithms.
\newblock In \emph{Artificial Neural Nets and Genetic Algorithms}, pages
  235--243, Vienna, 1999. Springer Vienna.
\newblock \doi{10.1007/978-3-7091-6384-9_40}.

\bibitem[Deb and Deb(2014)]{deb2014analysing}
K.~Deb and D.~Deb.
\newblock Analysing mutation schemes for real-parameter genetic algorithms.
\newblock \emph{International Journal of Artificial Intelligence and Soft
  Computing}, 4\penalty0 (1):\penalty0 1--28, 2014.

\bibitem[Gan and Lee(2002)]{gan}
P.~Y. Gan and K.~S. Lee.
\newblock Scheduling of flexible-sequenced process plans in a mould
  manufacturing shop.
\newblock \emph{International Journal of Advanced Manufacturing Technology},
  20:\penalty0 214--222, 2002.
\newblock \doi{10.1007/s001700200144}.

\bibitem[Gao et~al.(2008)Gao, Sun, and Gen]{gao2008hybrid}
J.~Gao, L.~Sun, and M.~Gen.
\newblock A hybrid genetic and variable neighborhood descent algorithm for
  flexible job shop scheduling problems.
\newblock \emph{Computers \& Operations Research}, 35\penalty0 (9):\penalty0
  2892--2907, 2008.
\newblock \doi{10.1016/j.cor.2007.01.001}.

\bibitem[Garey et~al.(1976)Garey, Johnson, and Sethi]{garey}
M.~R. Garey, D.~S. Johnson, and R.~Sethi.
\newblock The complexity of flowshop and jobshop scheduling.
\newblock \emph{Mathematics of Operations Research}, 1\penalty0 (2):\penalty0
  117--129, 1976.

\bibitem[Glover(1986)]{glover1986future}
F.~Glover.
\newblock Future paths for integer programming and links to artificial
  intelligence.
\newblock \emph{Computers \& Operations Research}, 13\penalty0 (5):\penalty0
  533--549, 1986.
\newblock \doi{10.1016/0305-0548(86)90048-1}.

\bibitem[Glover(1997)]{glover1997tabu}
F.~Glover.
\newblock Tabu search and adaptive memory programming --- {A}dvances,
  applications and challenges.
\newblock In R.~S. Barr, R.~V. Helgason, and J.~L. Kennington, editors,
  \emph{Interfaces in Computer Science and Operations Research: Advances in
  Metaheuristics, Optimization, and Stochastic Modeling Technologies}, pages
  1--75. Springer US, Boston, MA, 1997.
\newblock \doi{10.1007/978-1-4615-4102-8_1}.

\bibitem[Goldberg and Holland(1988)]{goldberg1988genetic}
D.~E. Goldberg and J.~H. Holland.
\newblock Genetic algorithms and machine learning.
\newblock \emph{Machine Learning}, 3:\penalty0 95--99, 1988.
\newblock \doi{10.1007/BF00113892}.

\bibitem[Gonz\'alez et~al.(2015)Gonz\'alez, Vela, and Varela]{gonza2015}
M.~A. Gonz\'alez, C.~R. Vela, and R.~Varela.
\newblock Scatter search with path relinking for the flexible job shop
  scheduling problem.
\newblock \emph{European Journal of Operational Research}, 245:\penalty0
  35--45, 2015.
\newblock \doi{10.1016/j.ejor.2015.02.052}.

\bibitem[Holland(1992)]{holland1992adaptation}
J.~H. Holland.
\newblock \emph{Adaptation in Natural and Artificial Systems --- {A}n
  introductory analysis with applications to biology, control, and artificial
  intelligence}.
\newblock MIT Press, Cambridge, MA, 1992.

\bibitem[Hurink et~al.(1994)Hurink, Jurisch, and Thole]{hurink1994tabu}
J.~Hurink, B.~Jurisch, and M.~Thole.
\newblock Tabu search for the job-shop scheduling problem with multi-purpose
  machines.
\newblock \emph{Operations-Research-Spektrum}, 15\penalty0 (4):\penalty0
  205--215, 1994.
\newblock \doi{10.1007/BF01719451}.

\bibitem[Kemmo\'e-Tchomté et~al.(2017)Kemmo\'e-Tchomté, Lamy, and
  Tchernev]{kemmoe2017}
S.~Kemmo\'e-Tchomté, D.~Lamy, and N.~Tchernev.
\newblock An effective multi-start multi-level evolutionary local search for
  the flexible job-shop problem.
\newblock \emph{Engineering Applications of Artificial Intelligence},
  62:\penalty0 80--95, 2017.
\newblock \doi{10.1016/j.engappai.2017.04.002}.

\bibitem[Kim et~al.(2003)Kim, Park, and Ko]{kim}
Y.~K. Kim, K.~Park, and J.~Ko.
\newblock A symbiotic evolutionary algorithm for the integration of process
  planning and job shop scheduling.
\newblock \emph{Computers \& Operations Research}, 30:\penalty0 1151--1171,
  2003.
\newblock \doi{10.1016/S0305-0548(02)00063-1}.

\bibitem[Laborie et~al.(2018)Laborie, Rogerie, Shaw, and Vil\'im]{Laborie2018}
P.~Laborie, J.~Rogerie, P.~Shaw, and P.~Vil\'im.
\newblock {IBM} {ILOG} {CP} {O}ptimizer for {S}cheduling.
\newblock \emph{Constraints}, 23\penalty0 (2):\penalty0 210--250, 2018.

\bibitem[Lee et~al.(2012)Lee, Moon, Bae, and Kim]{lee2012}
S.~Lee, I.~Moon, H.~Bae, and J.~Kim.
\newblock Flexible job-shop scheduling problems with ‘and’/‘or’
  precedence constraints.
\newblock \emph{International Journal of Production Research}, 50\penalty0
  (7):\penalty0 1979--2001, 2012.
\newblock \doi{10.1080/00207543.2011.561375}.

\bibitem[Li and Gao(2016)]{li2016effective}
X.~Li and L.~Gao.
\newblock An effective hybrid genetic algorithm and tabu search for flexible
  job shop scheduling problem.
\newblock \emph{International Journal of Production Economics}, 174:\penalty0
  93--110, 2016.
\newblock \doi{10.1016/j.ijpe.2016.01.016}.

\bibitem[Louren{\c{c}}o et~al.(2003)Louren{\c{c}}o, Martin, and
  St{\"u}tzle]{lourencco2003iterated}
H.~R. Louren{\c{c}}o, O.~C. Martin, and T.~St{\"u}tzle.
\newblock Iterated local search.
\newblock In \emph{Handbook of metaheuristics}, pages 320--353. Springer,
  Boston, MA, 2003.
\newblock \doi{10.1007/0-306-48056-5_11}.

\bibitem[Lunardi(2020)]{lunardi2020}
W.~T. Lunardi.
\newblock \emph{A Real-World Flexible Job Shop Scheduling Problem With
  Sequencing Flexibility: Mathematical Programming, Constraint Programming, and
  Metaheuristics}.
\newblock PhD thesis, University of Luxembourg, Luxembourg, 2020.
\newblock URL \url{https://orbilu.uni.lu/handle/10993/43893}.

\bibitem[Lunardi et~al.(2019)Lunardi, Voos, and Cherri]{lunardi2019}
W.~T. Lunardi, H.~Voos, and L.~H. Cherri.
\newblock An effective hybrid imperialist competitive algorithm and tabu search
  for an extended flexible job shop scheduling problem.
\newblock In \emph{Proceedings of the 34th ACM/SIGAPP Symposium on Applied
  Computing (SAC’19)}, pages 204--211, New York, NY, 2019. Association for
  Computing Machinery.
\newblock \doi{10.1145/3297280.3297302}.

\bibitem[Lunardi et~al.(2020{\natexlab{a}})Lunardi, Birgin, Laborie, Ronconi,
  and Voos]{ops1}
W.~T. Lunardi, E.~G. Birgin, P.~Laborie, D.~P. Ronconi, and H.~Voos.
\newblock Mixed integer linear programming and constraint programming models
  for the online printing shop scheduling problem.
\newblock \emph{Computers \& Operations Research}, 123:\penalty0 Article number
  105020, 2020{\natexlab{a}}.
\newblock \doi{10.1016/j.cor.2020.105020}.

\bibitem[Lunardi et~al.(2020{\natexlab{b}})Lunardi, Birgin, Ronconi, and
  Voos]{ops2supl}
W.~T. Lunardi, E.~G. Birgin, D.~P. Ronconi, and H.~Voos.
\newblock Metaheuristics for the online printing shop scheduling problem --
  {S}upplementary material.
\newblock Technical Report 10993/43275, University of Luxembourg,
  2020{\natexlab{b}}.
\newblock (Available at \url{https://orbilu.uni.lu/handle/10993/43275}.).

\bibitem[Mastrolilli and Gambardella(1999)]{FJSinstances}
M.~Mastrolilli and L.~M. Gambardella.
\newblock Effective neighborhood functions for the flexible job shop problem:
  Appendix.
\newblock \url{http://people.idsia.ch/~monaldo/fjspresults/fjsp\_result.ps},
  1999.
\newblock accessed on April 10, 2020.

\bibitem[Mastrolilli and Gambardella(2000)]{mastrolilli2000effective}
M.~Mastrolilli and L.~M. Gambardella.
\newblock Effective neighbourhood functions for the flexible job shop problem.
\newblock \emph{Journal of Scheduling}, 3\penalty0 (1):\penalty0 3--20, 2000.

\bibitem[{\"O}zg{\"u}ven et~al.(2010){\"O}zg{\"u}ven, {\"O}zbak{\i}r, and
  Yavuz]{ozguven2010mathematical}
C.~{\"O}zg{\"u}ven, L.~{\"O}zbak{\i}r, and Y.~Yavuz.
\newblock Mathematical models for job-shop scheduling problems with routing and
  process plan flexibility.
\newblock \emph{Applied Mathematical Modelling}, 34\penalty0 (6):\penalty0
  1539--1548, 2010.
\newblock \doi{10.1016/j.apm.2009.09.002}.

\bibitem[Palacios et~al.(2015)Palacios, Gonz{\'a}lez, Vela,
  Gonz{\'a}lez-Rodr{\'\i}guez, and Puente]{palacios2015genetic}
J.~J. Palacios, M.~A. Gonz{\'a}lez, C.~R. Vela, I.~Gonz{\'a}lez-Rodr{\'\i}guez,
  and J.~Puente.
\newblock Genetic tabu search for the fuzzy flexible job shop problem.
\newblock \emph{Computers \& Operations Research}, 54:\penalty0 74--89, 2015.
\newblock \doi{10.1016/j.cor.2014.08.023}.

\bibitem[Price et~al.(2006)Price, Storn, and Lampinen]{price2006differential}
K.~Price, R.~M. Storn, and J.~A. Lampinen.
\newblock \emph{Differential evolution -- A practical approach to global
  optimization}.
\newblock Springer Science \& Business Media, 2006.
\newblock \doi{10.1007/3-540-31306-0}.

\bibitem[Qin et~al.(2008)Qin, Huang, and Suganthan]{qin2008differential}
A.~K. Qin, V.~L. Huang, and P.~N. Suganthan.
\newblock Differential evolution algorithm with strategy adaptation for global
  numerical optimization.
\newblock \emph{IEEE Transactions on Evolutionary Computation}, 13\penalty0
  (2):\penalty0 398--417, 2008.
\newblock \doi{10.1109/TEVC.2008.927706}.

\bibitem[Reeves and Rowe(2002)]{reeves2002genetic}
C.~Reeves and J.~E. Rowe.
\newblock \emph{Genetic algorithms: Principles and perspectives -- A guide to
  GA theory}.
\newblock Springer US, 2002.
\newblock \doi{10.1007/b101880}.

\bibitem[Rossi and Lanzetta(2020)]{rossi}
A.~Rossi and M.~Lanzetta.
\newblock Integration of hybrid additive/subtractive manufacturing planning and
  scheduling by metaheuristics.
\newblock \emph{Computers \& Industrial Engineering}, 144:\penalty0 Article ID
  106428, 2020.
\newblock \doi{10.1016/j.cie.2020.106428}.

\bibitem[Storn and Price(1997)]{storn1997differential}
R.~Storn and K.~Price.
\newblock Differential evolution -- {A} simple and efficient heuristic for
  global optimization over continuous spaces.
\newblock \emph{Journal of Global Optimization}, 11\penalty0 (4):\penalty0
  341--359, 1997.
\newblock \doi{10.1023/A:1008202821328}.

\bibitem[Tsai et~al.(2013)Tsai, Fang, and Chou]{tsai2013optimized}
J.-T. Tsai, J.-C. Fang, and J.-H. Chou.
\newblock Optimized task scheduling and resource allocation on cloud computing
  environment using improved differential evolution algorithm.
\newblock \emph{Computers \& Operations Research}, 40\penalty0 (12):\penalty0
  3045--3055, 2013.
\newblock \doi{10.1016/j.cor.2013.06.012}.

\bibitem[Varrette et~al.(2014)Varrette, Bouvry, Cartiaux, and
  Georgatos]{VBCG_HPCS14}
S.~Varrette, P.~Bouvry, H.~Cartiaux, and F.~Georgatos.
\newblock Management of an academic hpc cluster: The ul experience.
\newblock In \emph{Proc. of the 2014 Intl. Conf. on High Performance Computing
  \& Simulation (HPCS 2014)}, pages 959--967, Bologna, Italy, July 2014. IEEE.

\bibitem[Vilcot and Billaut(2008)]{vilcot2008tabu}
G.~Vilcot and J.-C. Billaut.
\newblock A tabu search and a genetic algorithm for solving a bicriteria
  general job shop scheduling problem.
\newblock \emph{European Journal of Operational Research}, 190\penalty0
  (2):\penalty0 398--411, 2008.
\newblock \doi{10.1016/j.ejor.2007.06.039}.

\bibitem[Vital-Soto et~al.(2020)Vital-Soto, Azab, and Baki]{soto}
A.~Vital-Soto, A.~Azab, and M.~F. Baki.
\newblock Mathematical modeling and a hybridized bacterial foraging
  optimization algorithm for the flexible job-shop scheduling problem with
  sequencing flexibility.
\newblock \emph{Journal of Manufacturing Systems}, 54:\penalty0 74--93, 2020.
\newblock \doi{10.1016/j.jmsy.2019.11.010}.

\bibitem[Wang et~al.(2010)Wang, Pan, Suganthan, Wang, and Wang]{wang2010novel}
L.~Wang, Q.-K. Pan, P.~N. Suganthan, W.-H. Wang, and Y.-M. Wang.
\newblock A novel hybrid discrete differential evolution algorithm for blocking
  flow shop scheduling problems.
\newblock \emph{Computers \& Operations Research}, 37\penalty0 (3):\penalty0
  509--520, 2010.
\newblock \doi{10.1016/j.cor.2008.12.004}.

\bibitem[Yuan and Xu(2013)]{yuan2013flexible}
Y.~Yuan and H.~Xu.
\newblock Flexible job shop scheduling using hybrid differential evolution
  algorithms.
\newblock \emph{Computers \& Industrial Engineering}, 65\penalty0 (2):\penalty0
  246--260, 2013.
\newblock \doi{10.1016/j.cie.2013.02.022}.

\end{thebibliography}
}
\end{document}